\documentclass{article}

\usepackage{microtype}
\usepackage{graphicx}

\usepackage{soul}

\usepackage{subfig}
\usepackage{booktabs} % for professional tables

\usepackage{hyperref}

\usepackage[accepted]{icml2021}

\makeatletter
\renewcommand{\ICML@appearing}{\textit{Preprint under review.}}
\makeatother

\usepackage[section]{placeins}

\icmltitlerunning{Size Matters}

\begin{document}

\twocolumn[
\icmltitle{Size Matters}

% It is OKAY to include author information, even for blind
% submissions: the style file will automatically remove it for you
% unless you've provided the [accepted] option to the icml2021
% package.

% List of affiliations: The first argument should be a (short)
% identifier you will use later to specify author affiliations
% Academic affiliations should list Department, University, City, Region, Country
% Industry affiliations should list Company, City, Region, Country

% You can specify symbols, otherwise they are numbered in order.
% Ideally, you should not use this facility. Affiliations will be numbered
% in order of appearance and this is the preferred way.
\icmlsetsymbol{equal}{*}

\begin{icmlauthorlist}
%\icmlauthor{Aeiau Zzzz}{equal,to}
%\icmlauthor{Bauiu C.~Yyyy}{equal,to,goo}
\icmlauthor{Mats L. Richter}{uos}
\icmlauthor{Wolf Byttner}{rh}
\icmlauthor{Ulf Krumnack}{uos}
\icmlauthor{Ludwig Schallner}{uos}
\icmlauthor{Justin Shenk}{vis}
\end{icmlauthorlist}

\icmlaffiliation{uos}{Department of Cognitive Science, University of Osnabrueck, Lower Saxony, Germany}
\icmlaffiliation{rh}{Rapid Health, London, England, United Kingdom}
\icmlaffiliation{vis}{VisioLab, Berlin, Germany}

\icmlcorrespondingauthor{Mats L. Richter}{matrichter@uos.de}

% You may provide any keywords that you
% find helpful for describing your paper; these are used to populate
% the "keywords" metadata in the PDF but will not be shown in the document
\icmlkeywords{Machine Learning, ICML}

\vskip 0.3in
]

% this must go after the closing bracket ] following \twocolumn[ ...

% This command actually creates the footnote in the first column
% listing the affiliations and the copyright notice.
% The command takes one argument, which is text to display at the start of the footnote.
% The \icmlEqualContribution command is standard text for equal contribution.
% Remove it (just {}) if you do not need this facility.

%\printAffiliationsAndNotice{}  % leave blank if no need to mention equal contribution
\begin{NoHyper}% hack: make the hyperref warning "ignoring empty anchor" go away
\printAffiliationsAndNotice{\icmlEqualContribution} % otherwise use the standard text.
\end{NoHyper}
\setlength{\footskip}{3.30003pt}% hack: make the Fancyhdr Warning " \footskip is too small (0.0pt)" go away

\begin{abstract}
  Fully convolutional neural networks can process input of
  arbitrary size by applying a combination of downsampling and
  pooling.
  However, we find that fully convolutional image classifiers are
  not agnostic to the input size but rather show significant
  differences in performance:
  presenting the same image at different scales can result in
  different outcomes.
  A closer look reveals that there is no simple relationship between
  input size and model performance (no `bigger is better'), but that
  each each network has a preferred input size, for which it shows
  best results.
  We investigate this phenomenon by applying different methods, including
  spectral analysis of layer activations and probe classifiers,
  showing that there are characteristic features depending on the 
  network architecture.
  From this we find that the size of discriminatory features is critically influencing how the inference process is distributed among the layers.
\end{abstract}

\section{Introduction}

The relation between the neural architecture and the resolution of the input image is an often overlooked property of convolutional neural network (CNN) classifier. 
%The size of the input image and its relation to the neural architecture trained on this resolution is an often overlooked property in the training setup of a convolutional neural network classifier.
Early designs like \citet{alexnet, vgg} and \citet{inception} chose square resolutions between $224 \times 224$ and $299 \times 299$ pixels mainly as a compromise between image size and computational efficiency.
In recent publications like \citet{gpipe} the resolution is increased up to $600 \times 600$ pixels.

Furthermore \citet{efficentnet} show experimentally that the input resolution increases the performance reliably on various architectures.
The intuitive explanation offered by authors like \citet{inceptionv3} for this phenomenon is the addition of greater detail gained by increasing the size of the image.
%The intuitive explanation for why in many cases higher input size yields better performance is the addition of detail in the image. 
These details potentially yield additional features that can be detected by the classifier, resulting in higher performance. 

\citet{efficentnet} introduces also the idea that the size of the input image needs to match the neural architecture for optimal efficiency and performance; they use the size of the input image as one of 3 linearly related properties for scaling up neural network architectures.

\subsection{Contribution}
In this work we investigate the following questions addressing the relationship between input size and model architecture:
\begin{itemize}
    \item Has the size of the input image an effect on the predictive performance of CNN classifiers? Answer: yes, it is shown that not just increasing the amount of details but just altering the image resolution can improve classification accuracy (Section~\ref{sec:details}).
    % is independent from the amount of details added or removed by altering the size of the % input image?
    \item Does the size of the input image influence how the information is processed in the network? Answer: yes, we can show that processing significantly differs depending on input size (Section~\ref{sec:inference}) and the size of the depicted objects (Section~\ref{sec:objectsize}).
    \item Can we know in advance which layers will contribute to the quality of the prediction based on the input size? Answer: For strictly sequential architectures the receptive field size allows to identify unproductive layers (Section~\ref{sec:receptivefield}).
    \item Do residual connections influence the observed behavior? Yes, residual connections can help to involve more layers in the inference process (Section~\ref{sec:residual}).
\end{itemize}

%\subsection{Contribution}

%Our primary contribution is to show that the size of the input image has an effect on the predictive performance of CNN classifiers, which is independent from the amount of details added or removed by altering the size of the input image.
%We demonstrate that the input size affects how the inference is distributed in a given neural network architecture. 
%We show that a poorly chosen resolution results in underutilized layers and thus suboptimal performance.
%We experimentally attribute these findings to an interaction between the receptive field sizes of the convolutional layers and the size of discriminatory features on the image.
%We demonstrate that we can predict which layers will not contribute significantly to the inference process in sequential neural architectures.
%Furthermore we investigate the effect of residual connections on the observed inference dynamics. 
%We find that residual connections can distribute the inference process across more layers, leading to an increase in performance and better network utilization.

\section{Background}

\subsection{Fully Convolutional Networks}
This work focuses on fully convolutional neural network classifiers; the current de facto standard for CNN classifier architectures \cite{resnet, inceptionv3, efficentnet, densenet}.
While convolution is a local operation that can in principle deal with input
of arbitrary height and width, the fully connected dense layers used for the output operate globally
and hence enforce specific input and output shapes. 
Therefore architectures such as AlexNet have a fixed input size and require changes to the architecture in order to change the input size \cite{alexnet}.
In contrast, fully convolutional networks only apply a stack of convolutional filters
to the input, resulting in feature maps whose size depends on the size
of the input data. These feature maps can be summarized, usually by global average
pooling, to obtain a feature vector of fixed dimensionality that can be used 
for further processing, e.g. classification \cite{inceptionv3}.
This makes fully convolutional neural networks effectively agnostic to the size of the input.

Being able to change the size of the input image without altering the architecture of the CNN is important for avoiding potential artifacts induced by architectural changes.
For this reason we also modify non-fully convolutional architectures such as the VGG-family of networks by \citet{vgg}  to be fully convolutional.
This is done by simply replacing the readout, which reshapes the tensor produced by a convolutional layer into a vector by a global average pooling layer \cite{gap}.

\subsection{Probe Classifiers}
Probe classifiers were initially proposed by \citet{alain2016} as a tool for analyzing how the solution quality progresses while the data is propagated through neural networks.
In order to do this, logistic regression "probes"  are trained on the same task as the model, using the output of individual layers from the trained model as input.
Since the softmax layer and a cross-entropy minimizing logistic regression effectively solve the same task, we can use these probes to see how the quality of the intermediate solution evolves during the forward pass. 
Usually, probe accuracy will be low for the first layers and increase throughout the network to approximate the model accuracy in the final layers, as the example shown in figure~\ref{fig:vgg16_cifar10} demonstrates.

\subsection{Saturation and Tail Patterns}
\begin{figure}[htb!]
	\centering
	
	\label{fig:tails}
	\subfloat{
	    \includegraphics[width=0.98\columnwidth]{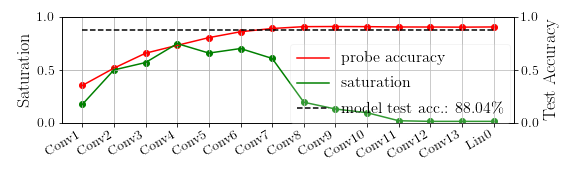}
	    }

	\caption{VGG16 trained on Cifar10 with $32 \times 32$ pixel input size. The model exhibits a tail pattern starting from Conv8. Note that the probe performance in the tail is roughly equal to the performance of the model itself.}
	\label{fig:vgg16_cifar10}
	
\end{figure}

Saturation is another technique for expressing the activity of a neural network layer in a single number, introduced by \citet{saturation}.
%In this work we use saturation as redundant technique to probes in order to analyze the inference process on a layer-by-layer basis.
The saturation metric of a layer is obtained by computing the number of eigendirections in the output required to explain 99\% of the total variance and dividing this number by the dimensionality of the layer's output.\footnote{In case of convolutional layers, each position of the convolutional filter is considered a single sample and the number of output channels is considered the dimensionality of the data.}
Similar to the accuracy of a probe classifier, this results in a value bound between 0 and 1.
Intuitively saturation measures a percentage of how much the output space of a layer is ``filled'' or ``saturated'' with the data.
%The overall saturation level is a function of the problem difficulty, the number of filters and the neural architecture \cite{saturation}. 
\citet{saturation} show that interesting patterns can be observed when the saturation of every layer is plotted in sequence of the forward pass.
Important for our work is the ``tail pattern'', which can be seen in figure \ref{fig:vgg16_cifar10}, as it indicates an inefficiently distributed forward pass.

A tail pattern is a sub-sequence of layers, generally found close to the input or the output of the network, with significantly less saturation than all other layers. 
By observing probe performances we find that these layers generally do not contribute significantly to the inference process by devolving to quasi-pass-through layers over the course of training.
We will refer to these layers as unproductive layers or unproductive sequences. %, when we refer to the tail.
%An example of which can be seen in figure \ref{fig:vgg16_cifar10}.
%We also find in \cite{saturation} that this phenomenon is caused by miss-match of input size and neural architecture.
%This mismatch can be removed by alteration of architecture and input size.
%We can see the effect of the latter in figure \ref{fig:vgg16_bigCifar10}.
%The change in input size has removed the tail pattern from the end of the networks structure, distributed the inference more evenly across the layers and improved the performance.
%Since the models are trained on Cifar10, which has a native resolution of 32 pixels, the effect can not be attributed to the content of the image.
%This work will further elaborate on the effects of the tail patterns.
%We will also test how much performance performance improvement can be attributed to the change in resolution when training on more realistic datasets like iNaturalist and ImageNet.

\subsection{The receptive field size}
The receptive field size is a property of a convolutional layer inside a neural network architecture. 
The receptive field size describes the maximum height and width of a square area\footnote{Technically a rectangle, however since square kernels are the norm we can make this simplification.} on an arbitrarily large input image.
Only pixels contained in this area may influence the output of a single convolution operation.
In other words, the receptive field size describes the spatial upper bound of visual patterns detectable by the respective layer.

For convolutional neural networks with a sequential structure (no multiple pathways during the forward pass) the receptive field size can be computed analytically.
We refer to the receptive field $r$ of the $l$th layer of sequential network structure as $r_l$ (with $r_0 = 1$, which is the "receptive field" of the input).
For all layers $l > 0$ in the convolutional part of a sequential network the receptive field can be computed with the following formula:

\begin{equation}
    r_l = r_{l-1} + ((k_{l} - 2) \prod^{l-1}_{i=0} s_{i})
\end{equation}

where $r_{l-1}$ is the receptive field of the previous layer, $k_l$ refers to the kernel size of layer $l$ (with potential dilation already accounted for) and $s_{i}$ the stride size of layer $i$.
In the simplest case, which is the first convolutional layer of the neural architecture, $r_1$ is exactly the size of the kernel.
When convolutional layers are stacked, the receptive field expands for every layer $l$ with a kernel size $k_l \neq 1$.
The stride size $s_l$ of a layer $l$ furthermore has a multiplicative effect on the growth rate of the receptive field, since a stride effectively downsamples the feature map.

While the ResNet-family of networks by \citet{resnet} is not strictly sequential due to their residual connections, we compute the receptive field size by simply ignoring the residual connections.
We can do this in the context of this work because residual connections do not change the receptive field size and we are only interested in the receptive field size as a spatial  upper bound for the integration of information into single output positions.

\subsection{Choice of Models}
The experiments are primarily conducted on VGG and ResNet-style architectures originally developed by \citet{vgg} and \citet{resnet}.

The purpose of the VGG-style architectures in this work is to serve as a baseline of simple sequential CNN architectures.
The VGG architectures are modified by using a global average pooling layer in the readout in order to make them fully convolutional. 
Furthermore we use variants that utilize batch-normalization after each layer, which has become a widespread pattern in many more recent architectures \cite{resnet, efficentnet, gpipe}.

For more complex architectures we choose the ResNet family and will mainly focus on ResNet18 for analysis in the main part of this work.
The reason for this decision is two-fold. 
First, ResNet networks utilize many ideas of modern neural architectures, like the building block design and residual connections, % that have since become widespread,
while being rather simple in their structure
\cite{efficentnet, densenet, inceptionv3}.
Second, the ResNet family and ResNet18\footnote{For the sake of consistency and comparability, when we refer to ResNet models like ResNet18, we specifically refer to the ImageNet versions of these architectures. In the initial publication, also a Cifar10-Version of ResNet models is proposed, which deviates drastically in the receptive field size and in performance on small image datasets.} in particular are easy to visualize, while many more complex derivative architectures feature a more non-sequential structure and a much larger amount of consecutive layers, making the experimental results harder to read and interpret.

To demonstrate that the claims made in this work generalize we reproduce some results on additional models such as EfficentNet-B0, ResNet34 and ResNet50 and provide them either in the main part of this work or in the supplementary material.

\subsection{Training Setup}
Training on any dataset is conducted using a batch size of 64 (see supplementary material for more details). If no information on the input size is given it can be assumed that the native resolution of the dataset is fed into the network.
The training data is augmented by random cropping, horizontal flipping and channel wise normalization of the data points using ImageNet computed means and standard deviations of every color channel.
When images are resized linear interpolation is used and the resize is applied after data augmentation and any other preprocessing.
We use the stochastic gradient decent optimizer for training. 
The learning rate is set initially to 0.1 and reduced after one and two thirds of the trained epochs by a factor 10.
The models are trained for 30 epochs with the exception of ResNet50, which is trained for 60 epochs due to slow convergence.
No hyperparameter optimization or fine tuning on hyperparameters is conducted, since it is not the goal of the paper to achieve maximum or state-of-the-art performance.
We find that changing the learning rate, optimizer, batch size, data augmentation and number of epochs trained does not influence the conclusions of the experiments in any significant way as long as the model is able to converge to a solution better than chance-level (see supplementary material).

\subsection{Datasets}
Multiple datasets are used to conduct and verify the experiments presented in this work.
Experiments that require small base images were conducted on Cifar10 and reproduced on MNIST and TinyImageNet \cite{tinyimagenet, cifar, mnist}.
Experiments requiring images of higher resolution were conducted on iNaturalist as well as ImageNet \cite{iNaturalist, ImageNet}.

\section{Experiments}
A series of experiments has been conducted to investigate the relationship of input size and model architecture.
We first demonstrate that input size affects the predictive performance of models even though no additional information is added by scaling the images.
We then explore this effect further by reproducing these results on Cifar10 and analyzing the saturation and probe performance.
Based on these results we investigate the role of locality of discriminatory features in the image.
Finally we discuss the role of the receptive field in these observations and analyze the influence of residual connections.

\subsection{Image Size and additional Details have independent effects on the predictive performance}
\label{sec:details}
\citet{efficentnet} showed that increasing the input resolution of a model leads to increased performance.
The intuitive explanation for this effect is that adding details (i.e. information) leads to better performance.
The working hypothesis is that more details is the only factor increasing performance when increasing the image size.
We investigate by increasing the image size, with or without added detail.

We train multiple models on ImageNet and iNaturalist in three different settings A, B and C.
Models of set A, are trained on images of the original design size of $224 \times 224$ pixels, providing the performance baseline.
Set B models are trained on images of size $32 \times 32$. We expect the relative performance of these models to drop significantly in all scenarios, since the image detail is reduced.
The third set of models C is trained on images that are first resized to $32 \times 32$ pixels; then up-sampled to $224 \times 224$ pixels.
These images have as much detail as the images used for training the models of set B while keeping the larger size used for training the models of set A. 
According to the working hypothesis, performance should not increase relative to groups two.
%If the addition of details is the primary source of predictive performance gain, we would expect a performance drop of roughly equal proportion from models of set two and three relative to the baseline model of the same architecture and trained on the same dataset.

\begin{table}
\caption{Relative Top1-Accuracy:
The table shows the predictive performance of trained models relative to a baseline model. 
The baseline uses the same model and architecture and is trained on images resized to $224 \times 224$ pixels.
Training the models on $32 \times 32$ pixel images results in less than half the predictive performance in all tested scenarios relative to the baseline. Resizing the $32 \times 32$ images back to $224 \times 224$ results in a significant recovery of lost performance in all tested scenarios, despite no information is added by upscaling.}
\label{tab:relative_performances}
\begin{tabular}{llccc}
%\multicolumn{4}{c}{Relative Top1-Accuracy} \\
\hline
Model    &  Dataset & \hspace{-2ex}downscaled\hspace{-2ex} & upscaled \\
         &          & $32\times32$ & $224\times224$\\
\cline{1-3}

\hline
VGG16 & ImageNet   & 15.35\% & 66.08\%          \\
\hline
VGG16 & iNaturalist & 36.83\% & 58.38\%         \\
\hline
ResNet18 & ImageNet  & 45.74\% & 67.39\%        \\
\hline
ResNet18 & iNaturalist  & 30.4\% & 52.68\%      \\
\hline
ResNet50 & ImageNet & 28.21\% & 71.8\%          \\
\hline
ResNet50 & iNaturalist & 33.87\% & 54.38\%      \\ 
\hline
EfficientNet-B0\hspace{.3ex} & ImageNet & 19.32\% & 64.45\%      \\ 
\hline
EfficientNet-B0\hspace{.3ex} & iNaturalist & 15.34\% & 63.79\%     \\ 
\hline

\end{tabular}
\end{table}

While performance suffers drastically when reducing the input size, we can also see in table \ref{tab:relative_performances} that decreasing, then increasing the image size regains some of the lost performance with all models.
Based on this observation we can conclude that the size of the input images has an independent effect on the predictive performance.
This also confirms observations made by \citet{saturation}, were the same effect was observed on the Food101 dataset.

\subsection{Input size affects the inference process of the CNN}
\label{sec:inference}
To further understand what effect the input size has on the trained model and its performance, we train the same models on Cifar10.
Cifar10 has a native resolution of $32 \times 32$, therefore no additional details can be added by increasing the size of the image.
We train ResNet18 on Cifar10s native resolution as well as $224 \times 224$ pixels.
We also train the model a third time using an input size of $1024 \times 1024$ pixels, to understand what happens when the image is drastically oversized.

\begin{figure}[htb!]
	\centering
	\subfloat[$32 \times 32$ (Cifar10 native resolution)]{
	    \includegraphics[width=0.8\columnwidth]{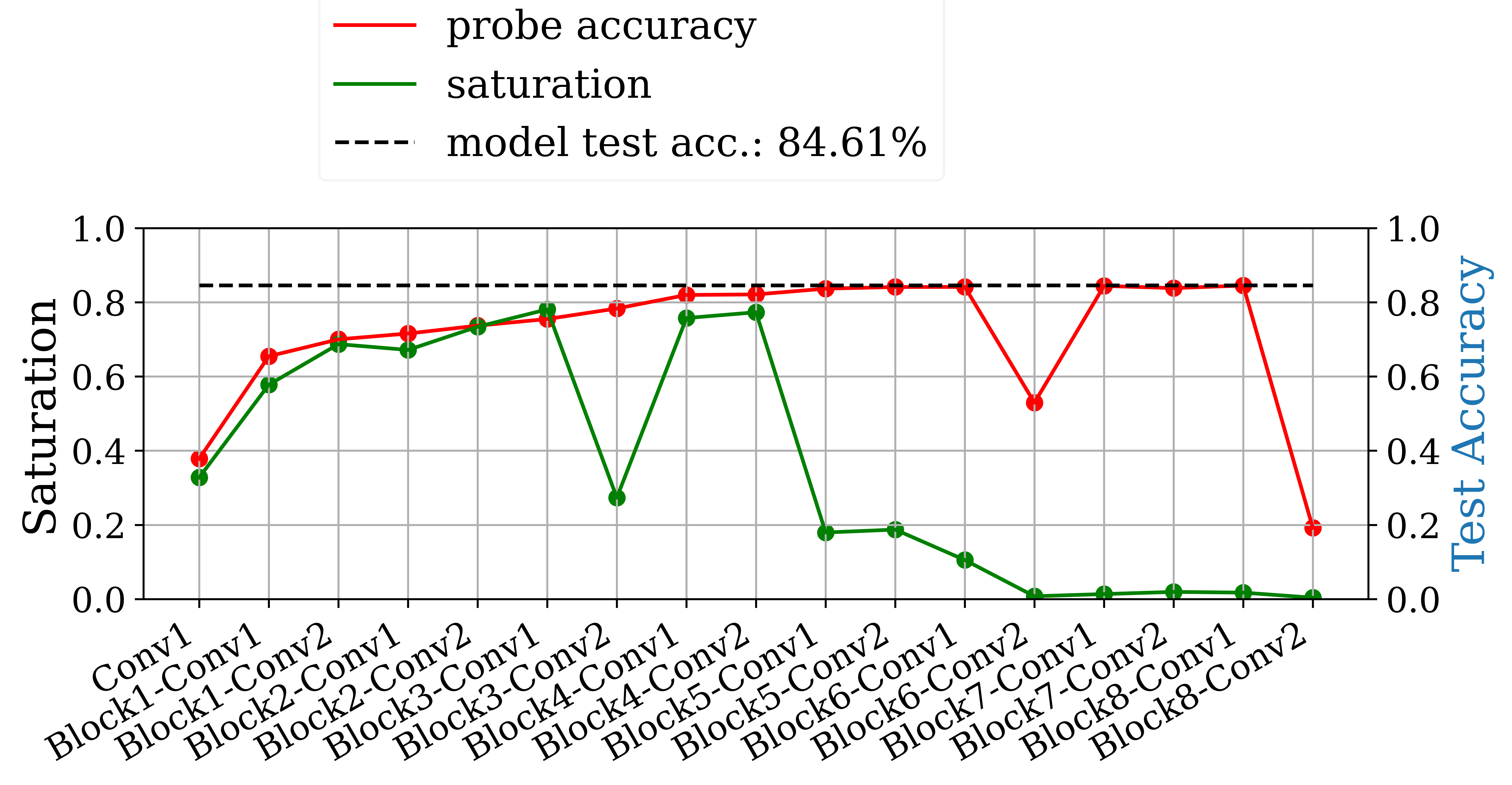}
	    \label{fig:resnet18_cifar10_small}
	    %\label{fig:iNaturalistVgg16_small}
	} \qquad
	\subfloat[$224 \times 224$ (ResNet standard)]{
	    \includegraphics[width=0.8\columnwidth]{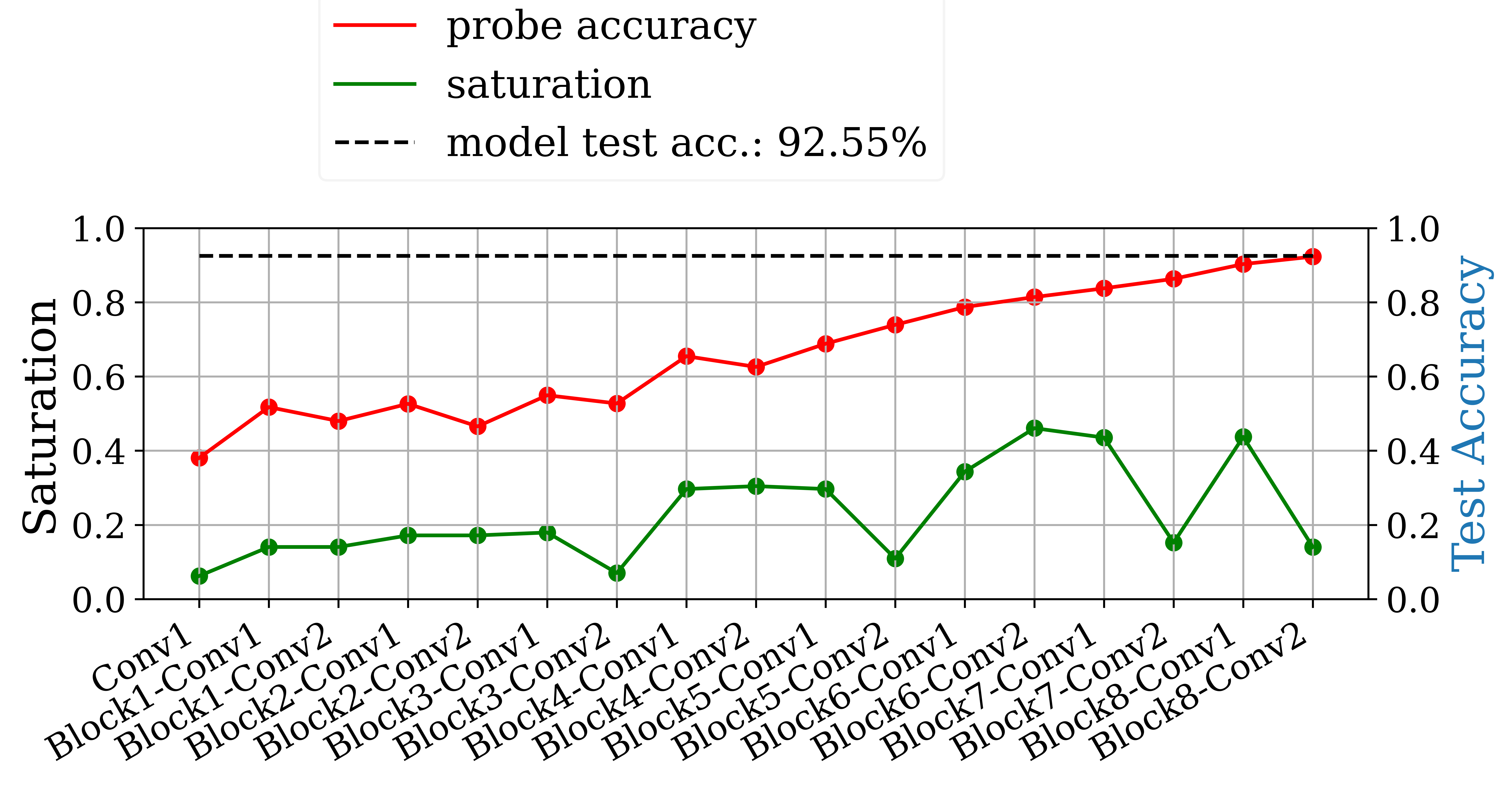}
	    \label{fig:resnet18_cifar10_medium}
	}
	\quad
	\subfloat[$1024 \times 1024$]{
	    \includegraphics[width=0.8\columnwidth]{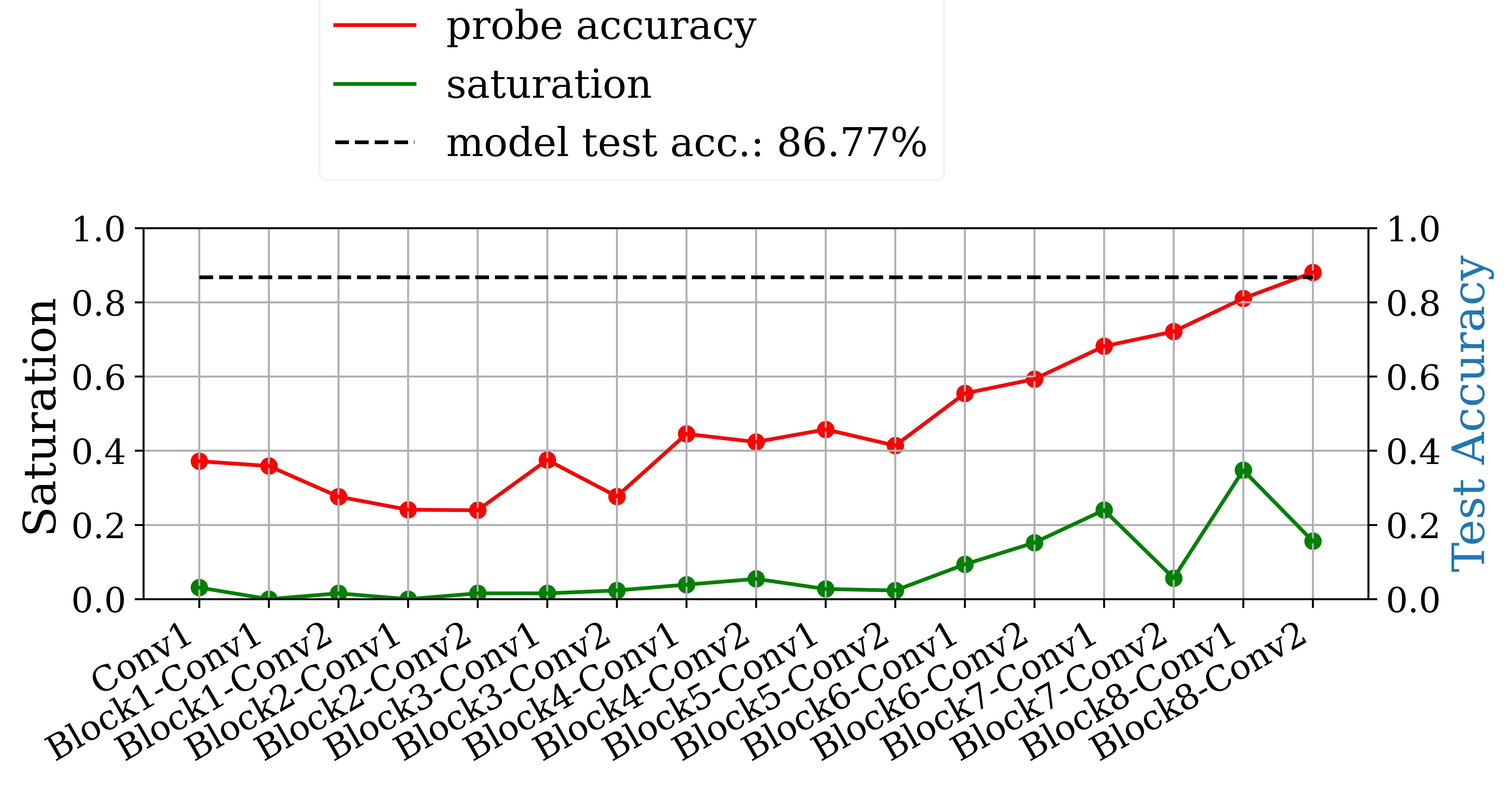}
	    \label{fig:resnet18_cifar10_big}

	    %\label{fig:iNaturalistVgg16_big}
	}
	\caption{Changing the input size changes how the inference is distributed among the layers. 
	The resolution ResNet18 was designed for (b) distributes the inference most evenly, while too small (a) and too large (c) resolution shift the bulk of the inference process to early and later layers respectively, resulting in worse predictive performance.}
	\label{fig:resnet18_cifar10_resolution}

\end{figure}

The results in figure \ref{fig:resnet18_cifar10_resolution} show that the inference process is distributed differently among the layers depending on the input size.
Best predictive performance is achieved at $224 \times 224$, where inference is also distributed most evenly.
At a substantially smaller size of $32 \times 32$ pixels, the performance suffers, dropping from 92.79\% accuracy to 84.64\%.
We also observe in figure \ref{fig:resnet18_cifar10_small} that only the first (highly saturated) two thirds of the network improve the predictive performance of the probes substantially from layer to layer. The rest of the network is lower saturated and does not contribute to improving the quality of the solution.\footnote{The anomalies of single layers dropping drastically in probe performance are an artifact of residual connections. This was first observed by \citet{alain2016} and also described by \citet{saturation}}
Thus shrinking the input size has shifted the bulk of the qualitative inference process to the front of the network.
Similarly, when drastically increasing the resolution to $1024 \times 1024$ pixels (see figure \ref{fig:resnet18_cifar10_big}), the opposite effect can be observed.
The qualitative inference process is shifted closer to the output, while the first half of the network does not substantially improve the predictive performance.
The impact on the predictive performance is also negative, reducing the accuracy from 92.79\% to 86.77\% compared to the model trained on $224 \times 224$ pixel input images. 

\subsection{The role of the object size in the relation of model and input resolution}
\label{sec:objectsize}
Processing similar features with different sizes is a common challenge in the related field of object detection, where objects may strongly vary in size and position on the image \cite{yolo, yolov2, yolov3, yolov4}.
One possible solution is the pyramid network concept introduced in \cite{pyramidnets}, which is also employed in popular models like YoloV3 by \citet{yolov3}.
By utilizing feature maps from different parts of the network the model becomes more stable when the object size varies significantly.
This indicates that processing objects of different size is handled by different parts of the network.
%The problem of varying object sizes is generally less severe for classifiers. Most commonly used classification datasets in computer vision like Cifar10, Cifar100, MNIST, iNaturalist, Food101 and ImageNet feature only a single object which is in most cases occupying a large (and often centered) portion of the image. \footnote{We interpret the word "object" and by extension "object size" very loosely in this work. 
%CNN Classifiers only detect discriminatory patterns indicative of certain classes and are not required to have a concept of object or objectness. 
%Therefore "size" of an "object" in this sense refers to how local discriminatory patterns are on an image.}
Based on our previous observations, we hypothesize that the uneven distribution of the inference is caused by the size of features on the images used for classification.

\begin{figure}[htb!]
	\centering
	\subfloat[$32 \times 32$ without canvas]{
	    \includegraphics[width=0.8\columnwidth]{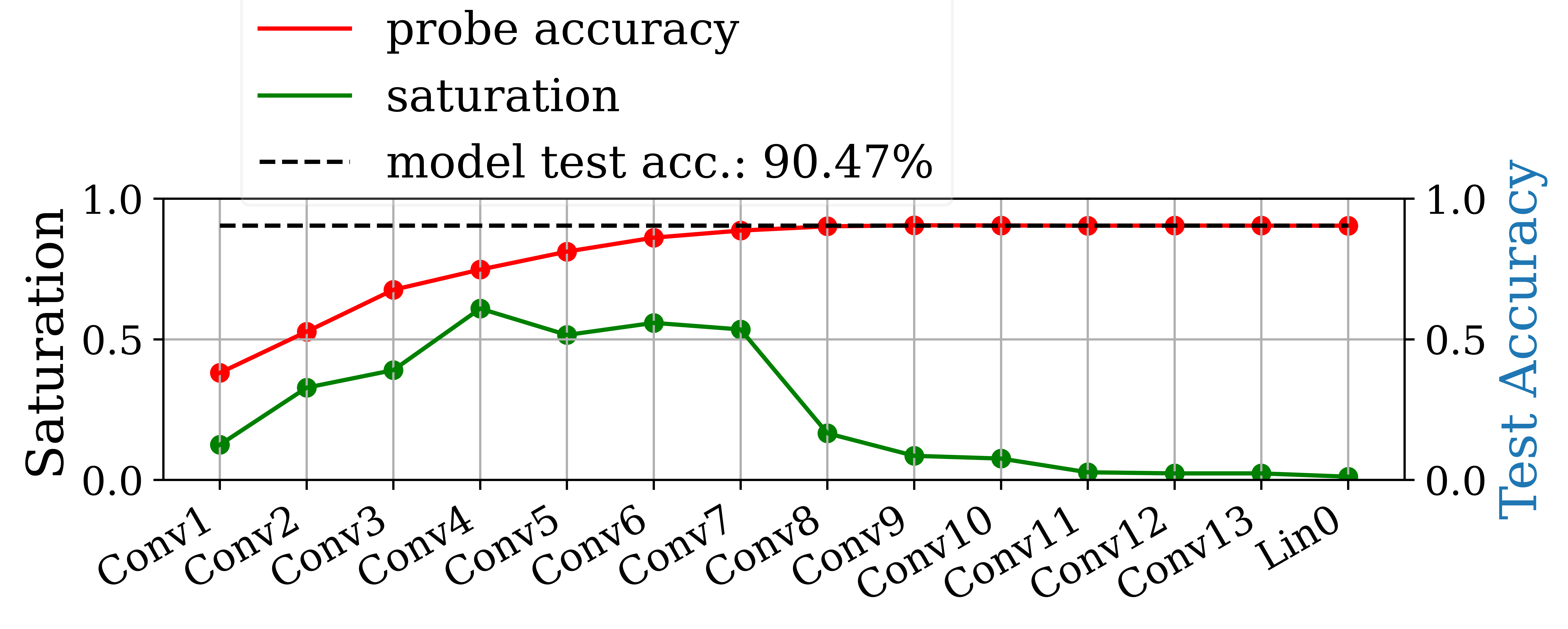}
	    %\label{fig:iNaturalistVgg16_resized}
	}
	\quad
	\subfloat[$160 \times 160$ with canvas]{
	    \includegraphics[width=0.8\columnwidth]{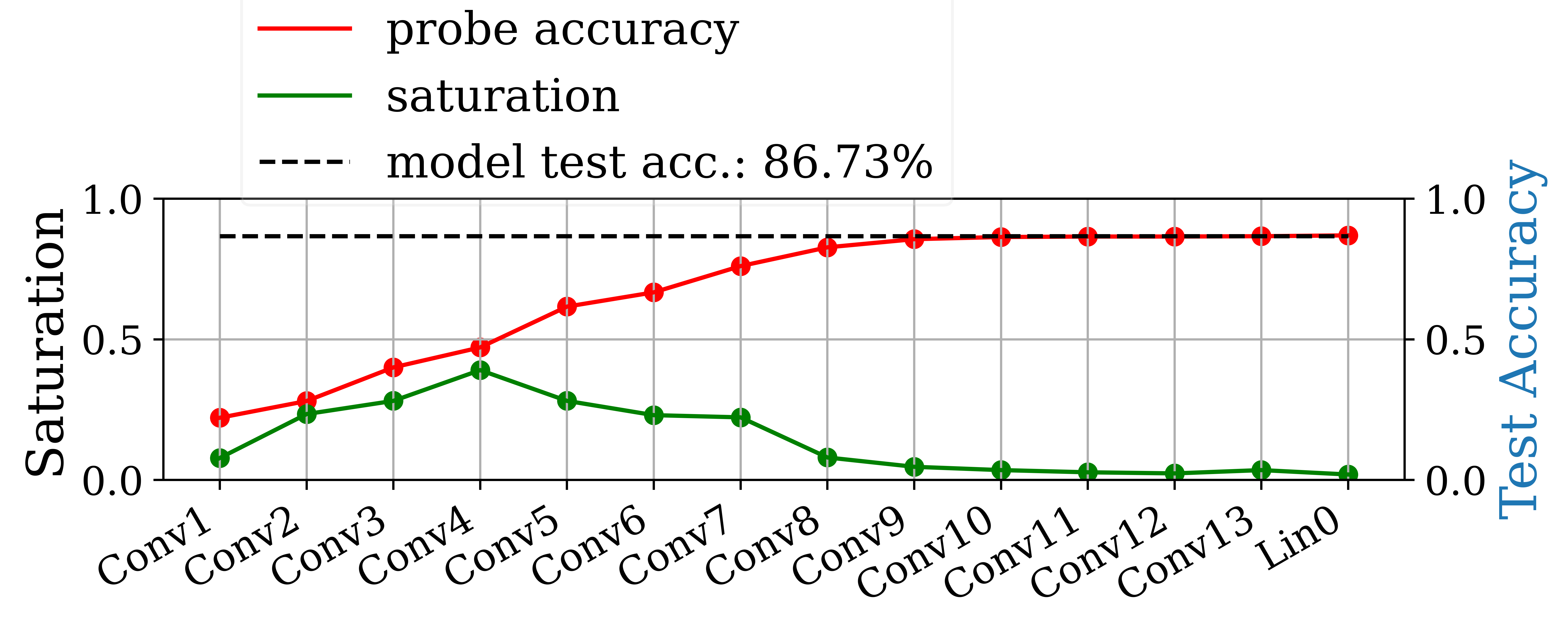}
	    %\label{fig:iNaturalistVgg16_big}
	}\quad
	\subfloat[$160 \times 160$ upsampled]{
	    \includegraphics[width=0.8\columnwidth]{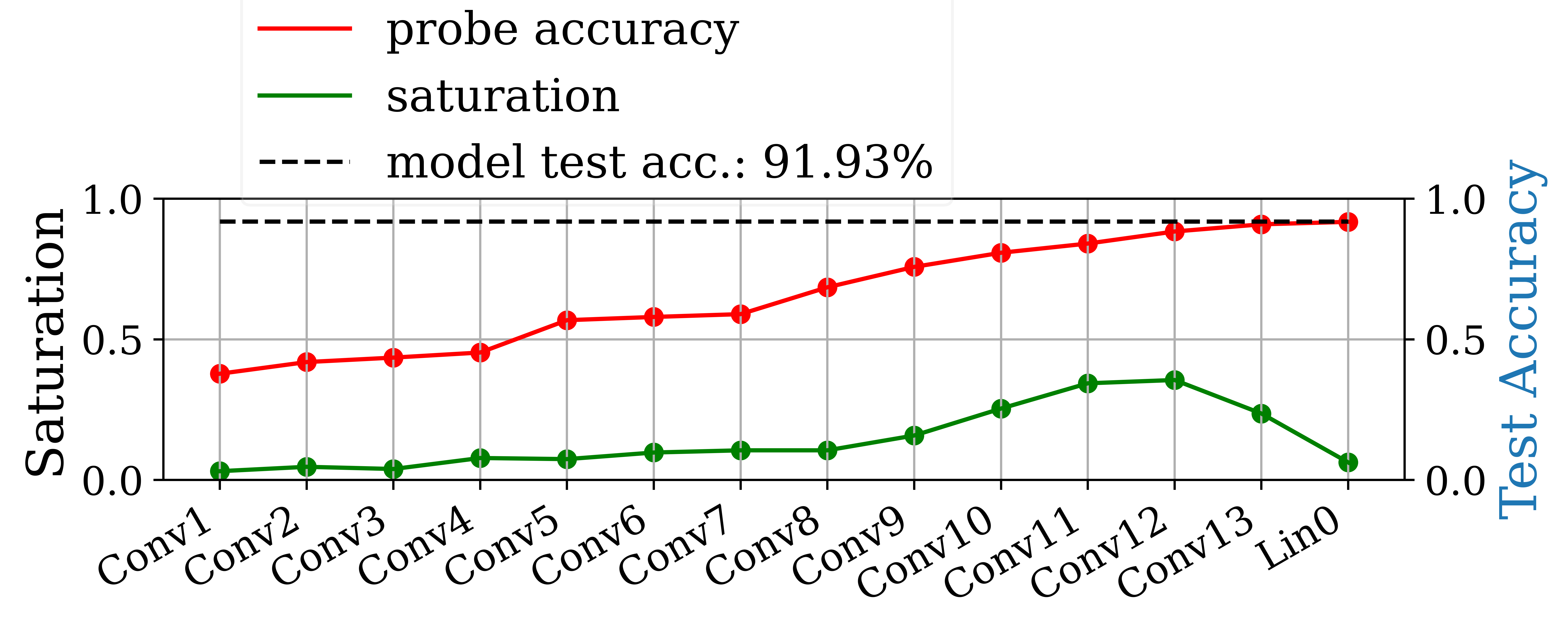}
	    %\label{fig:iNaturalistVgg16_big}
	    \label{fig:object_size_upsampled}

	}
	\caption{A tail of unproductive layers can be produced by placing the Cifar10 images on a $160 \times 160$ pixel canvas. This indicates that locality of discriminatory features (essentially the size of the object) is responsible for the observed effect on the inference process.}
	\label{fig:object_size}

\end{figure}

To verify this experimentally we train models on a modified version of the Cifar10 dataset. In this version the Cifar10 images are randomly placed on a black canvas of $160 \times 160$ pixels.
This restricts the size of the object of interest and therefore all potentially discriminatory features to a square region of 32 pixels per side.
If the absolute size of the object is influencing which parts of the network process the information, the resulting tail pattern and evolution of the probe performances should be similar to the same model trained on Cifar10 on its native resolution.
The results in figure \ref{fig:object_size} show that this is the case.
Saturation and probe performances behave very similar to training the model on $32 \times 32$ pixels, while scaling the entire image to $160 \times 160$ pixels distributes the inference process very differently (see figure \ref{fig:object_size_upsampled}).
The same effect can be reproduced with other models and datasets (see supplementary material).

We can conclude from this that the relevance of the input size when training classifiers comes from the size of the discriminatory features detected by the model.
In other words, the size of the image is a heuristic for the size of the object of interest.
This also means that the results depicted in this work are unlikely to generalize to tasks where objects of interest are more localized and varied like in most object detection datasets.

%While this experiment shows that the observed relationship of the input size with the neural architecture is merely an artifact of the relation of the detected patterns and the neural architecture, we will proceed to use the input size as heuristic measure for this property.
%There are multiple underlying reasons for this decision.
%First, a convolutional neural network minimizing cross entropy is not bound to any definition of object or "objectness" and may therefore recognize any pattern. 
%This is also the case when the recognized patterns do not technically belong to an object by any definition of human perception as demonstrated by \cite{rethinking_generalization}. 
%This makes any rigorous, formal definition of object size impossible.
%Second, as we mentioned before, many classification datasets are relatively biased towards more homogeneous object sizes (compared to object detection datasets). \cite{food101, affectnet, cifar, mnist, fashionmnist, iNaturalist}
%While the object is in most cases still smaller than the image and may vary in size from class to class, we can still use the resolution of the image as an upper-bound heuristic for the size of the object.
%We are aware that this heuristic is neither flawless nor perfect. 
%However, for our purposes this technique is sufficient to put the observed differences in context.

\subsection{The role of the receptive field in relation to the object size}
\label{sec:receptivefield}
From the previous section we can derive that features of different sizes are recognized by different layers in the model.
In this section we will investigate the causes of this phenomenon from an architectural point of view.

From the perspective of the architecture, the receptive field of a layer can be considered an upper bound for the size of recognizable features.
Since the receptive field expands with every layer that has stride and/or kernel size greater than 1, increasingly larger features can be recognized.
We hypothesize that for simple, sequential architectures like the VGG-family of networks \footnote{We define simple architectures as networks that are strictly sequential without multiple pathways, featuring only convolutional layers (with ReLU and BatchNorm) and downsampling (max-pooling or otherwise) layers in the feature extractor. The feature extractor is separated by a global average pooling layer and followed only by fully connected layers.}, the receptive field is the dominating factor influencing presence and position of unproductive subsequences of layers.

\begin{figure}[htb!]
	\centering
	\subfloat[VGG11]{
	    \includegraphics[width=0.8\columnwidth]{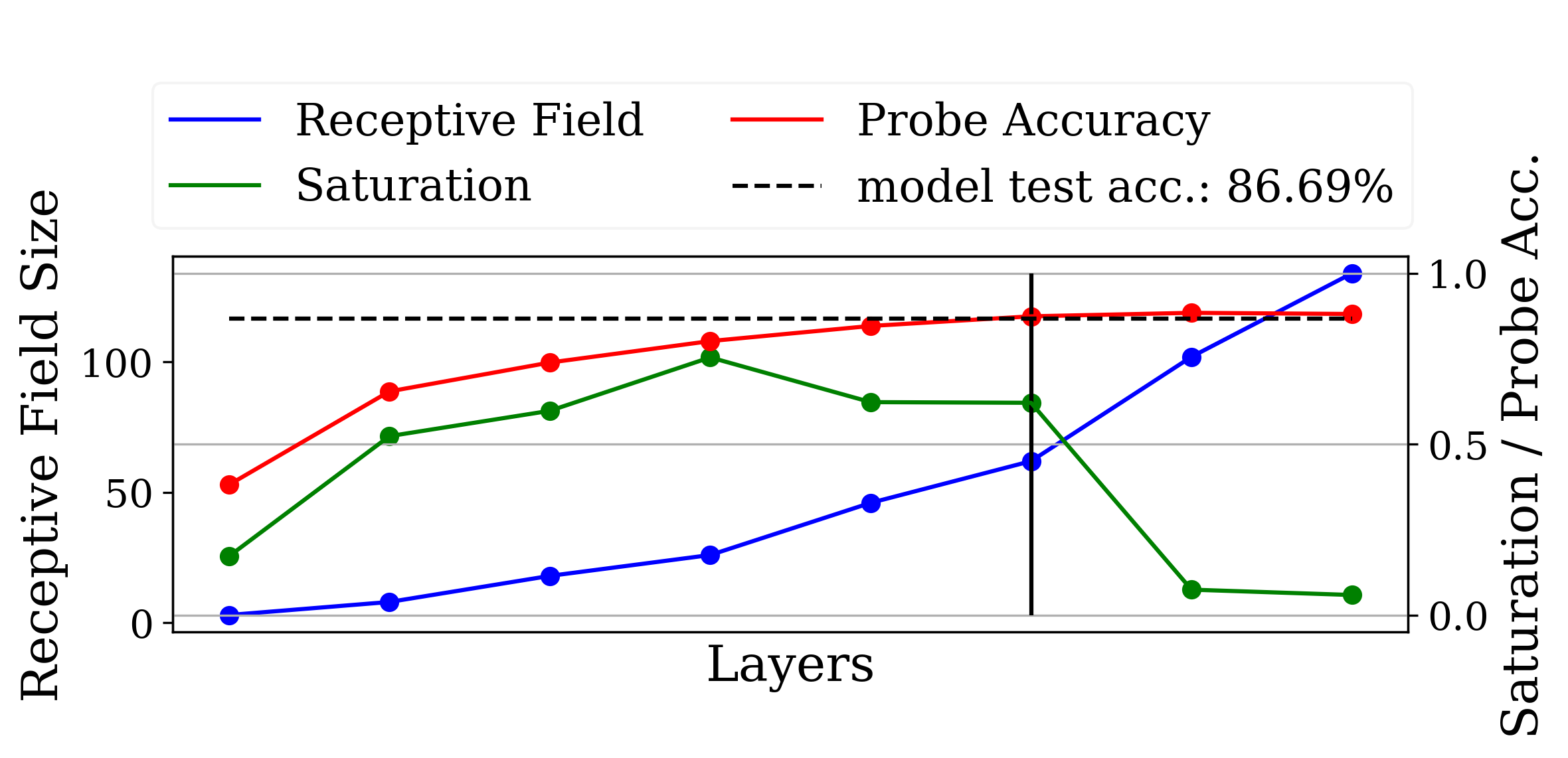}
	}\quad
		\subfloat[VGG13]{
	    \includegraphics[width=0.8\columnwidth]{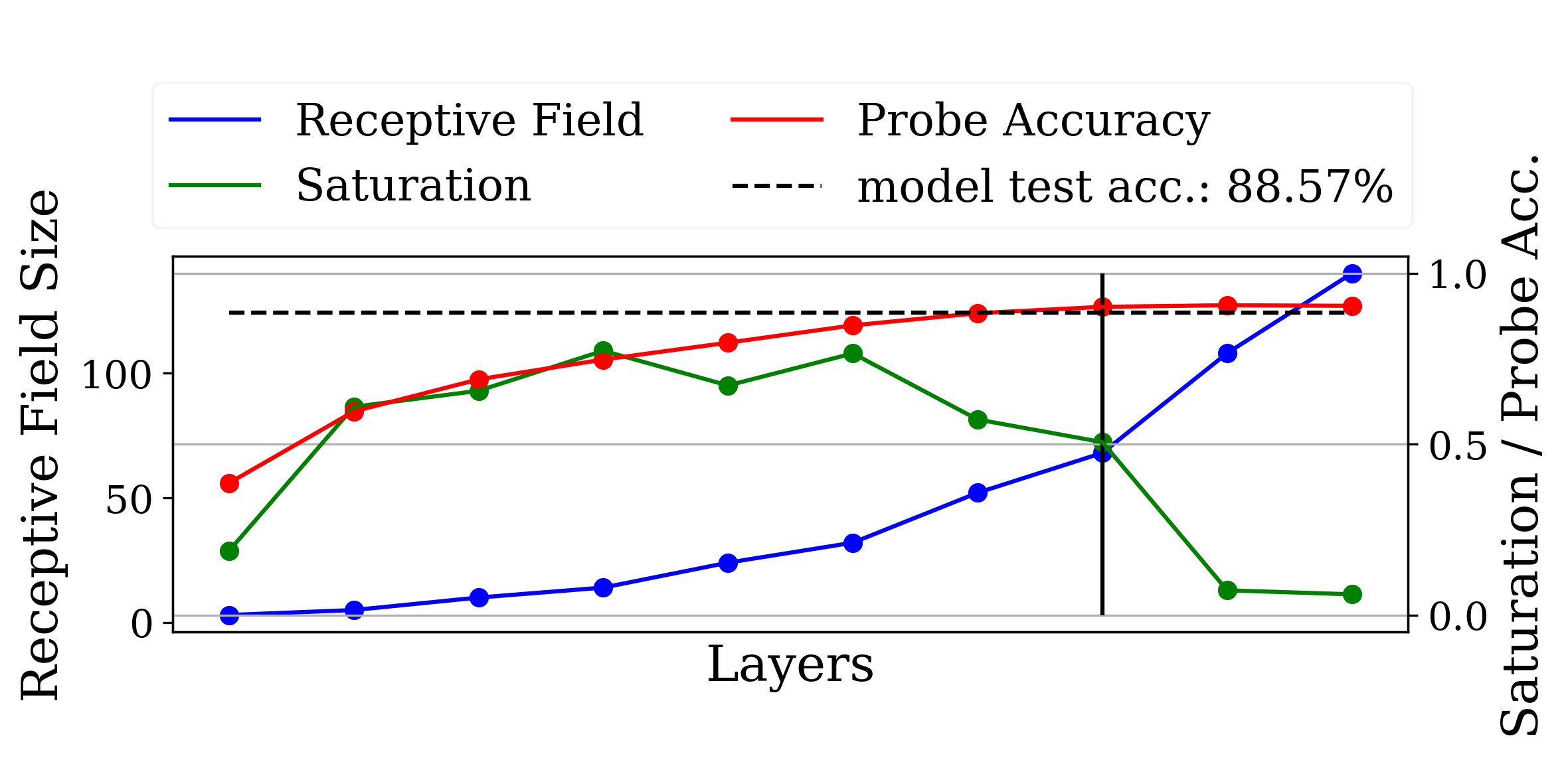}
	}\quad
	\subfloat[VGG16]{
	    \includegraphics[width=0.8\columnwidth]{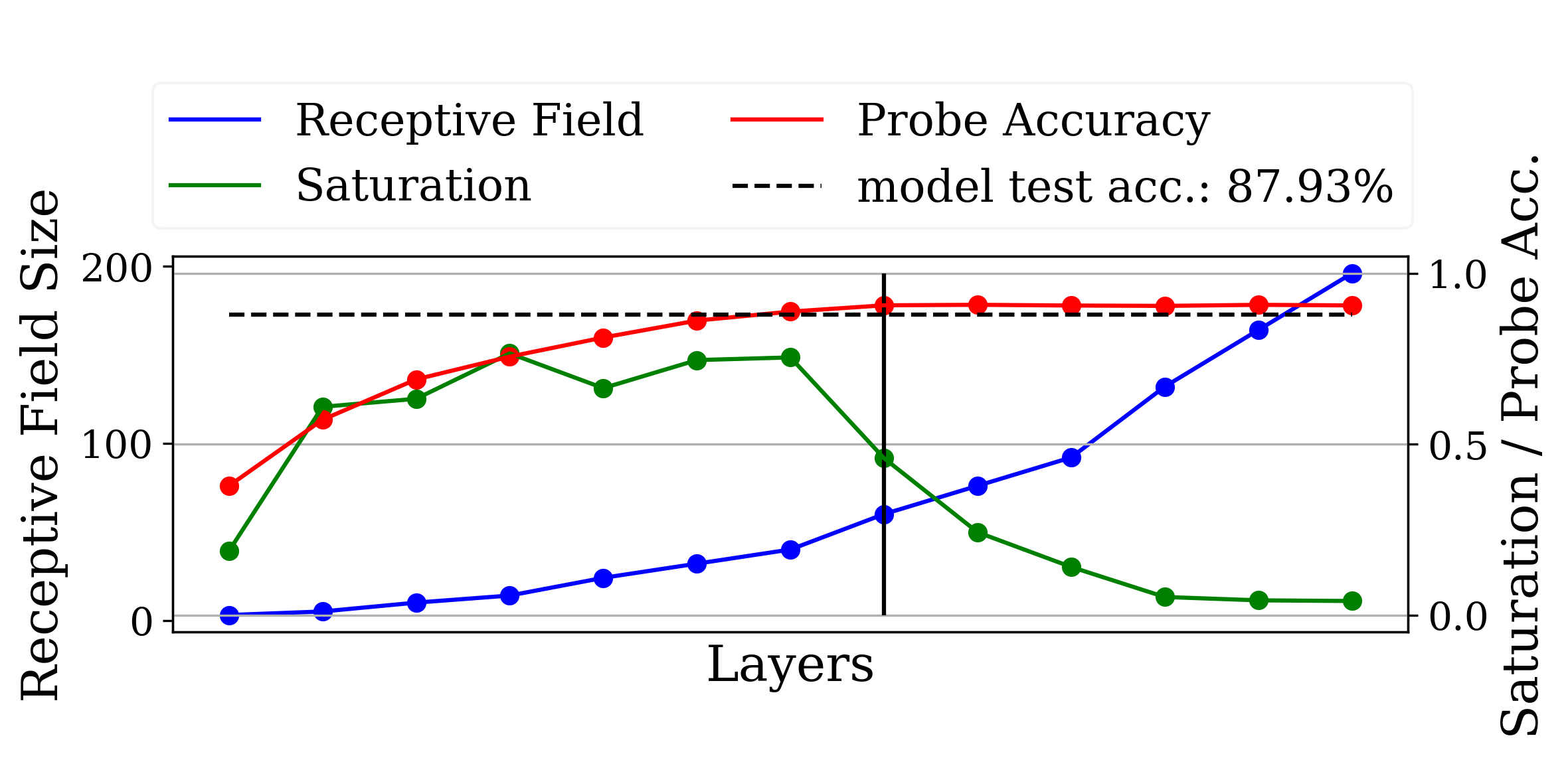}
	}\quad
	\subfloat[VGG19]{
	    \includegraphics[width=0.8\columnwidth]{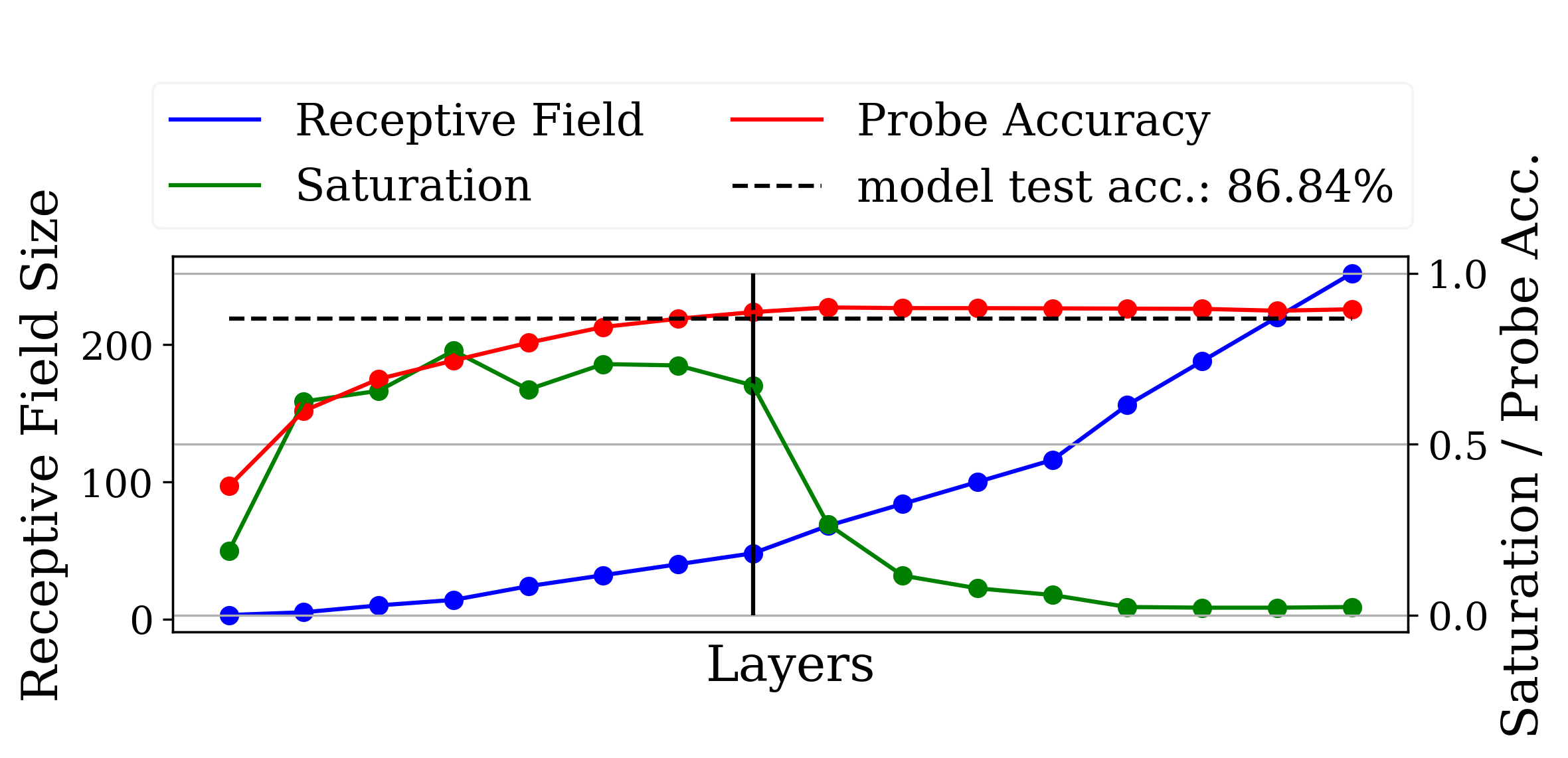}
	}
	\caption{The first convolutional layer processing input from a layer with a receptive field size greater than the input size can be seen as a border (marked with a black vertical line) between the "solving" and the "compressing" stage of the network. Only the highly saturated solving stage improves on the intermediate solutions as shown by the probes. The depicted models are from the VGG-family of networks by \citet{vgg} and trained on Cifar10 with an input size of $32 \times 32$ pixels.}
	\label{fig:vgg_receptive_field}

\end{figure}

We study this property of the receptive field by training the VGG-family of networks on Cifar10 and add the receptive field as an additional information to our analysis. 

\begin{figure}[htb!]
	\centering
	\subfloat[ResNet18 (no res. connections)]{
	    \includegraphics[width=0.8\columnwidth]{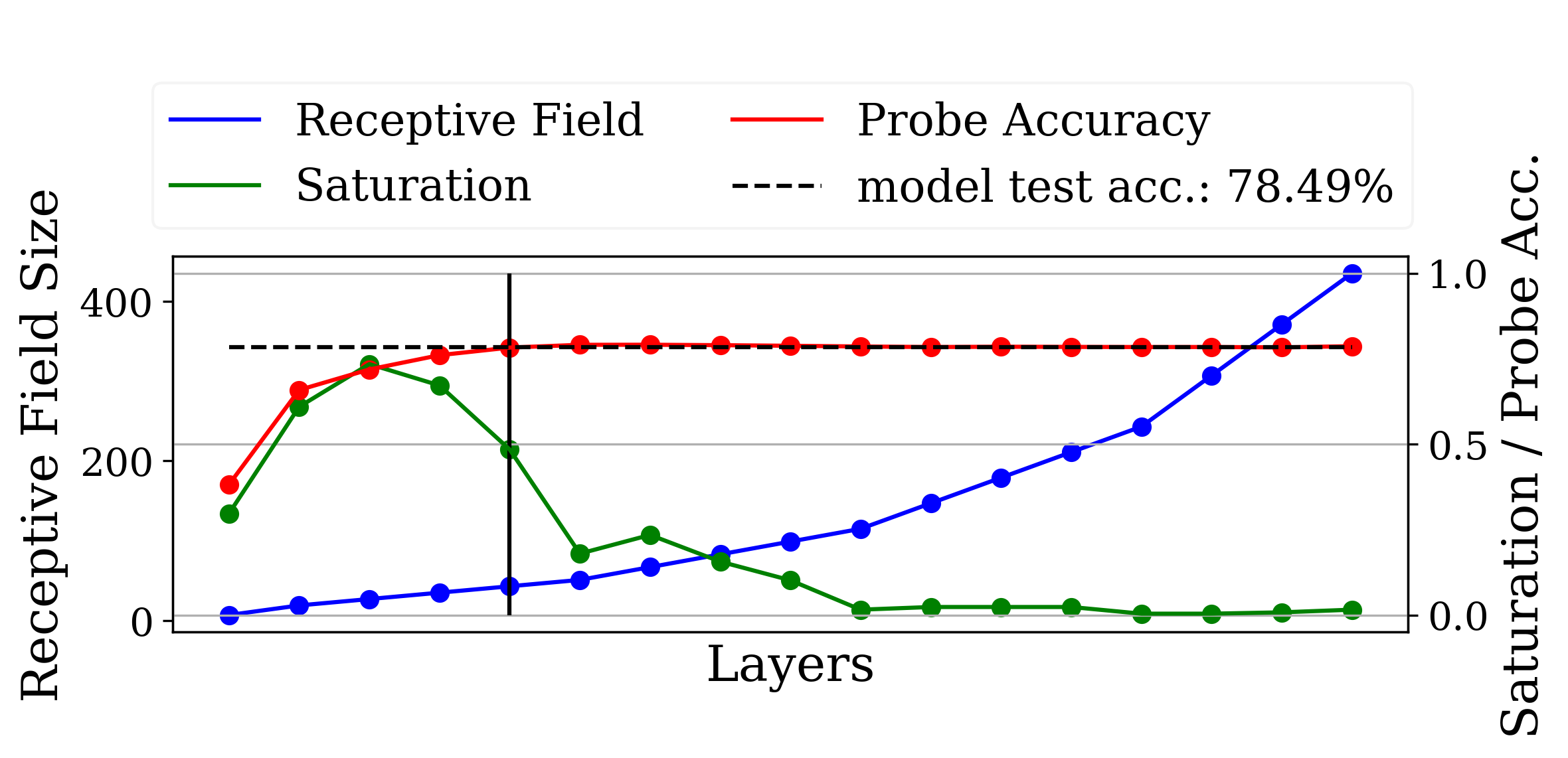}
	} \quad
	\subfloat[VGG19 (dilation=3)]{
	    \includegraphics[width=0.8\columnwidth]{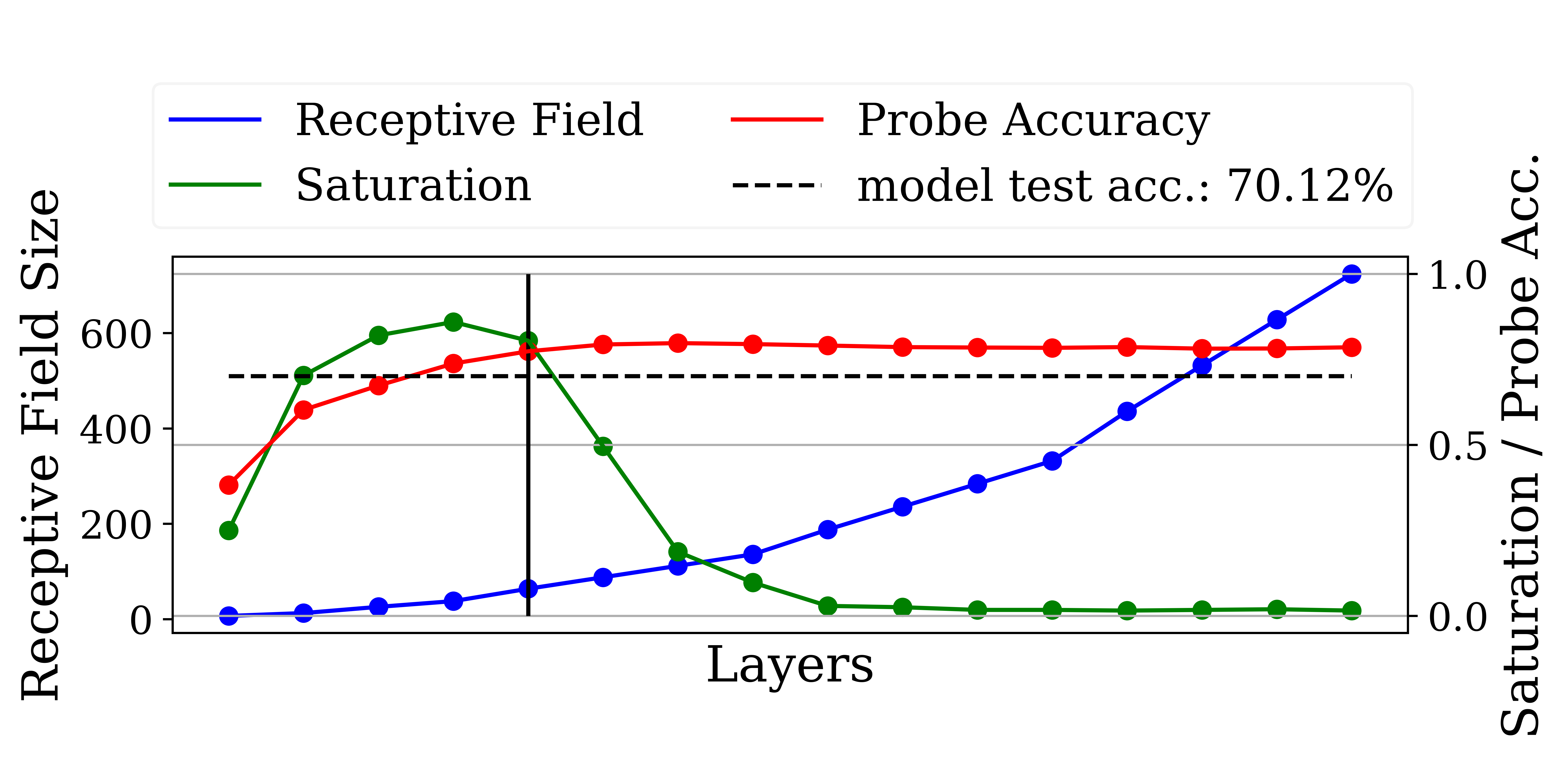}
	}
	\caption{The behavior described in figure \ref{fig:vgg_receptive_field} translates to different sequential architectures as well. 
	For instance ResNet18 with disabled residual connections uses mainly strided convolutional layers and has a significantly faster receptive field expansion due to the first two layers being downsampling layers. 
	The VGG19 variant displayed here has the border layer shifted closer to the input compared to the original version displayed in figure \ref{fig:vgg_receptive_field}, since the dilated convolutions increase the kernel size and therefore the expansion of the receptive field. All models are trained on Cifar10 with $32 \times 32$ pixel input size.}
	\label{fig:vgg16resnetnoskip}
\end{figure}

In figure \ref{fig:vgg_receptive_field} we mark the first layer that processes input from a layer with a receptive field size greater than the input size with a black border. 
We will refer to this layer as the border layer.
For all architectures, this border separates layers contributing to the inference process from layers that do not contribute to the quality of the inference significantly. 
This suggests that a layer in a simple, sequential architecture can only substantially improve the performance when novel information is integrated into positions on its feature map.

We investigate this observation further by testing architectures, which alter the receptive field size in different ways than the VGG-family of networks.
We test the effect of dilated convolutions by training a modified VGG19 with a dilation rate of 2 in all convolutional layers.
This modification effectively increases the kernel sizes of all layers, which in turn increases the receptive field size without changing the number of parameters.
We also turn ResNet18 into a sequential model by removing the residual connections.
The ResNet-family mainly uses strided convolutions for downsampling and also features more downsampling layers that are spaced differently compared to VGG-style models.
This results in a much faster growth of the receptive field as well as a higher overall receptive field size.

From the results in figure \ref{fig:vgg16resnetnoskip} we can see that the behavior of both models is consistent with previous observations.
%Figure \ref{fig:vgg16resnetnoskip} shows two additional examples of this behavior (for more see supplementary materials). The left model is ResNet18 with removed residual connections and the right model is VGG19 using dilated convolutions with a dilation rate of two. 
%Both networks are trained on Cifar10.
%In both cases the border layer separates the productive and unproductive part, consistent with previous observations.

\begin{figure}[htb!]
	\centering
	\subfloat[Pre Border]{
	    \includegraphics[width=0.225\columnwidth]{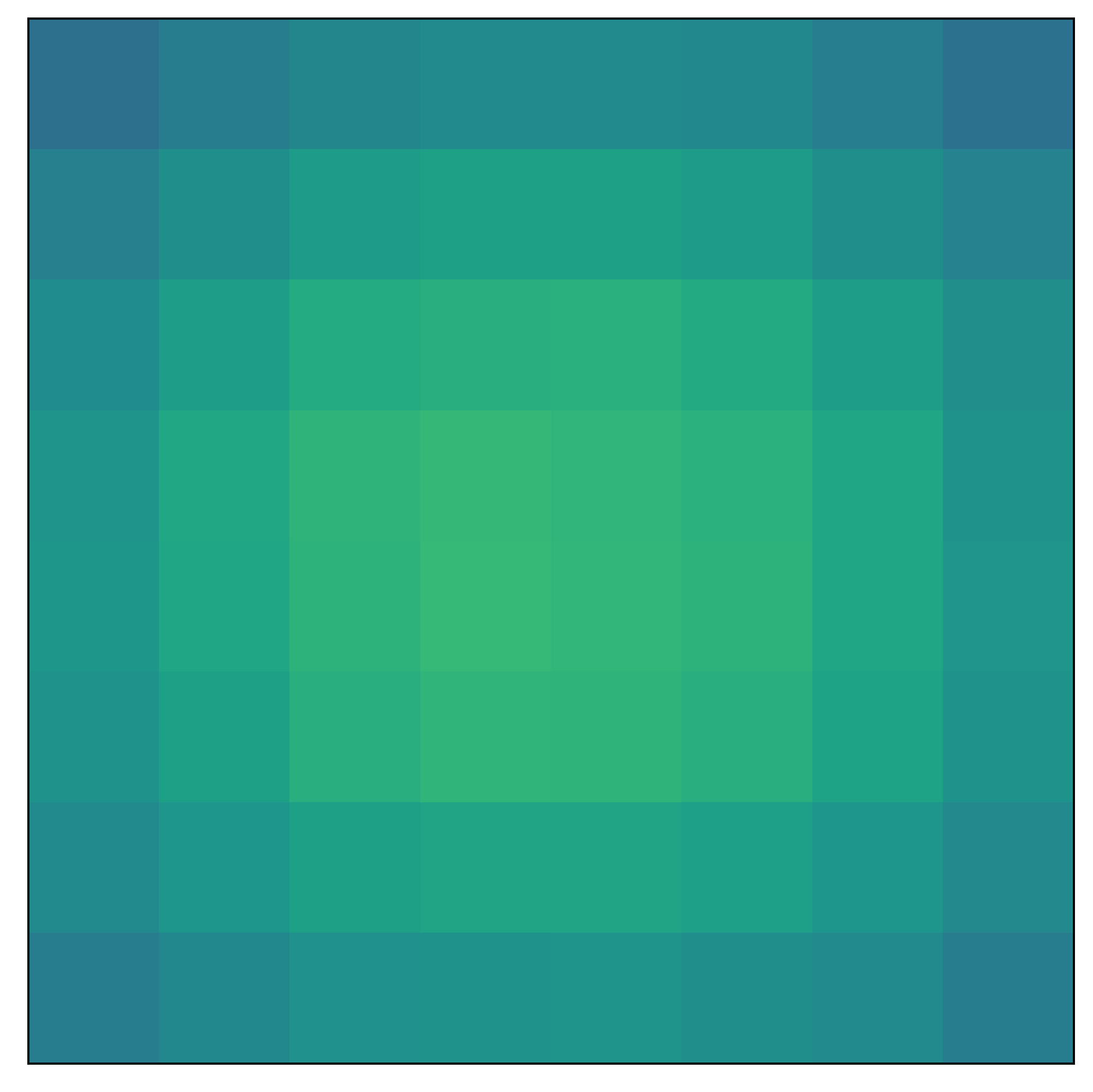}
	}
		\subfloat[Border]{
	    \includegraphics[width=0.225\columnwidth]{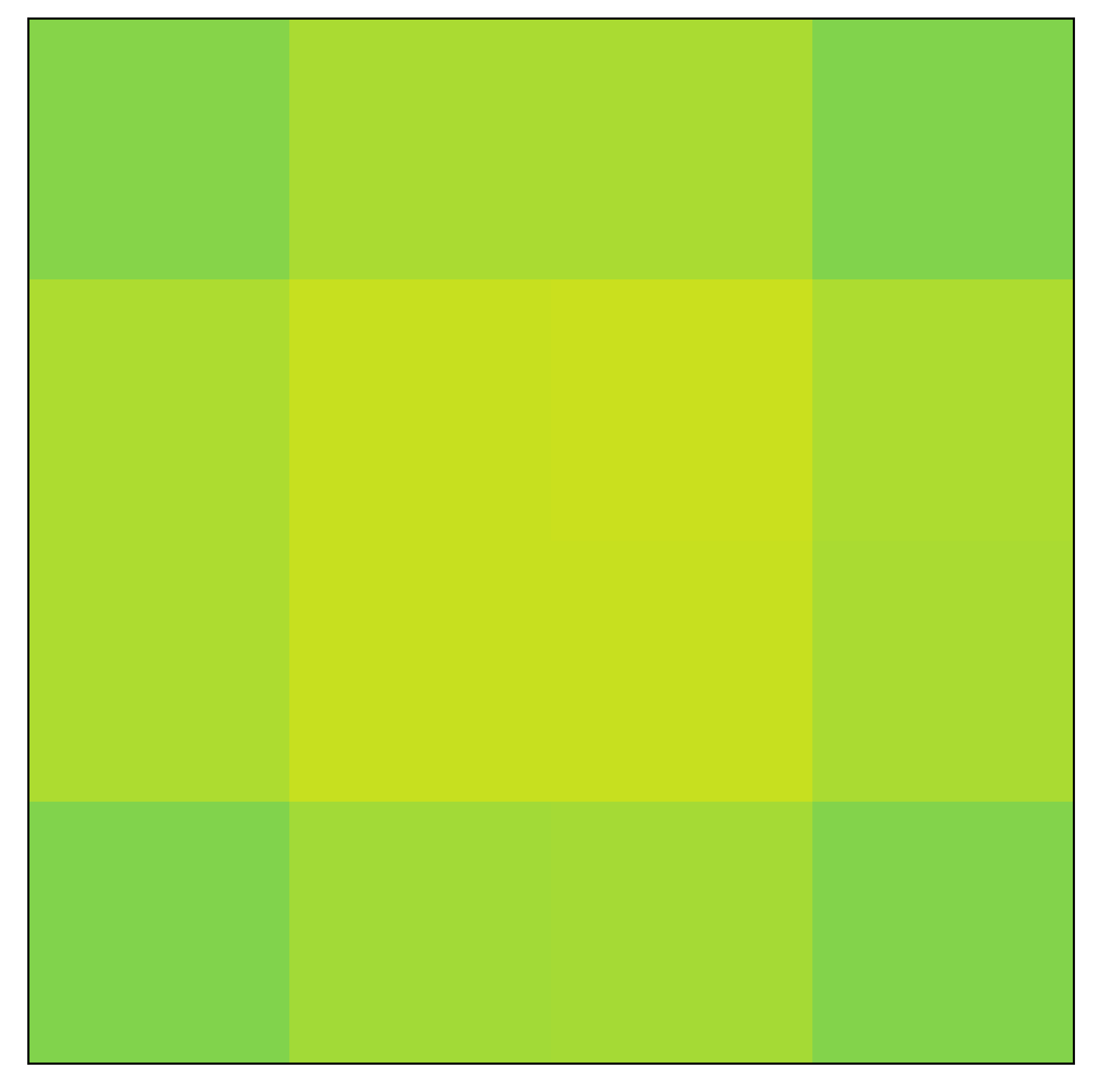}
	}
	\subfloat[Post Border]{
	    \includegraphics[width=0.225\columnwidth]{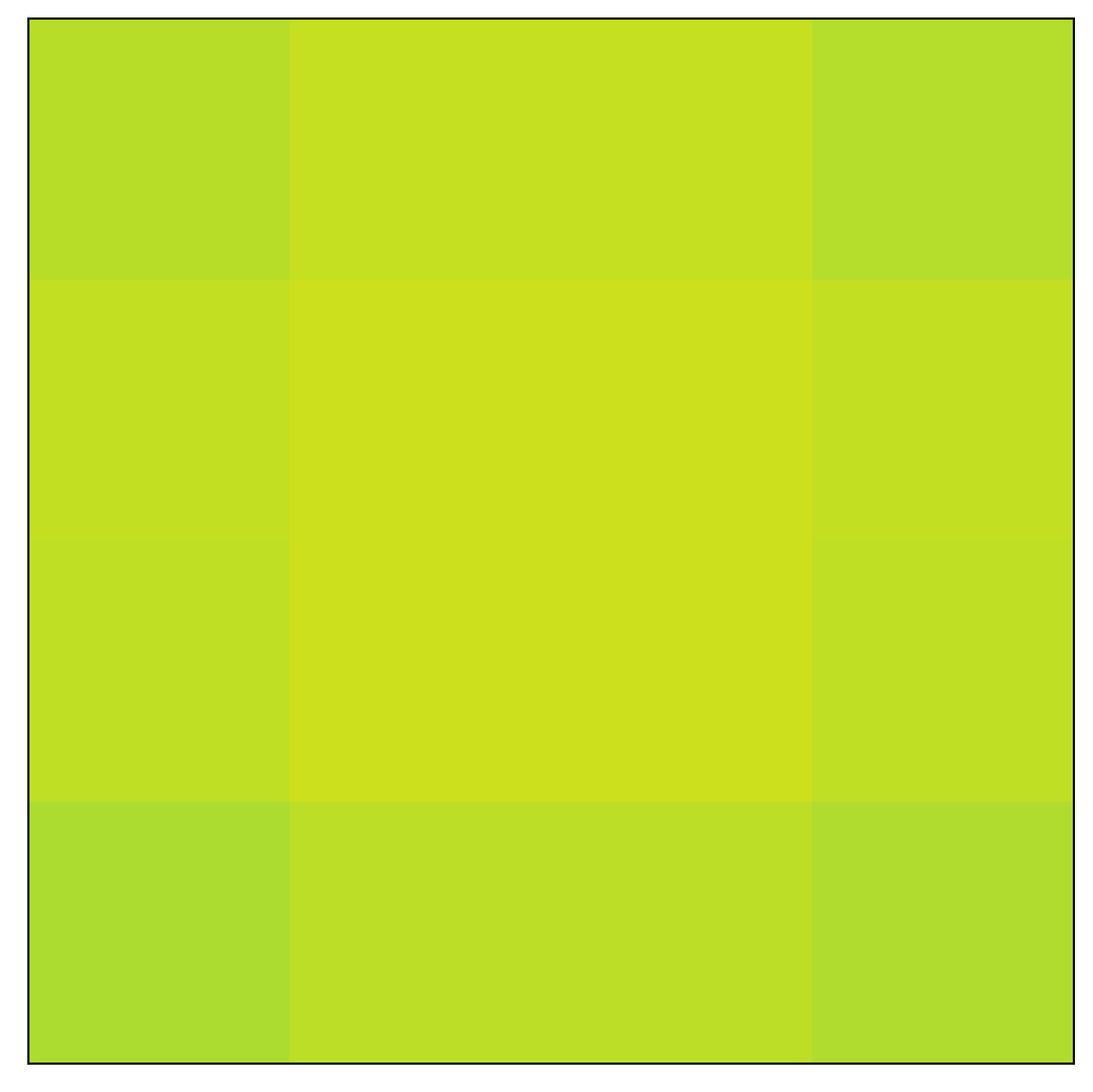}
	}
	\subfloat[Last Layer]{
	    \includegraphics[width=0.225\columnwidth]{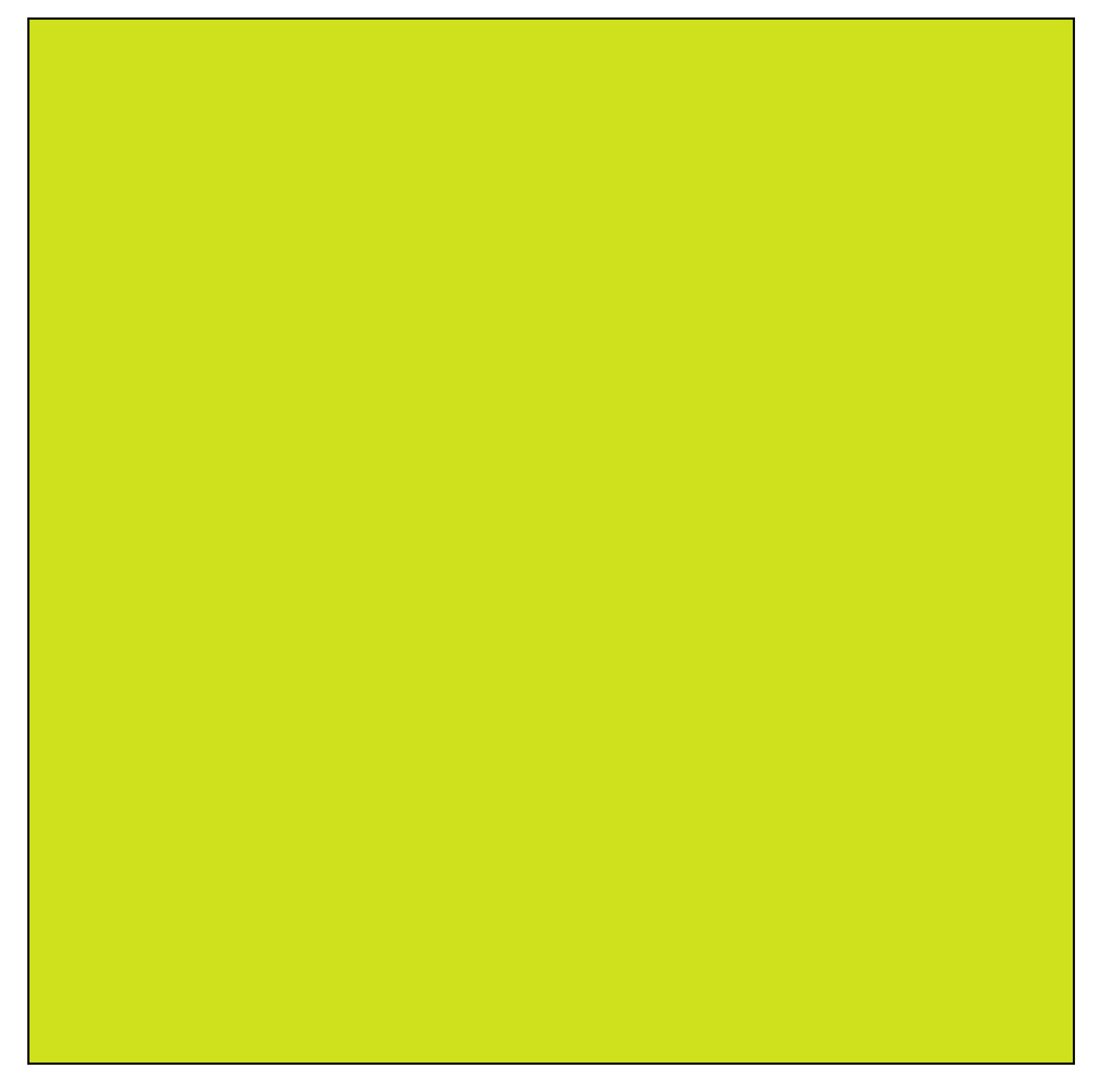}
	} \quad
	\subfloat{
	    \includegraphics[width=0.8\columnwidth]{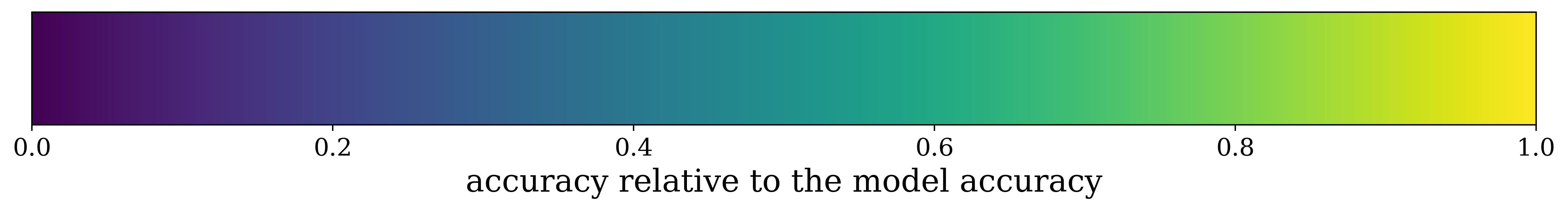}
	}
	\caption{The heatmaps display the relative performance of probes trained on individual positions of feature map to the performance of the model. The depicted model is VGG16 trained on Cifar10 and displayed also in figure \ref{fig:vgg_receptive_field}. The depicted layers are the 5th, 8th, 9th and the last convolutional layer.
	We observe that the performance of probes on early layers is best in the center, where the receptive field can integrate the most information since the receptive field will not exceed the bounds of the image.
	This is still true for the border layer. However, all four inner positions of this feature map integrate information from the entire image and have reached the performance of the softmax output of the model as a result. The last two depicted layers visualize the activity in the low saturated tail. 
	As the receptive field expands more positions on the feature map encode information from the entire image and achieve similar probe performances. This homogenisation of partial solutions results in a decreasing number of high variance eigendirections which in turn results in low saturation. This explains the low saturated tail of layers seen on networks trained on too low resolution.}
	\label{fig:vgg_probe_heatmaps}
\end{figure}

Finally, we investigate how the solution develops inside the feature maps of different parts of the network.
We modify our method of logistic regression probes by applying a single probe to every position of every layer's feature map.
We then compute the relative performance of the probes by dividing the probe accuracy with the network`s accuracy (both evaluated on the test set).
By doing so we can visualize the quality of the partial solutions contained in every position of the feature map based on their position.
We observe in figure \ref{fig:vgg_probe_heatmaps} that the central positions on the feature map generally perform best in early layers, while outer positions perform increasingly worse, with the corner positions generally being the worst.
We suspect that the receptive field is at least partially responsible for this, since outer positions on the feature map will receive more black padding and thus less information with the receptive field expansion as a center pixel.

Another interesting observation in figure \ref{fig:vgg_probe_heatmaps} is that the center most positions on the feature map contain partial solutions roughly equivalent to the performance of the entire model.
As the saturation drops and layers become part of the low saturated tail, the partial solution quality becomes increasingly homogeneous across feature map positions. In the last layer each vector on the feature map is as linearly separable with regards to the classification task as the average pooled solution.
We conclude based on these measurements that this homogenisation of partial solution quality is also responsible for the drop in saturation.

Based on these observations we can conclude that simple, sequential neural networks may develop two stages of inference when the image is smaller than the receptive field size of the model: The first being the solving stafe, where the data is processed incrementally to achieve loss minimization. The second stage, starting from the border layer, is the compressing stage. This stage compresses the latent space by homogenisation of the partial solutions for every position in the feature map.

\subsection{The role of residual connections}
\label{sec:residual}
Residual connections are a popular component used in many neural architectures \cite{resnet, densenet, inceptionv3, efficentnet}.
According to \citet{resnet} the networks with residual connections are able to add "deltas" to the existing representation of the data rather than transforming it entirely.
The residual connection itself does not expand or change the receptive field. 
After the residual connection is added to the output of a convolutional layer, information based on multiple receptive field sizes is present in the feature map.
This allows features based on lower receptive field sizes to "skip" layers and to be processed later in the network, resulting in the network being able to distribute the inference on more layers.

The aforementioned claims about the effects of residual connections suggest, that models utilizing residual connections are able to utilize layers for improving the prediction after the border layer.
The border layer is the first layer to receive an input produced by a layer with a receptive field size larger than the image.
In sequential architectures it is also the layer separating the solving and compressing stage of the model.

\begin{figure}[htb!]
	\centering
		\subfloat[ResNet18]{
	    \includegraphics[width=0.8\columnwidth]{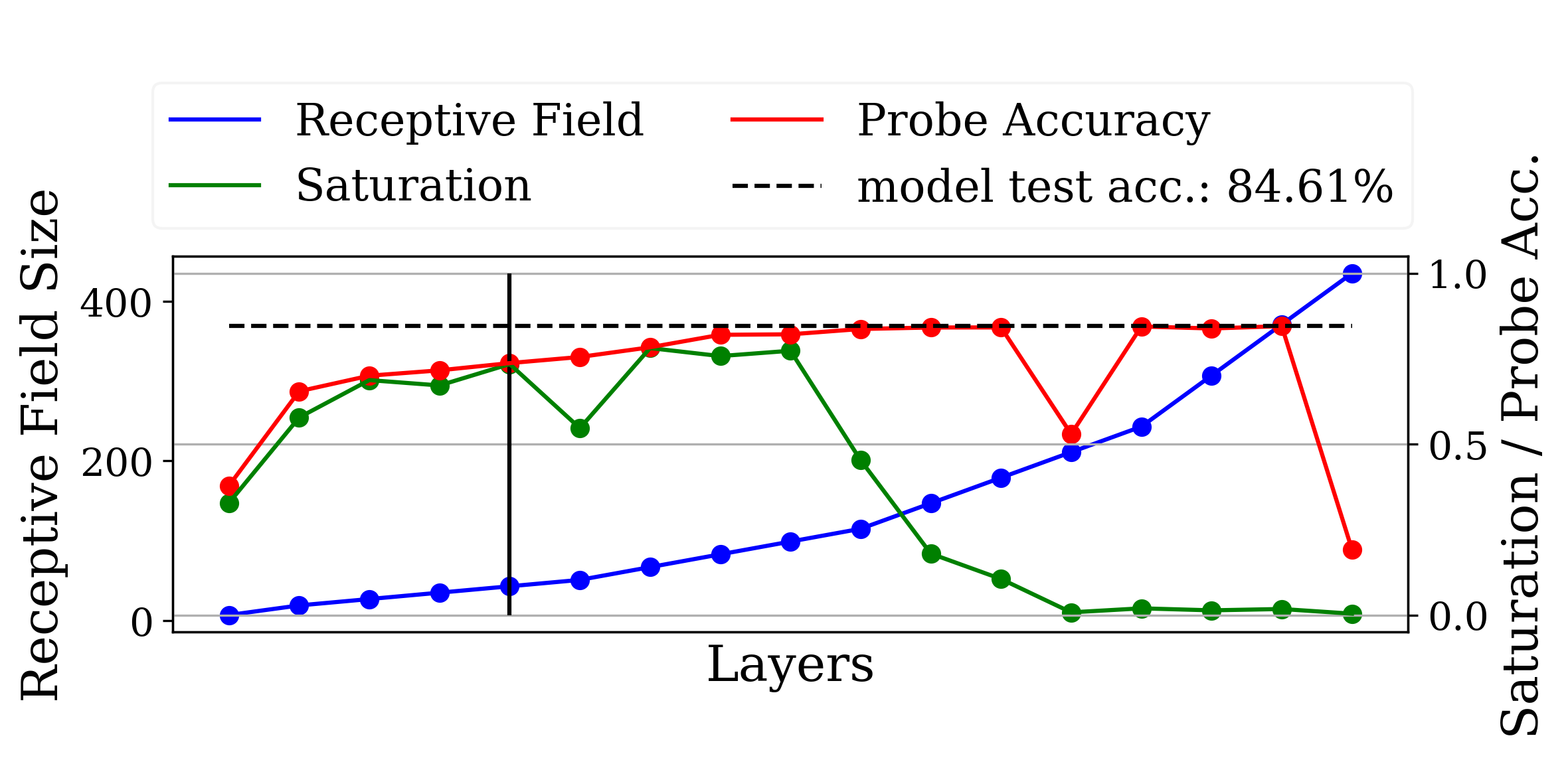}
	} \quad
	\subfloat[ResNet34]{
	    \includegraphics[width=0.8\columnwidth]{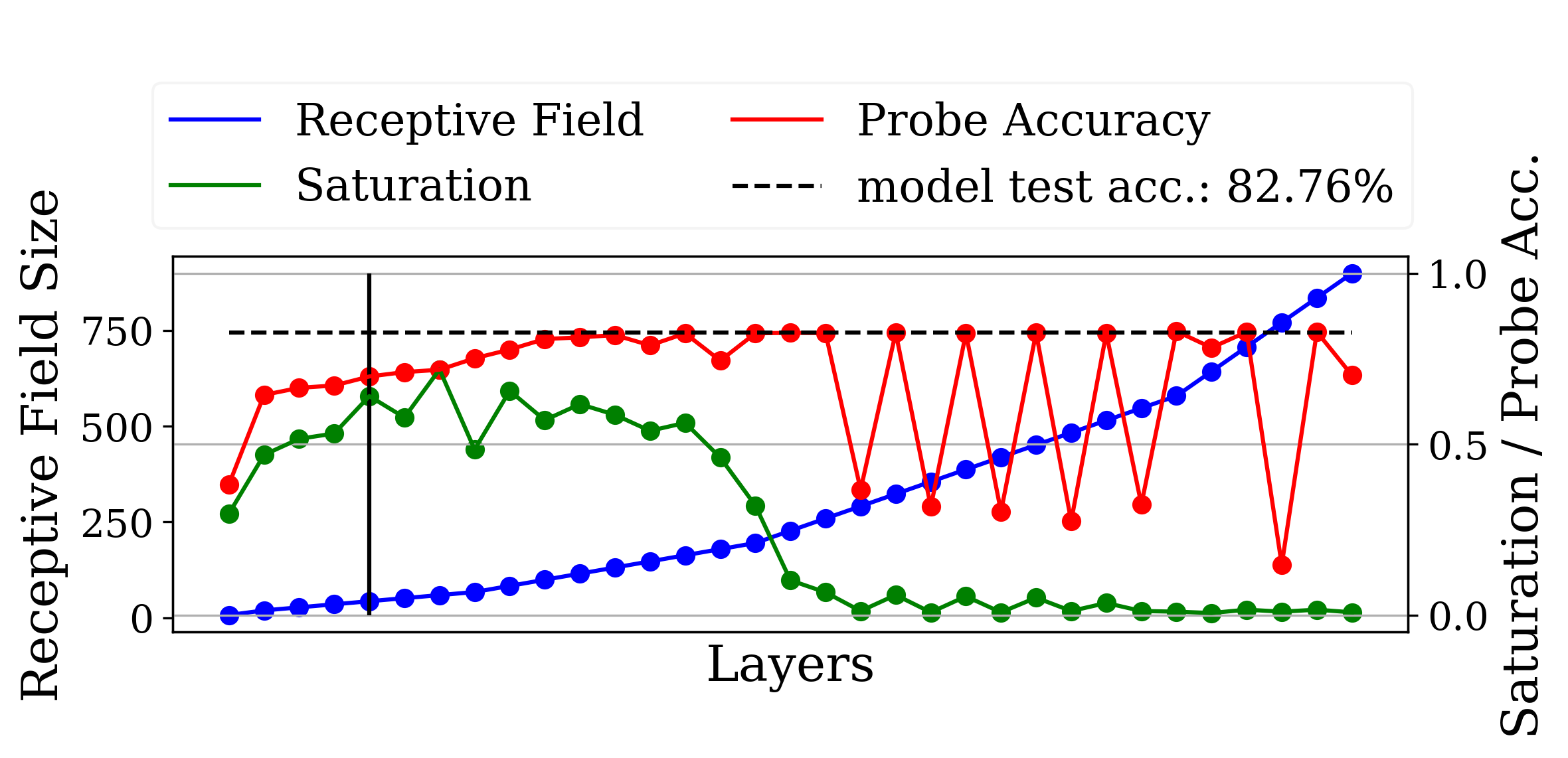}
	}
	\caption{Residual connections allow networks to utilize layers past the border layer (marked with the horizontal bar).
	The zick-zack-pattern and drops of probe performances observed in both networks are artifacts previously observed by \citet{saturation} and \citet{alain2016}.
	These indicate that the networks attempt to "skip" the convolutional layers in the low saturated tail.
	It is also worth noting that these skips are only present on later layers which are part of the tail.}
	\label{fig:resnet_receptive field}

\end{figure}

We can confirm this by looking at the results in figure \ref{fig:resnet_receptive field}.
Both networks improve the probe performances and stay highly saturated long after the border layer has processed the data.
%The receptive field size of ResNet-style networks expands faster compared to VGG-style architectures due to two consecutive downsampling layers close to the input.
%This results in the receptive field size exceed the input size of Cifar10 very early.
We can attribute this behavior to the residual connections, since we tested ResNet18 without them in the previous section (see figure \ref{fig:vgg16resnetnoskip}), resulting in behavior consistent with other sequential architectures (and lower performance).

While the presence of residual connections has a positive effect on the predictive performance of ResNet18 (84.61\% accuracy with and 79.05\% accuracy without residual connections), the performance still remains worse than VGG-style models.
We attribute this to lower overall receptive field size of VGG19, which is only 252 pixels.
ResNet models downsample more aggressively in the beginning, resulting in receptive field sizes of 413 (ResNet18) and over 800 pixels (Resnet34) respectively.
This suggests that matching the receptive field size with the input size remains important when using residual connections in the architecture.

\begin{figure}[htb!]
	\centering
		\subfloat[Cifar10 optimized ResNet18]{
	    \includegraphics[width=0.8\columnwidth]{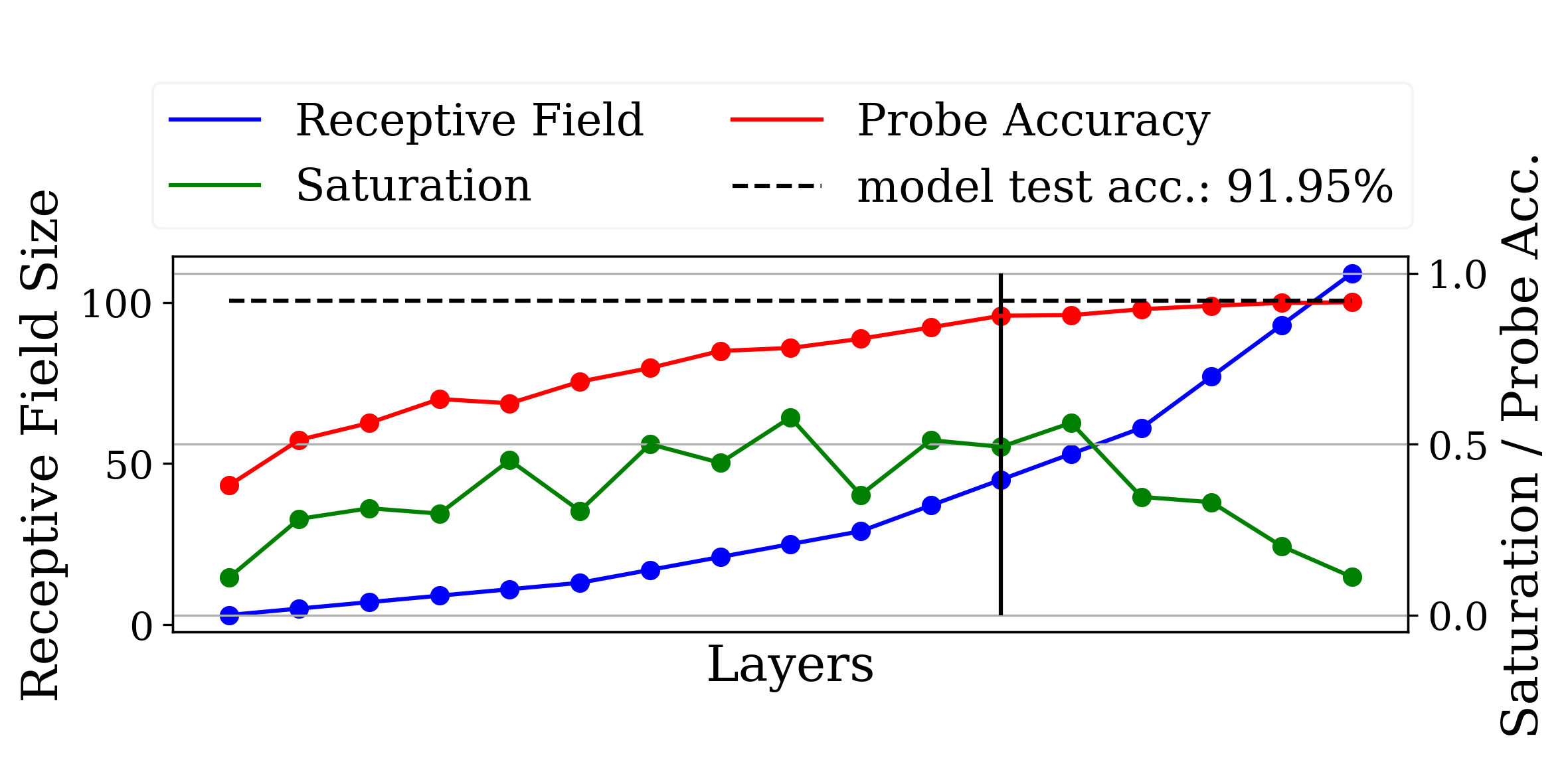}
	} \qquad
	\subfloat[Cifar10 optimized ResNet34]{
	    \includegraphics[width=0.8\columnwidth]{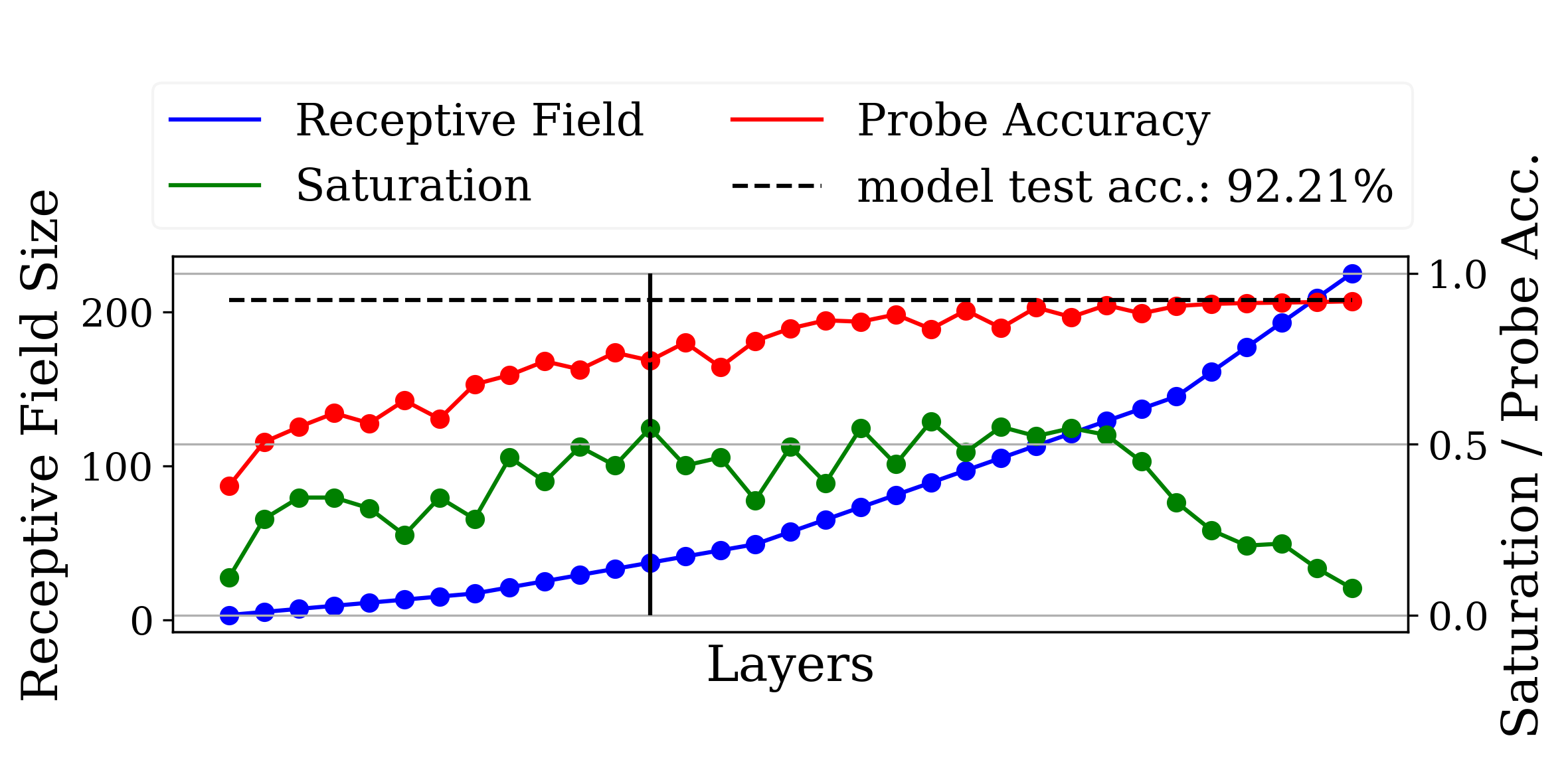}
	}
	\caption{Cifar10 optimized version ResNet18 and 34 have a roughly quartered receptive field size in every layer. These networks still display similar behavior as the ImageNet optimized version when it comes to the border layer. However, the reduction in the receptive field size has further removed the low saturated tail and lead in both cases to a steady increase in performance over the entire networks structure.}
	\label{fig:resnet_cif_receptive field}

\end{figure}

Therefore it should be possible to improve the performance of the (ImageNet optimized) ResNet18 and 34 by reducing the size of the receptive field.
This was implicitly done by \citet{resnet}, who proposed additionally to the ImageNet optimized versions of ResNet a Cifar10 optimized variant.
The Cifar10 variant of ResNet has no max-pooling layer and the first layer has its stride size in kernel size halved.
By doing so the receptive field of ResNet18 for example was reduced from 413 (ImageNet optimized) to 109 pixels (Cifar10 optimized).
The effect of these reductions can be seen in figure \ref{fig:resnet_cif_receptive field}: the proportion of low saturated layers is drastically reduced for both models and the inference process is now well/more evenly distributed.
Thus the performance increases in both cases compared to the ImageNet optimized models. The accuracy of ResNet18 improves from 84.61\% (ImageNet optimized) to 91.95\% (Cifar10 optimized)
ResNet34 improves from 82.76\% to 92.21\%.
This example illustrates that a mismatch of dataset and architecture can be resolved by altering the architecture in a way that brings the receptive field size closer to the input resolution.

\section{Conclusion}
For the longest time in deep learning, developing increasingly bigger and more complex models to improve performance has been the norm.
Our work provides a counterpoint, showing that the neural architecture and the visual properties of the dataset must match in order to avoid inefficiencies like layers not contributing to the quality of the solution and sub-optimal predictive performance.
We show that the receptive field size is a primary architectural factor in this interaction.
We further show that the residual connections can counteract (but not entirely mitigate) mismatches of input size and receptive field size by allowing more layers to contribute to the inference qualitatively.
Finally, we demonstrate that probe classifiers and saturation are useful tools to analyze the relationship between dataset and model architecture allowing the optimization of CNN design in a principled way.
We conclude that the size of the input image matters for optimal predictive and computational performance.

\bibliography{references}

\begin{thebibliography}{24}
\providecommand{\natexlab}[1]{#1}
\providecommand{\url}[1]{\texttt{#1}}
\expandafter\ifx\csname urlstyle\endcsname\relax
  \providecommand{\doi}[1]{doi: #1}\else
  \providecommand{\doi}{doi: \begingroup \urlstyle{rm}\Url}\fi

\bibitem[Alain \& Bengio(2016)Alain and Bengio]{alain2016}
Alain, G. and Bengio, Y.
\newblock Understanding intermediate layers using linear classifier probes.
\newblock \emph{ArXiv}, abs/1610.01644, 2016.

\bibitem[Bochkovskiy et~al.(2020)Bochkovskiy, Wang, and Liao]{yolov4}
Bochkovskiy, A., Wang, C.-Y., and Liao, H.-Y.~M.
\newblock {YOLOv4}: Optimal speed and accuracy of object detection, 2020.

\bibitem[Bossard et~al.(2014)Bossard, Guillaumin, and Van~Gool]{food101}
Bossard, L., Guillaumin, M., and Van~Gool, L.
\newblock Food-101 -- mining discriminative components with random forests.
\newblock In Fleet, D., Pajdla, T., Schiele, B., and Tuytelaars, T. (eds.),
  \emph{Computer Vision -- ECCV 2014}, pp.\  446--461, Cham, 2014. Springer
  International Publishing.
\newblock ISBN 978-3-319-10599-4.

\bibitem[Han et~al.(2016)Han, Kim, and Kim]{pyramidnets}
Han, D., Kim, J., and Kim, J.
\newblock Deep pyramidal residual networks.
\newblock \emph{CoRR}, abs/1610.02915, 2016.
\newblock URL \url{http://arxiv.org/abs/1610.02915}.

\bibitem[He et~al.(2015)He, Zhang, Ren, and Sun]{resnet}
He, K., Zhang, X., Ren, S., and Sun, J.
\newblock Deep residual learning for image recognition.
\newblock \emph{CoRR}, abs/1512.03385, 2015.

\bibitem[Horn et~al.(2017)Horn, Aodha, Song, Shepard, Adam, Perona, and
  Belongie]{iNaturalist}
Horn, G.~V., Aodha, O.~M., Song, Y., Shepard, A., Adam, H., Perona, P., and
  Belongie, S.~J.
\newblock The {iNaturalist} challenge 2017 dataset.
\newblock \emph{CoRR}, abs/1707.06642, 2017.
\newblock URL \url{http://arxiv.org/abs/1707.06642}.

\bibitem[Huang et~al.(2018)Huang, Cheng, Chen, Lee, Ngiam, Le, and Chen]{gpipe}
Huang, Y., Cheng, Y., Chen, D., Lee, H., Ngiam, J., Le, Q.~V., and Chen, Z.
\newblock Gpipe: Efficient training of giant neural networks using pipeline
  parallelism.
\newblock \emph{CoRR}, abs/1811.06965, 2018.
\newblock URL \url{http://arxiv.org/abs/1811.06965}.

\bibitem[Iandola et~al.(2014)Iandola, Moskewicz, Karayev, Girshick, Darrell,
  and Keutzer]{densenet}
Iandola, F.~N., Moskewicz, M.~W., Karayev, S., Girshick, R.~B., Darrell, T.,
  and Keutzer, K.
\newblock {DenseNet}: Implementing efficient convnet descriptor pyramids.
\newblock \emph{CoRR}, abs/1404.1869, 2014.
\newblock URL \url{http://arxiv.org/abs/1404.1869}.

\bibitem[Iandola et~al.(2016)Iandola, Moskewicz, Ashraf, Han, Dally, and
  Keutzer]{squeezenet}
Iandola, F.~N., Moskewicz, M.~W., Ashraf, K., Han, S., Dally, W.~J., and
  Keutzer, K.
\newblock {SqueezeNet}: {AlexNet}-level accuracy with 50x fewer parameters and
  {\textless}1mb model size.
\newblock \emph{CoRR}, abs/1602.07360, 2016.
\newblock URL \url{http://arxiv.org/abs/1602.07360}.

\bibitem[Krizhevsky et~al.(2010)Krizhevsky, Nair, and Hinton]{cifar}
Krizhevsky, A., Nair, V., and Hinton, G.
\newblock {CIFAR-10} (canadian institute for advanced research), 2010.
\newblock URL \url{http://www.cs.toronto.edu/~kriz/cifar.html}.

\bibitem[Krizhevsky et~al.(2012)Krizhevsky, Sutskever, and Hinton]{alexnet}
Krizhevsky, A., Sutskever, I., and Hinton, G.~E.
\newblock {ImageNet} classification with deep convolutional neural networks.
\newblock In Pereira, F., Burges, C. J.~C., Bottou, L., and Weinberger, K.~Q.
  (eds.), \emph{Advances in Neural Information Processing Systems 25}, pp.\
  1097--1105. Curran Associates, Inc., 2012.

\bibitem[Le \& Yang(2015)Le and Yang]{tinyimagenet}
Le, Y. and Yang, X.
\newblock Tiny {ImageNet} visual recognition challenge, 2015.

\bibitem[LeCun \& Cortes(2010)LeCun and Cortes]{mnist}
LeCun, Y. and Cortes, C.
\newblock {MNIST} handwritten digit database.
\newblock http://yann.lecun.com/exdb/mnist/, 2010.
\newblock URL \url{http://yann.lecun.com/exdb/mnist/}.

\bibitem[Lin et~al.(2013)Lin, Chen, and Yan]{gap}
Lin, M., Chen, Q., and Yan, S.
\newblock Network in network, 2013.
\newblock URL \url{http://arxiv.org/abs/1312.4400}.
\newblock cite arxiv:1312.4400Comment: 10 pages, 4 figures, for iclr2014.

\bibitem[Redmon \& Farhadi(2016)Redmon and Farhadi]{yolov2}
Redmon, J. and Farhadi, A.
\newblock {YOLO9000:} better, faster, stronger.
\newblock \emph{CoRR}, abs/1612.08242, 2016.
\newblock URL \url{http://arxiv.org/abs/1612.08242}.

\bibitem[Redmon \& Farhadi(2018)Redmon and Farhadi]{yolov3}
Redmon, J. and Farhadi, A.
\newblock {YOLOv3}: An incremental improvement.
\newblock \emph{CoRR}, abs/1804.02767, 2018.
\newblock URL \url{http://arxiv.org/abs/1804.02767}.

\bibitem[Redmon et~al.(2015)Redmon, Divvala, Girshick, and Farhadi]{yolo}
Redmon, J., Divvala, S.~K., Girshick, R.~B., and Farhadi, A.
\newblock You only look once: Unified, real-time object detection.
\newblock \emph{CoRR}, abs/1506.02640, 2015.
\newblock URL \url{http://arxiv.org/abs/1506.02640}.

\bibitem[Richter et~al.(2020)Richter, Shenk, Byttner, Arpteg, and
  Huss]{saturation}
Richter, M.~L., Shenk, J., Byttner, W., Arpteg, A., and Huss, M.
\newblock Feature space saturation during training, 2020.

\bibitem[Russakovsky et~al.(2015)Russakovsky, Deng, Su, Krause, Satheesh, Ma,
  Huang, Karpathy, Khosla, Bernstein, Berg, and Fei-Fei]{ImageNet}
Russakovsky, O., Deng, J., Su, H., Krause, J., Satheesh, S., Ma, S., Huang, Z.,
  Karpathy, A., Khosla, A., Bernstein, M., Berg, A.~C., and Fei-Fei, L.
\newblock {ImageNet} large scale visual recognition challenge.
\newblock \emph{International Journal of Computer Vision (IJCV)}, 115\penalty0
  (3):\penalty0 211--252, 2015.
\newblock \doi{10.1007/s11263-015-0816-y}.

\bibitem[Simonyan \& Zisserman(2014)Simonyan and Zisserman]{vgg}
Simonyan, K. and Zisserman, A.
\newblock Very deep convolutional networks for large-scale image recognition.
\newblock \emph{CoRR}, abs/1409.1556, 2014.

\bibitem[Szegedy et~al.(2015)Szegedy, Liu, Jia, Sermanet, Reed, Anguelov,
  Erhan, Vanhoucke, and Rabinovich]{inception}
Szegedy, C., Liu, W., Jia, Y., Sermanet, P., Reed, S., Anguelov, D., Erhan, D.,
  Vanhoucke, V., and Rabinovich, A.
\newblock Going deeper with convolutions.
\newblock In \emph{Proceedings of the IEEE Conference on Computer Vision and
  Pattern Recognition (CVPR)}, June 2015.

\bibitem[Szegedy et~al.(2016)Szegedy, Vanhoucke, Ioffe, Shlens, and
  Wojna]{inceptionv3}
Szegedy, C., Vanhoucke, V., Ioffe, S., Shlens, J., and Wojna, Z.
\newblock Rethinking the inception architecture for computer vision.
\newblock In \emph{IEEE Conference on Computer Vision and Pattern Recognition
  (CVPR)}, pp.\  2818--2826, 2016.
\newblock \doi{10.1109/CVPR.2016.308}.

\bibitem[Tan \& Le(2019)Tan and Le]{efficentnet}
Tan, M. and Le, Q.
\newblock {E}fficient{N}et: Rethinking model scaling for convolutional neural
  networks.
\newblock In Chaudhuri, K. and Salakhutdinov, R. (eds.), \emph{Proceedings of
  the 36th International Conference on Machine Learning}, volume~97 of
  \emph{Proceedings of Machine Learning Research}, pp.\  6105--6114. PMLR,
  09--15 Jun 2019.

\bibitem[Zhang et~al.(2017)Zhang, Zhou, Lin, and Sun]{shufflenet}
Zhang, X., Zhou, X., Lin, M., and Sun, J.
\newblock {ShuffleNet}: An extremely efficient convolutional neural network for
  mobile devices.
\newblock \emph{CoRR}, abs/1707.01083, 2017.
\newblock URL \url{http://arxiv.org/abs/1707.01083}.

\end{thebibliography}
\bibliographystyle{icml2021}

%%%%%%%%%%%%%%%%%%%%%%%%%%%%%%%%%%%%%%%%%%%%%%%%%%%%%%%%%%%%%%%%%%%%%%%%%%%%%%%
%%%%%%%%%%%%%%%%%%%%%%%%%%%%%%%%%%%%%%%%%%%%%%%%%%%%%%%%%%%%%%%%%%%%%%%%%%%%%%%
% DELETE THIS PART. DO NOT PLACE CONTENT AFTER THE REFERENCES!
%%%%%%%%%%%%%%%%%%%%%%%%%%%%%%%%%%%%%%%%%%%%%%%%%%%%%%%%%%%%%%%%%%%%%%%%%%%%%%%
%%%%%%%%%%%%%%%%%%%%%%%%%%%%%%%%%%%%%%%%%%%%%%%%%%%%%%%%%%%%%%%%%%%%%%%%%%%%%%%

\clearpage
\appendix

\onecolumn

\section{Code and Experimental Setup}
The full code used for conducting all depicted experiments as well as a guide for reproducing the results can be found here:

\url{https://anonymous.4open.science/r/d04fa611-aa00-4c22-9158-5af363b68c3a/}

\section{Handling convolutional feature maps when computing probe performance}
A practical problem for computing the performance of probes is the size of convolutional feature maps. In order to make the experiments computationally tractable, it is necessary to reduce the dimensionality of those feature maps.

When probes were first proposed by \citet{alain2016} two ways of handling this practical problem are tested, Global Average Pooling (GAP) and randomized subsampling of the feature maps. 
However, random sub sampling introduces artefacts \cite{saturation}.  %We also decided against GAP, since we suspected that this reduction strategy may introduce increasingly strong artifacts on ealier layers with larger resolutions and more local features. %% Johan - this is a strong statement with no evidence to back it
Also, \citet{saturation} conducted GAP experiments by looking on the ablative effects of downsampling and adaptive pooling strategies on VGG and ResNet model trained on input sizes between $32 \times 32$ pixel $224 \times 224$ pixel on Food101 and Cifar10 \cite{food101, cifar}.
We conclude that the effect on Global Pooling are diminishing as the feature map increases but the overall structure does not change.
In order to reduce this ablative effect while still making probe performance computation practical as much as possible we decided on a $4 \times 4$ adaptive average pooling over the feature maps. 
The experiments showed diminishingly small changes in the features and the overall performance levels of probes compared to larger downsampling strategies.

In experiments for showing the evolution of the of the feature map in conjunction with the receptive field expansion, we also trained probes on each position of all feature maps, which allows us to generate performance heat maps, where we are able to show how much information required for solving the task is contained on a specific position of the feature map.

\subsection{A more detailed explanation on receptive field size and its computation}
The receptive field size is a property of a layer inside a neural network architecture. 
It describes the size (height and width) of an rectangle on an infinitely large input image. This area contains all pixels that can influence the output of a single position of the convolutional kernel on the feature map.
Simply speaking, it describes the spatial limits of visual features detectable by the respective layer.

It is worth noting that the size (in theory) is a rectangle, since the kernels of CNNs may have an arbitrary height and width. therefore the receptive field size is technically a 2-tuple consisting of two integers describing height and width of the rectangle. 
However, conventionally networks have square kernels and are fed square images, which are either padded, cropped or scaled to the correct aspect ratio using data augmentation \cite{alexnet, vgg, resnet, inception, inceptionv3, efficentnet, squeezenet, densenet}.
We thus refer to the receptive field size as an integer for the sake of simplicity.

\subsubsection{Computing the receptive field}
For convolutional neural networks with a sequential structure (no multiple pathways during the forward pass) the receptive field size can be computed analytically.
We refer to the receptive field $r$ of the $l$th layer of sequential network structure as $r_l$ (with $r_0 = 1$, which is the "receptive field" of the input).
For all layers $l > 0$ in the convolutional part of the network the receptive field can be computed with the following formula:

\begin{equation}
    r_l = r_{l-1} + ((k_{l} - 2) \prod^{l-1}_{i=0} s_{i})
\end{equation}

Where $r_{l-1}$ is the receptive field of the previous layer, $k_l$ refers to the kernel size of layer $l$ (with potential dilation already accounted for) and $s_{i}$ the stride size of layer $i$.
In the simplest case, which is the first convolutional layer of the neural architecture $r_1$ and stride size of 1, the receptive field is exactly the size of the kernel, therefore the receptive field size is the kernel size.
When convolutional layers are stacked, the receptive field for every layer $l$ with a kernel size $k_l \neq 1$ expands the receptive field for following layers, because the filter integrates information of an area of the input feature map into a single output position.
Downsampling and pooling layers that reduce the size of the feature map will have $s_i \neq 1$, which has a multiplicative effect on the receptive field expansion of consecutive layers. 
In simple cases like VGG16, where the max-pooling layers are $2 \times 2$ filters with $s_i = 2$, this can be intuitively imagined as 4 adjacent pixels being summarized into a single "super pixel".
This effect on the growth of $r$ is also compounding exponentially when multiple downsampling layers are employed in the neural architecture, which is generally the case for most popular architectures \cite{vgg, resnet, inception, inceptionv3, densenet, squeezenet, shufflenet, efficentnet, inceptionv3, squeezenet}.

\subsubsection{The receptive field in network with multiple pathways}
The method described previously allows the computation of the receptive field for sequential architectures with each layer having at most a single previous layer.
However, since the advent of deep learning multiple pathways have been common in neural architectures.
In this scenario there $s_{i}$ and $r_{l-1}$ may be tuples of values with each element describing the stride size and the receptive field size  of a pathway.
However, the common strategies for funneling multiple paths into a single successive layer are simple element-wise operation (like additions and multiplications) or concatenations of the respective stacks of feature maps produced by each path. 
Both cases have in common that the size of the feature maps must agree, which is achieved by either different means influencing the size of the receptive field.
However, we are not interested in knowing exactly which feature map in an output layer considers which information.
We are only interested in the size of the area that contains all information that can theoretically influence a position in the feature map. 
Hence, we ignore all pathways except the one with the largest receptive field.
It is worth noting that we make the assumption that all pathways downsample the image identically $s_{l, 1} = s_{l, 2} = ... = s_{l, n}$, so that the growth rate of the receptive field is increased evenly on different pathways.
While this assumption is not true for all possible architectures it holds true for all architectures considered in this paper and most popular architectures like VGG, ResNet, InceptionV3 and EfficentNet \cite{vgg, resnet, inceptionv3, efficentnet}.

\clearpage

\section{Resolution Shifts the Distribution of the Inference Process on different Models}
In Section 3.2 we show on a ResNet18 architecture how resolution affects the distribution of the inference process by training ResNet18 on 3 different resolutions on Cifar10.

To show that this observation does generalize we repeat the experiment. 
We use TinyImageNet as a more complex but still small resolution problem.
We further show that the hypothesis holds also true for VGG16 and ResNet50 models.

\subsection{VGG16 - Cifar10}
\FloatBarrier

\begin{figure}[!htb]
	\centering
	\subfloat[$32 \times 32$ (Cifar10 native resolution)]{
	    \includegraphics[width=0.5\columnwidth]{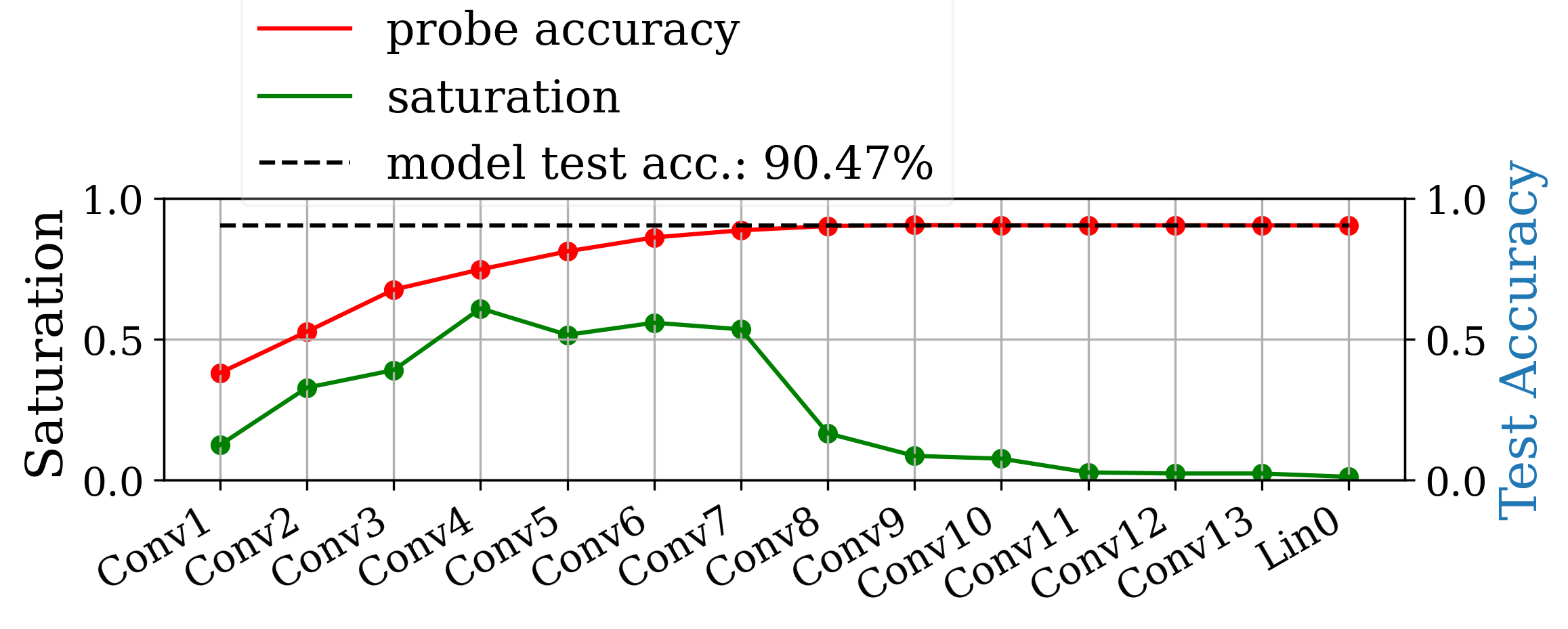}
	    \label{fig:vgg16_cifar10_small}
	    %\label{fig:iNaturalistVgg16_small}
	} \qquad
	\subfloat[$160 \times 160$]{
	    \includegraphics[width=0.5\columnwidth]{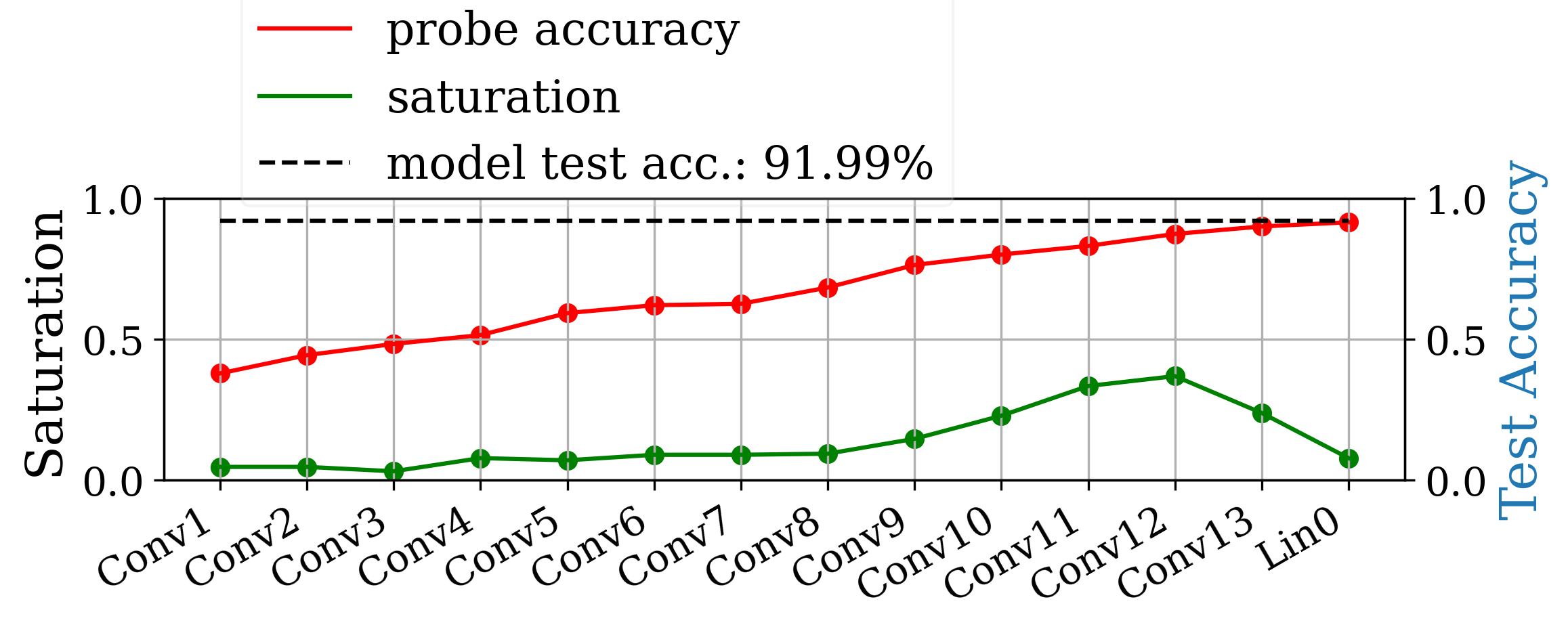}
	    \label{fig:vgg16_cifar10_medium}
	}
	\quad
	\subfloat[$1024 \times 1024$]{
	    \includegraphics[width=0.5\columnwidth]{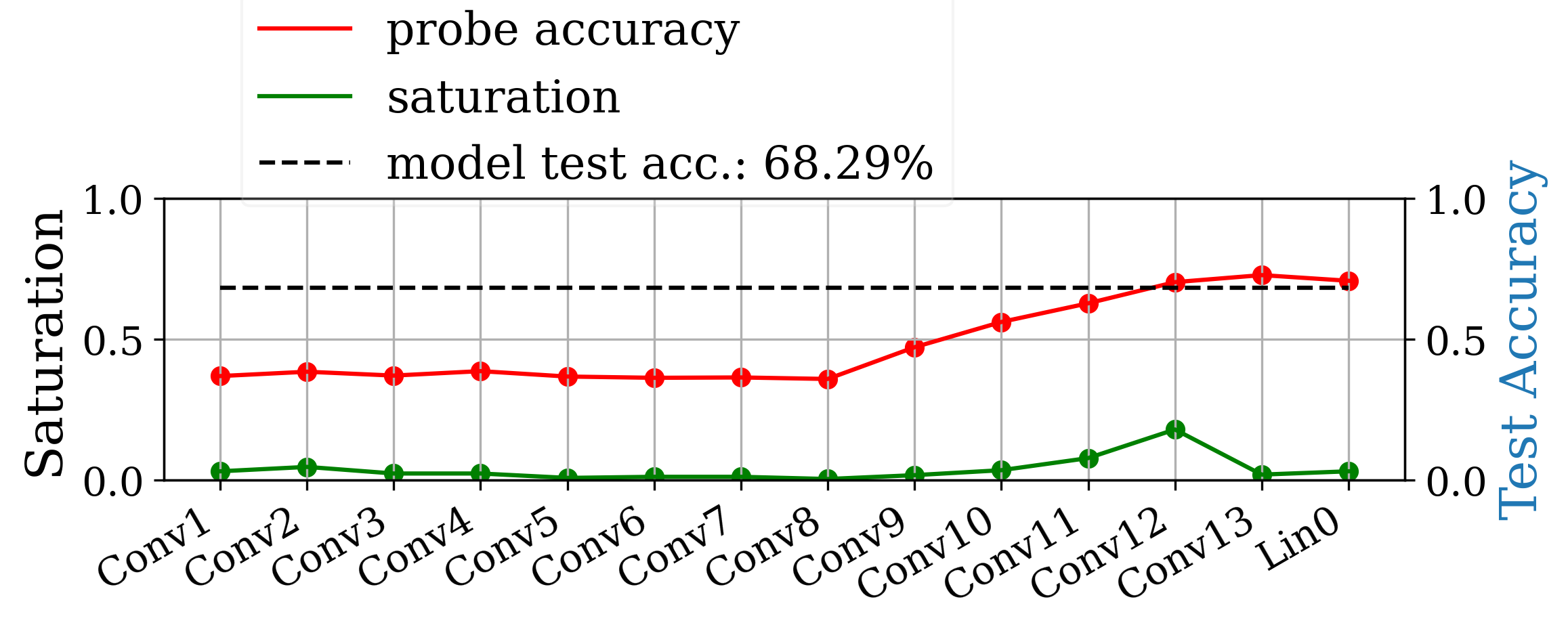}
	    \label{fig:vgg16_cifar10_big}

	    %\label{fig:iNaturalistVgg16_big}
	}
	\caption{Reproduction of the experiment depicted in figure \ref{fig:resnet18_cifar10_resolution} using the VGG16 architecture. Because of the larger memory footprint of the model, the batch size of this experiment is reduced to 20. The observed pattern stays the same. Low resolution results in a tail close to the output, while high resolution exhibits a tail at the input.} The (medium sized) resolution of $160 \times 160$ pixels performs best.
	\label{fig:vgg16_cifar10_resolution}

\end{figure}
\FloatBarrier
\clearpage

\subsection{ResNet18 - TinyImageNet}
TinyImageNet consists of 200 classes of $64 \times 64$ images.
Therefore the first experiment is conducted on $64 \times 64$ pixel input resolution instead of $32 \times 32$ pixel resolution, since this is the basic resolution of the dataset.
\FloatBarrier

\begin{figure}[!htb]
	\centering
	\subfloat[$64 \times 64$ (TinyImageNet native resolution)]{
	    \includegraphics[width=0.5\columnwidth]{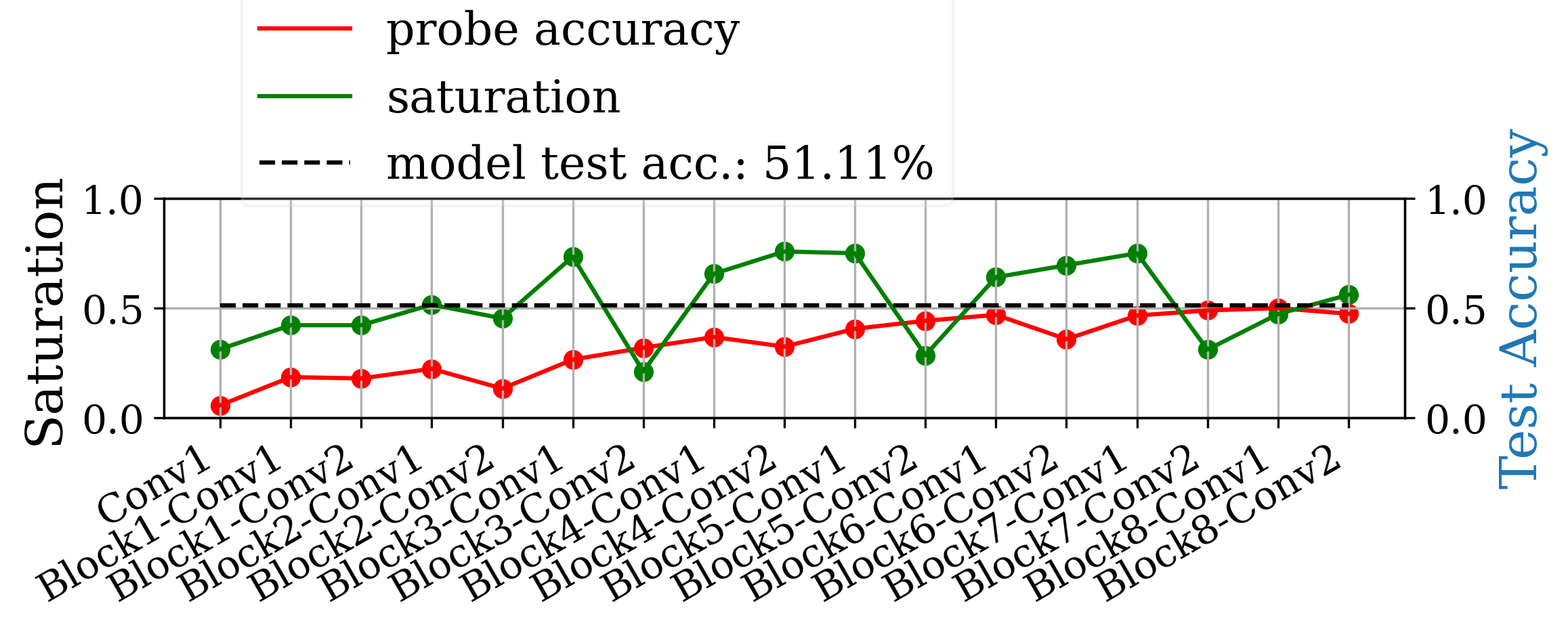}
	    \label{fig:resnet18_tinyimagenet_small}
	    %\label{fig:iNaturalistVgg16_small}
	} 
	\quad
	\subfloat[$1024 \times 1024$]{
	    \includegraphics[width=0.5\columnwidth]{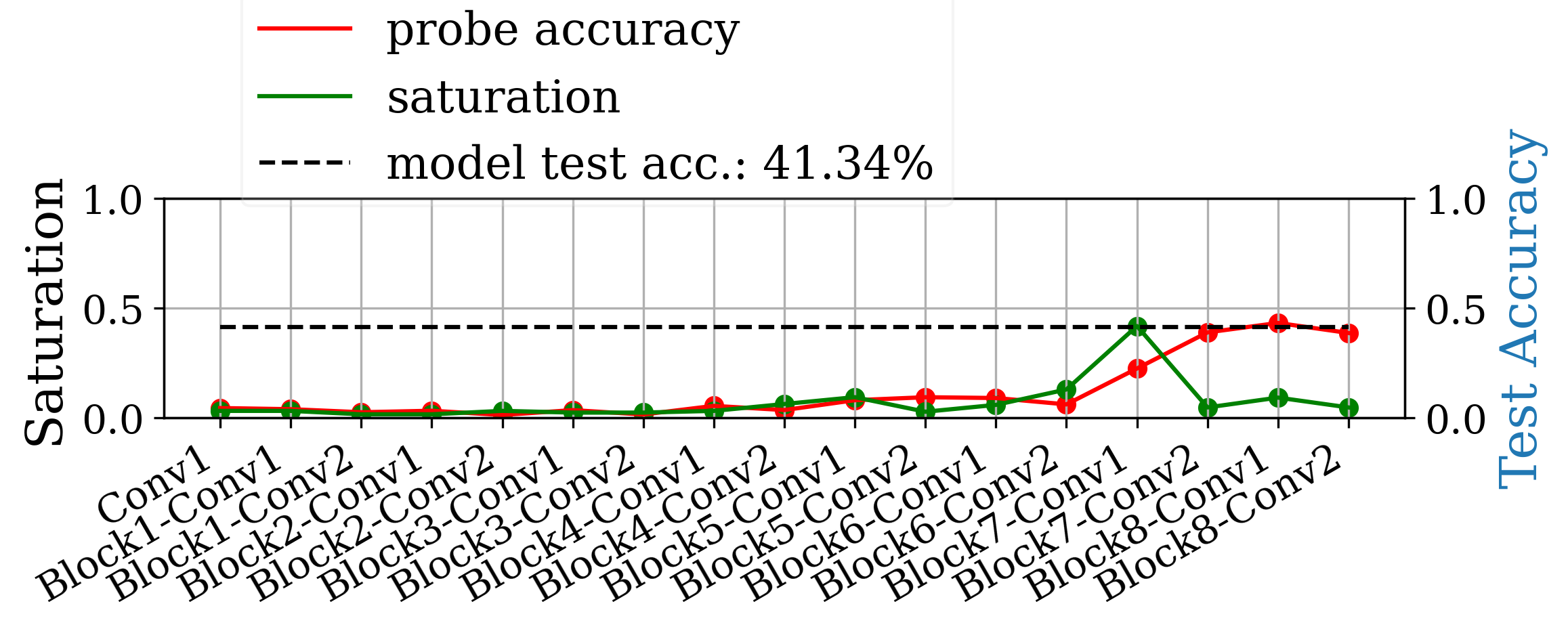}
	    \label{fig:resnet18_tinyimagenet_big}

	    %\label{fig:iNaturalistVgg16_big}
	}
	\caption{The higher native resolution combined with the residual connection removes the tail pattern even at $64 \times 64$ pixels for ResNet18. We see that a tail pattern can be produced in the input part of the network when the resolution is increased drastically to $1024 \times 1024$, confirming observations from Cifar10.}
	\label{fig:resnet18_tinyimagenet_resolution}

\end{figure}
\FloatBarrier
\clearpage

\subsection{ResNet50 - Cifar10}
Due to an error in the setup, the mid-sized image experiment was conducted with a $160 \times 160$ pixel resolution instead of $224 \times 224$.
However, even thought the resolution was altered it still shows the exspected behavior observed with other models.
\FloatBarrier

\begin{figure}[!htb]
	\centering
	\subfloat[$32 \times 32$ (Cifar10 native resolution)]{
	    \includegraphics[width=0.8\columnwidth]{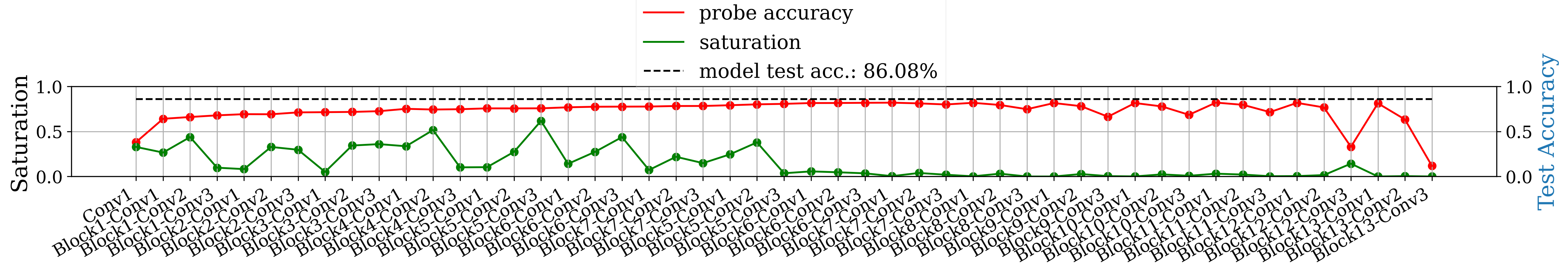}
	    \label{fig:resnet50_Cifar10_small}
	    %\label{fig:iNaturalistVgg16_small}
	} \qquad
	\subfloat[$160 \times 160$ (ResNet standard)]{
	    \includegraphics[width=0.8\columnwidth]{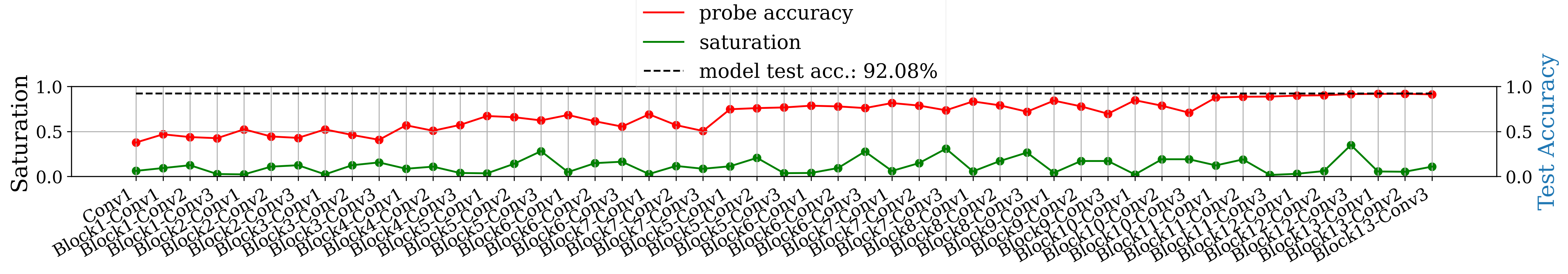}
	    \label{fig:resnet50_Cifar10_medium}
	}
	\quad
	\subfloat[$1024 \times 1024$]{
	    \includegraphics[width=0.8\columnwidth]{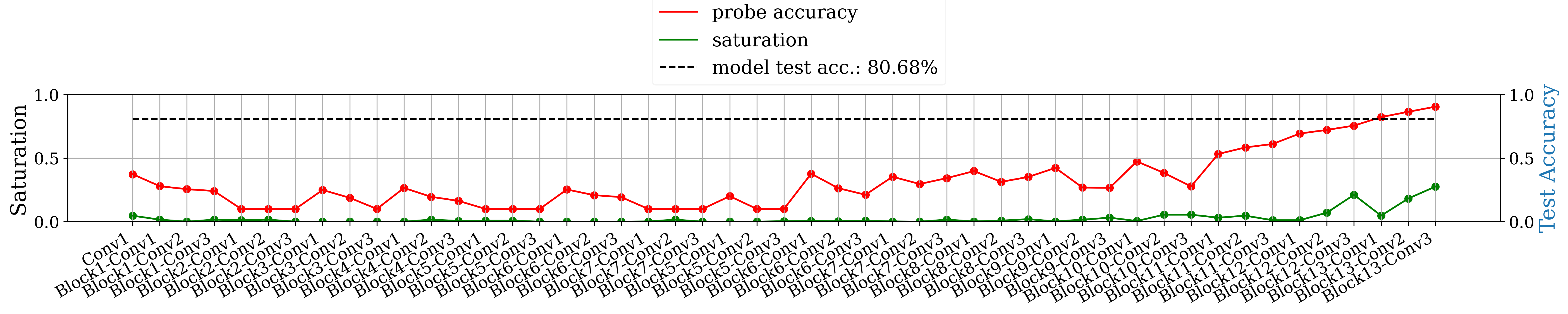}
	    \label{fig:resnet50_Cifar10_big}

	    %\label{fig:iNaturalistVgg16_big}
	}
	\caption{Producing a tail pattern in the near the input and the output is also possible for very deep models like ResNet50 with the exspected effects on the performance. The Bottleneck-Modules with $1 \times 1$ convolutions and varying filter sizes induce additional noise into saturation patterns.}
	\label{fig:resnet50_Cifar10_resolution}

\end{figure}
\FloatBarrier
\clearpage

\section{Object Size Experiments on MNIST and ResNet models}
\begin{figure}[!htb]
	\centering
	\subfloat[$32 \times 32$ without canvas]{
	    \includegraphics[width=0.5\columnwidth]{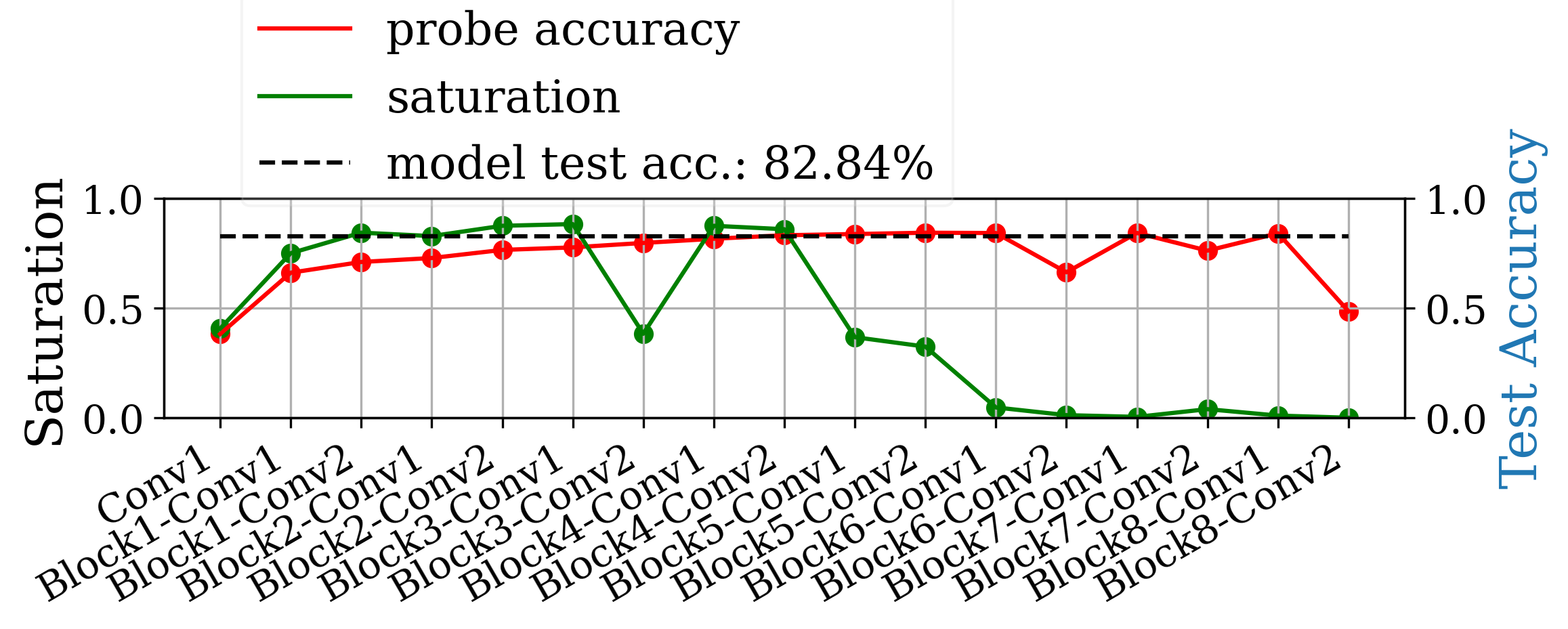}
	    \label{fig:resnet18_rpCifar10_small}
	    %\label{fig:iNaturalistVgg16_small}
	} \qquad
		\subfloat[$160 \times 160$ with canvas]{
	    \includegraphics[width=0.5\columnwidth]{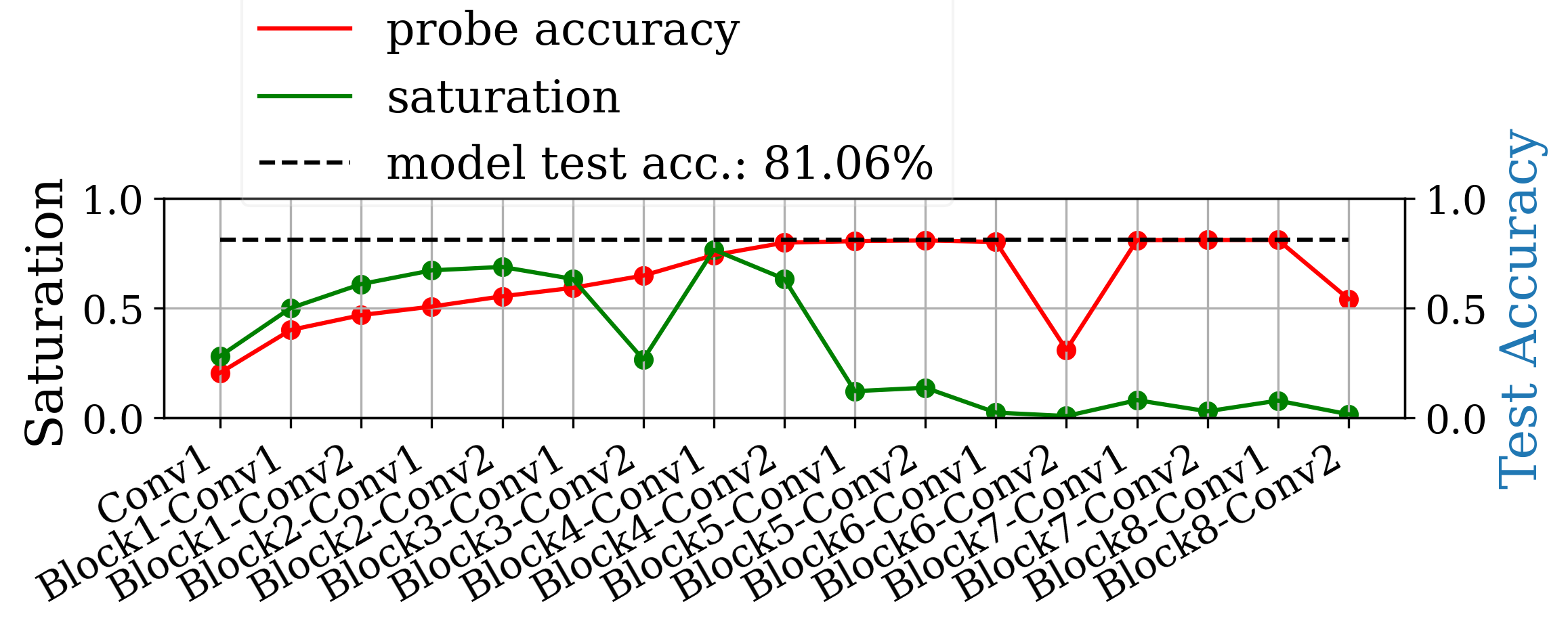}
	    \label{fig:resnet18_rpCifar10_big}

	    %\label{fig:iNaturalistVgg16_big}
	} \quad
	\subfloat[$160 \times 160$ upsampled]{
	    \includegraphics[width=0.5\columnwidth]{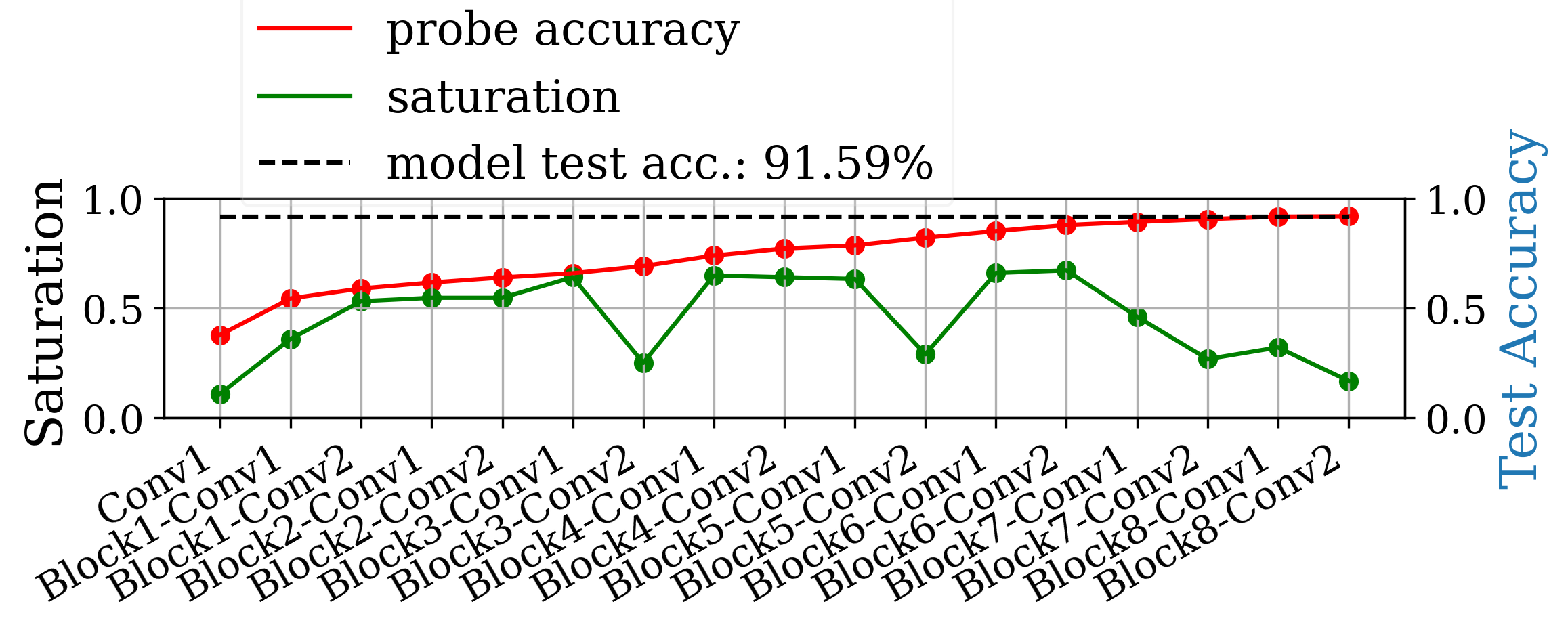}
	    \label{fig:resnet18_rpCifar10_medium}
	}

	\caption{Random Positioning Experiments conducted with ResNet18 on Cifar10 data.}
	\label{fig:resnet18_rpCifar10_resolution}

\end{figure}

\begin{figure}[!htb]
	\centering
	\subfloat[$32 \times 32$ without canvas]{
	    \includegraphics[width=0.5\columnwidth]{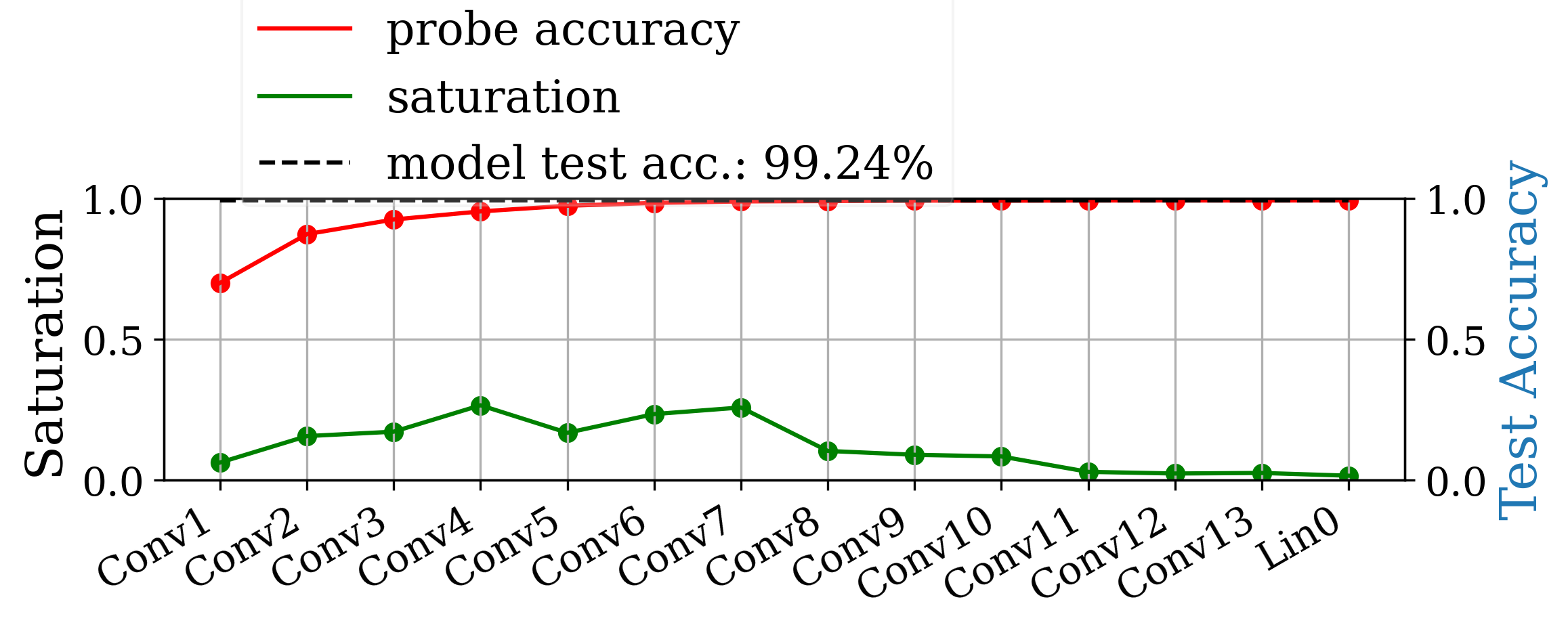}
	    %\label{fig:iNaturalistVgg16_resized}
	}
	\quad
	\subfloat[$160 \times 160$ with canvas]{
	    \includegraphics[width=0.5\columnwidth]{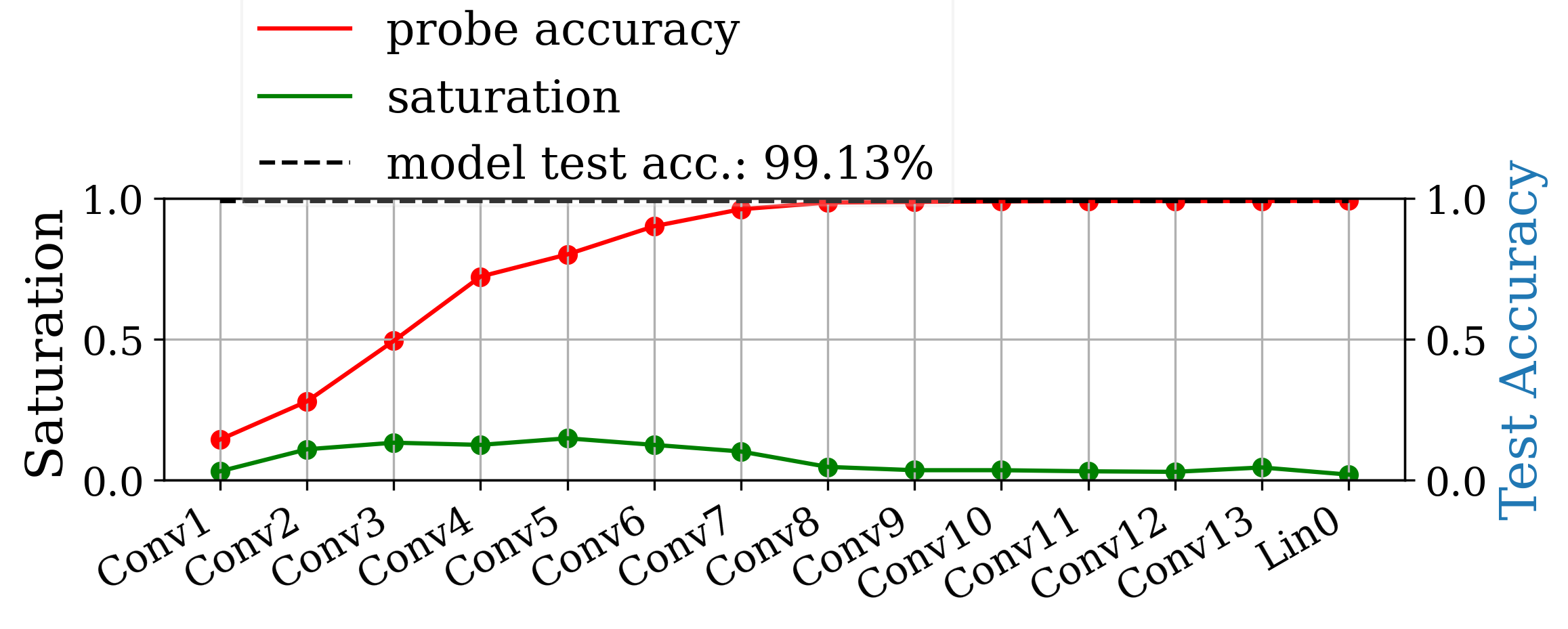}
	    %\label{fig:iNaturalistVgg16_bigvgg}
	}\quad
	\subfloat[$160 \times 160$ upsampled]{
	    \includegraphics[width=0.5\columnwidth]{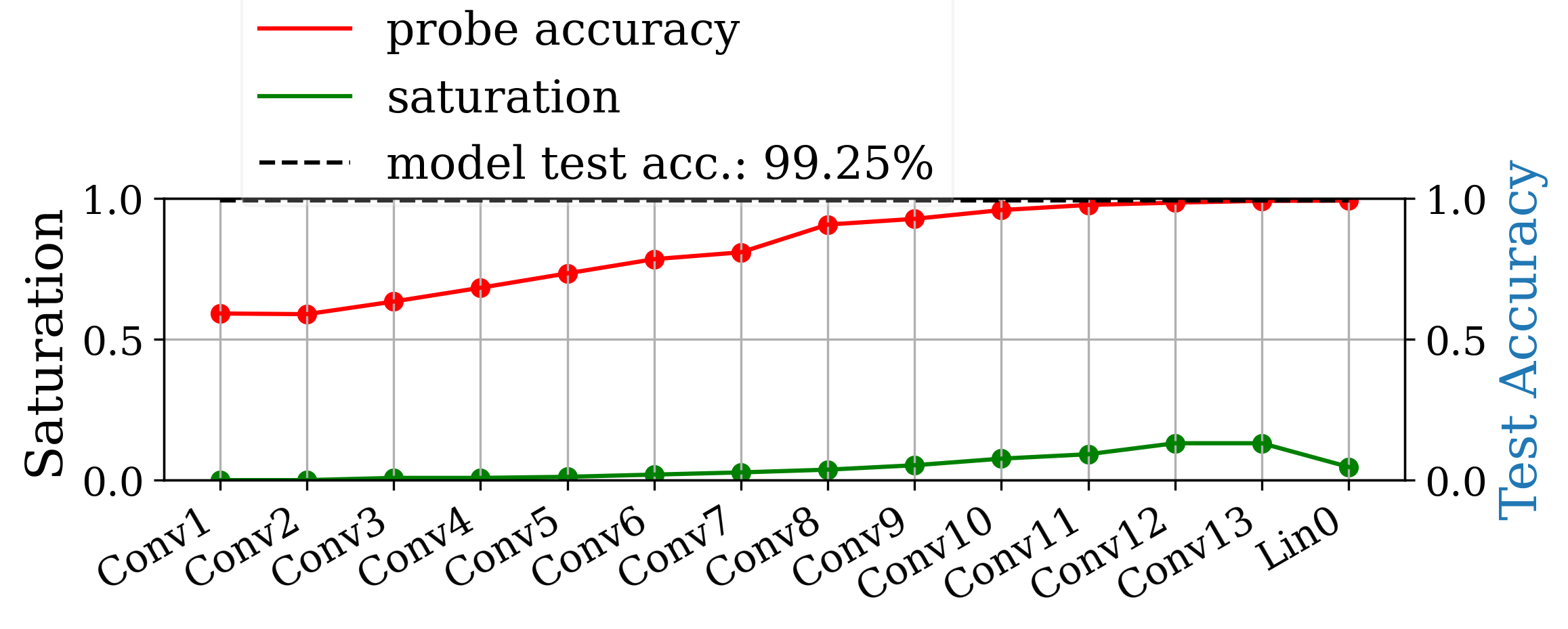}
	    %\label{fig:iNaturalistVgg16_big}
	    \label{fig:object_size_upsampledvgg}

	}
	\caption{Random Positioning Experiments conducted with VGG16 on MNIST data.}
	\label{fig:object_size3vgg}

\end{figure}

\begin{figure}[!htb]
	\centering
	\subfloat[$32 \times 32$ (Cifar10 native resolution)]{
	    \includegraphics[width=0.8\columnwidth]{figures/resolution_exp/ResNet50_Cifar10_32.png}
	    \label{fig:resnet50_rpCifar10_small}
	    %\label{fig:iNaturalistVgg16_small}
	} \qquad
	\subfloat[$160 \times 160$ with canvas]{
	    \includegraphics[width=0.8\columnwidth]{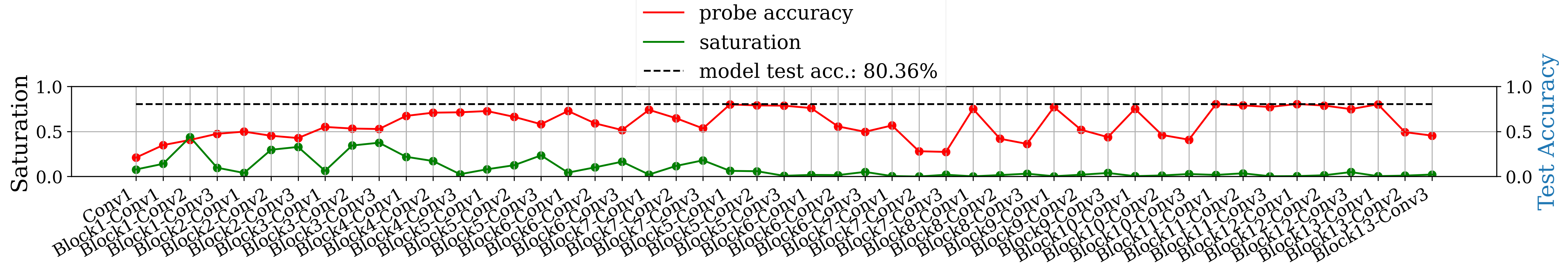}
	    \label{fig:resnet50_rpCifar10_medium}
	}
	\quad
	\subfloat[$160 \times 160$ upsampled]{
	    \includegraphics[width=0.8\columnwidth]{figures/resolution_exp/ResNet50_Cifar10_160.png}
	    \label{fig:resnet50_rpCifar10_big}

	    %\label{fig:iNaturalistVgg16_big}
	}
	\caption{Random Positioning Experiments repeated on ResNet50.}
	\label{fig:object_size4}

\end{figure}

In Section 3.3 we make the claim that the observed relation of input resolution and the performance of the model is an artifact caused by the underlying interaction of object size and neural architecture.
We repeat the experiments that demonstrate this claim on ResNet-style models as well as on the MNIST dataset.
As we can see in figure~\ref{fig:resnet18_rpCifar10_resolution} and figure~\ref{fig:object_size3vgg} and figure~\ref{fig:object_size4}, the results are consistent with the observations shown in the paper in all three scenarios.

\subsection{ResNet18 - Cifar10}

\FloatBarrier
\FloatBarrier
\clearpage

\subsection{VGG16 - MNIST}
\FloatBarrier

\FloatBarrier
\clearpage

\subsection{ResNet50 - Cifar10}

\FloatBarrier
\FloatBarrier
\clearpage

\section{Additional Experiments Regarding the Relationship of Receptive Field and Input Resolution}
\begin{figure}[htb!]
	\centering
	\subfloat[VGG11]{
	    \includegraphics[width=0.45\columnwidth]{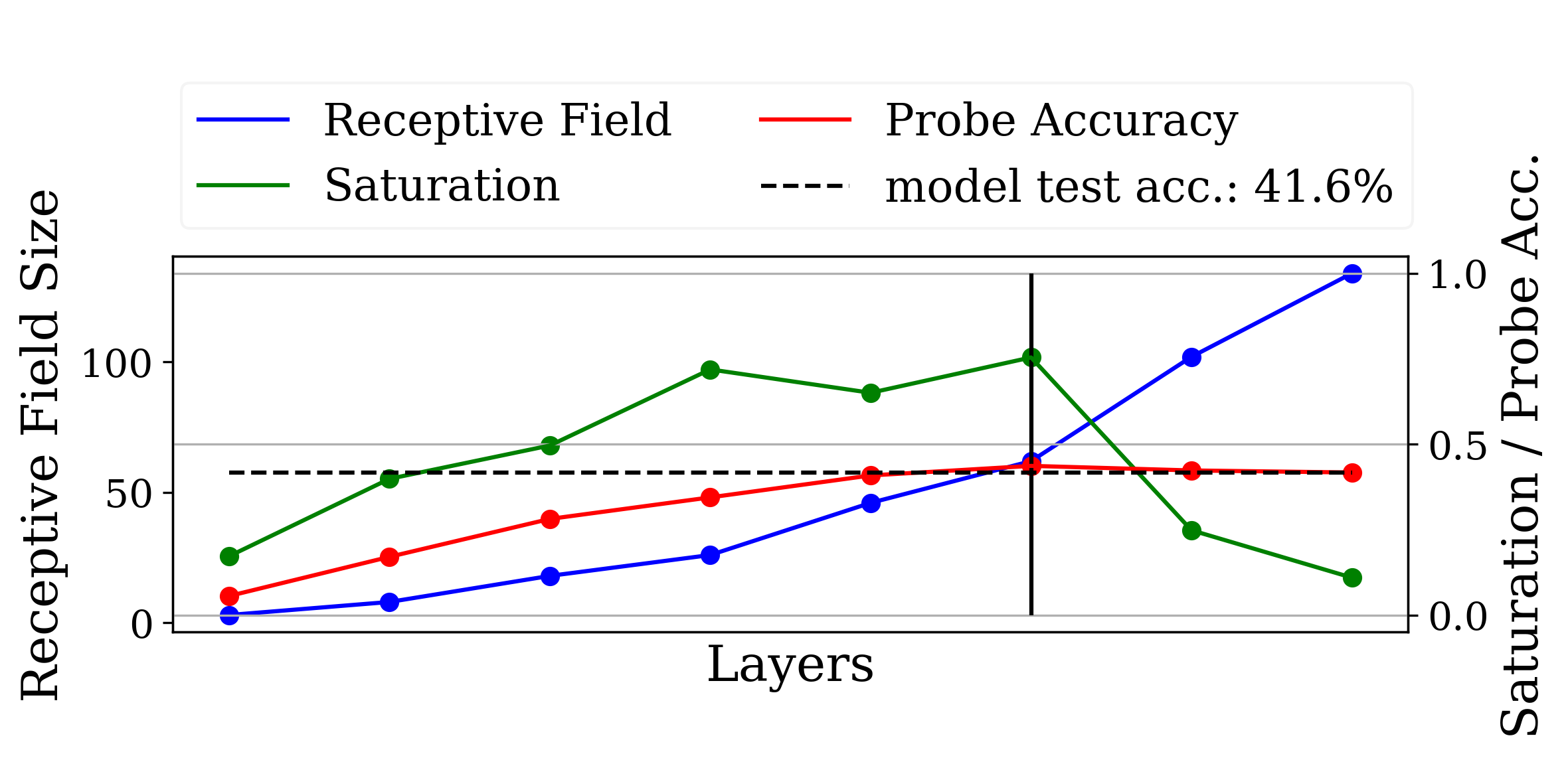}
	}\quad
		\subfloat[VGG13]{
	    \includegraphics[width=0.45\columnwidth]{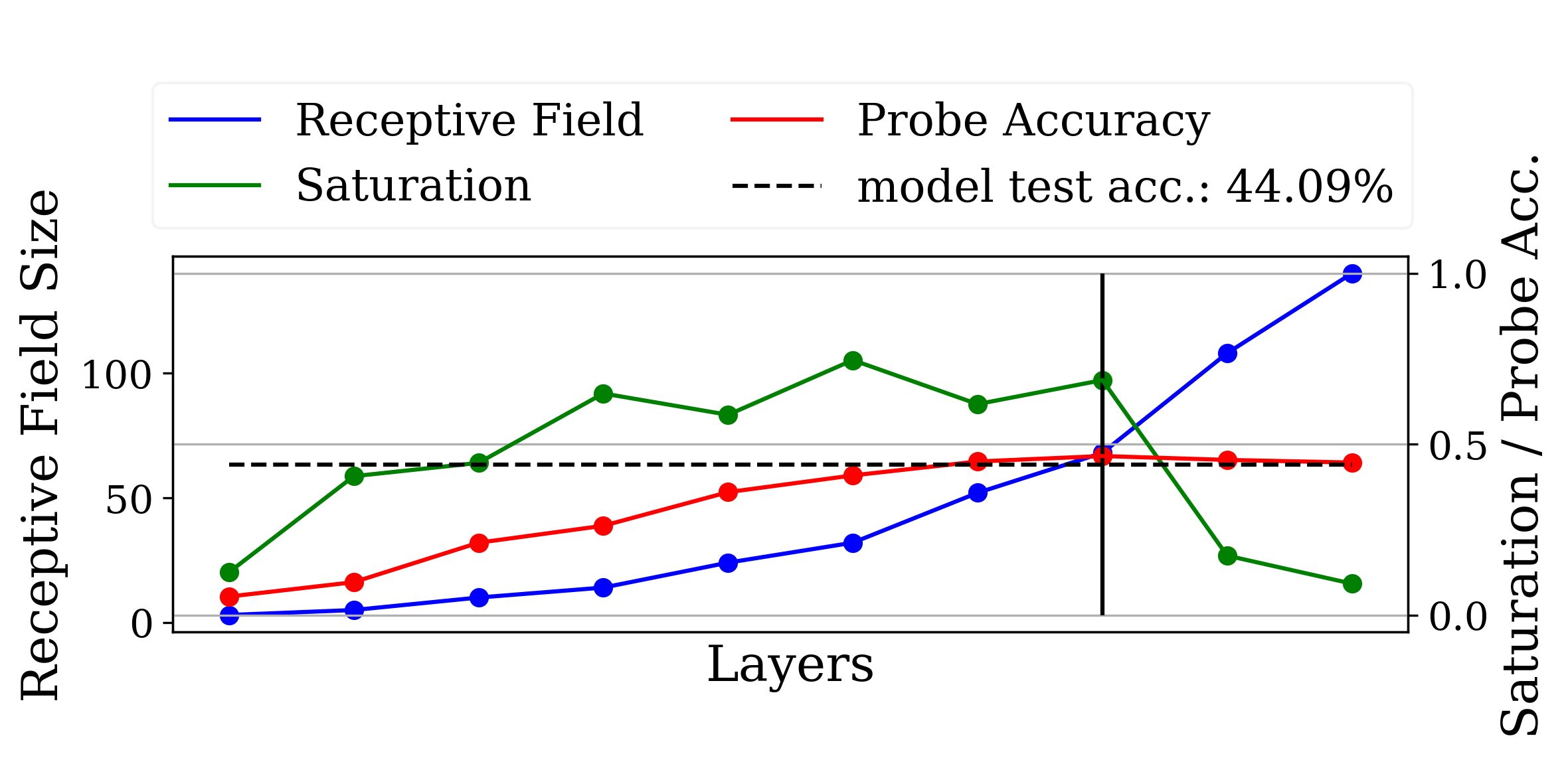}
	}\quad
	\subfloat[VGG16]{
	    \includegraphics[width=0.45\columnwidth]{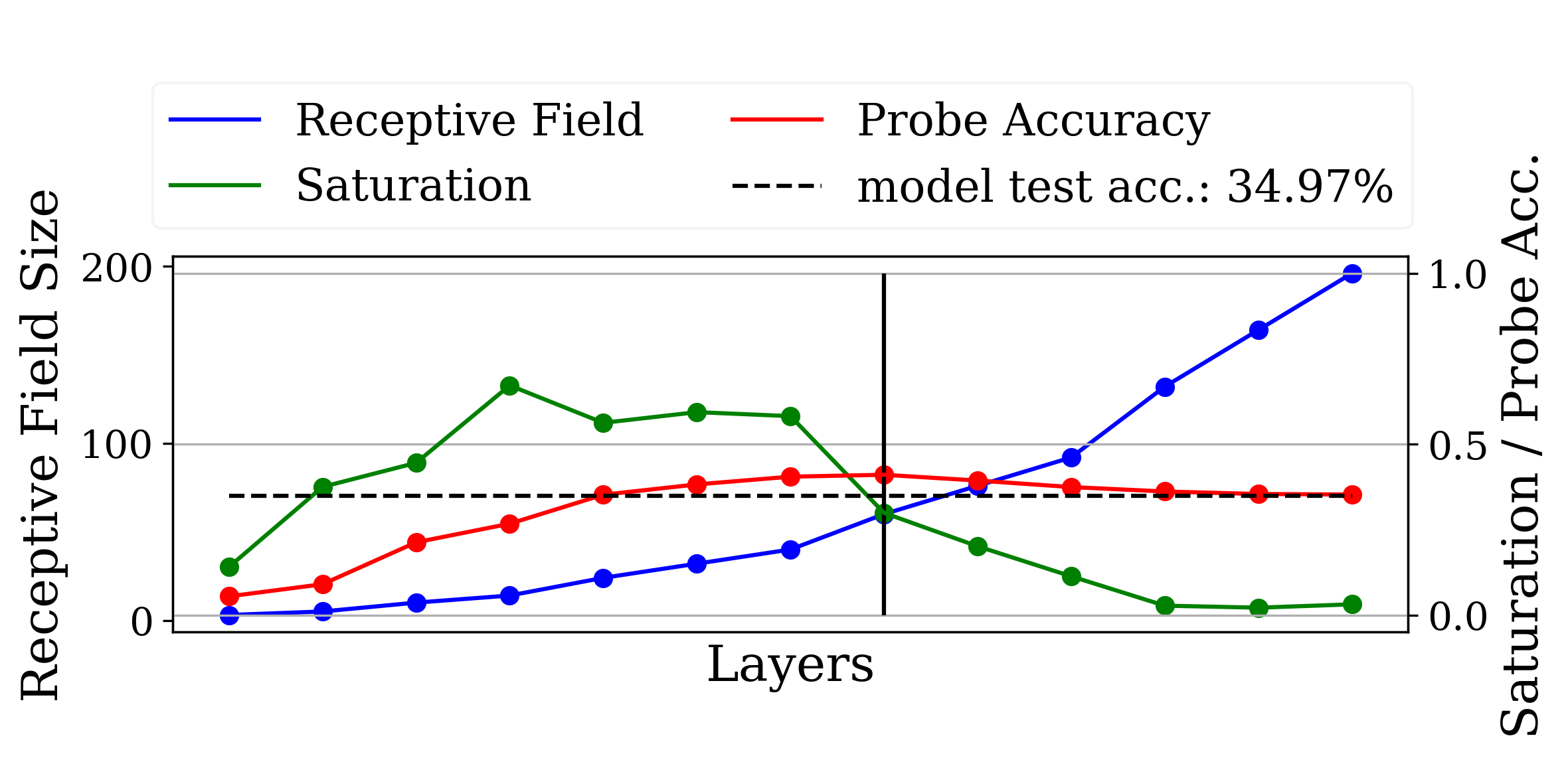}
	}
	\caption{The first convolutional layer processing input from a layer with a receptive field size greater than the input size marks a border between an "productive" and an "unproductive" part of the network. Productive meaning relatively high saturated and visible improving probe performance. The depicted models are from the VGG-family of networks by \citet{vgg} and trained on Cifar10 with an input size of $32 \times 32$ pixels.}
	\label{fig:vgg_receptive_field_TinyImageNet}
	
\end{figure}

\begin{figure}[htb!]
	\centering
	\subfloat[VGG11]{
	    \includegraphics[width=0.4\columnwidth]{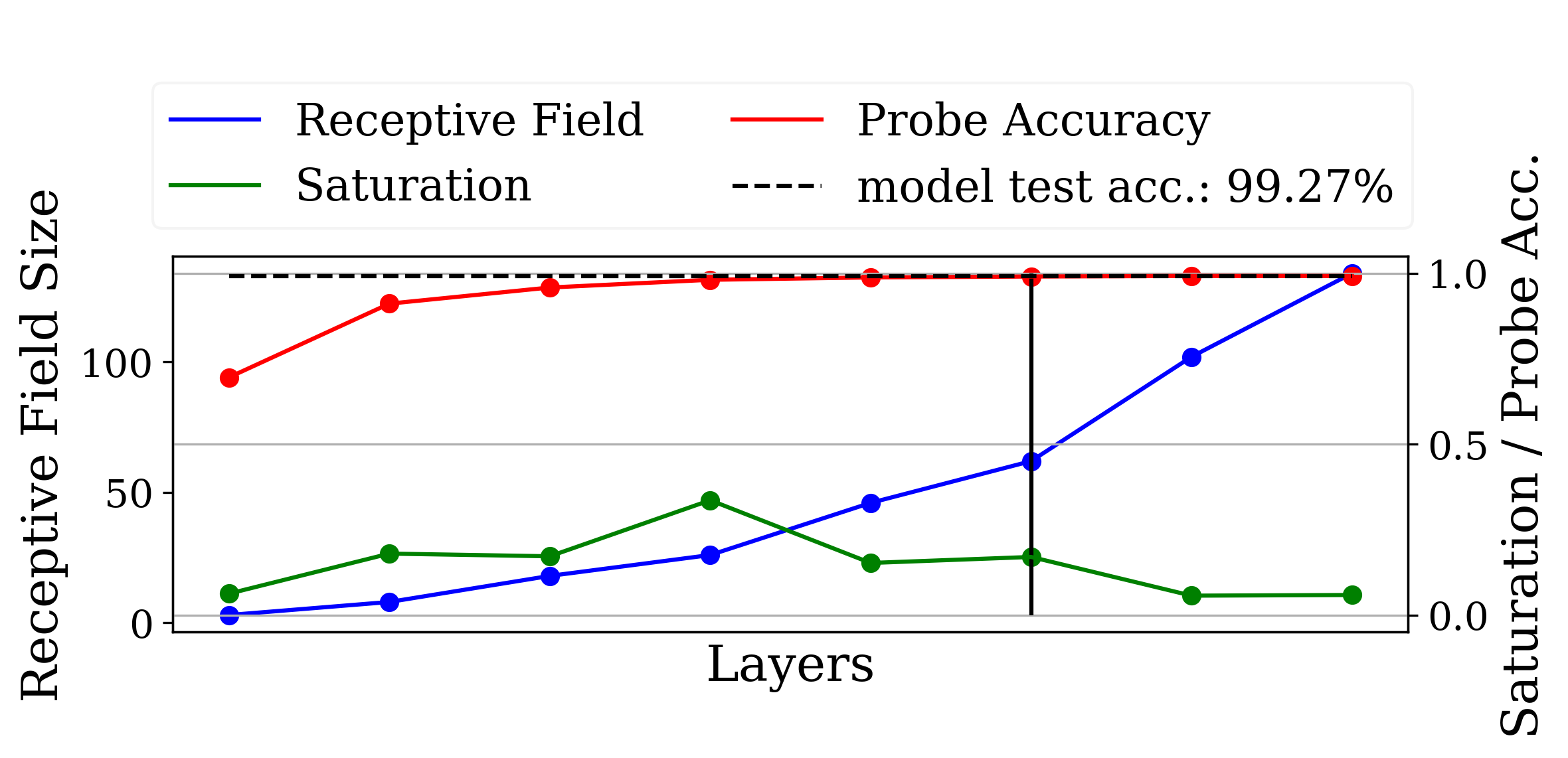}
	}\quad
		\subfloat[VGG13]{
	    \includegraphics[width=0.4\columnwidth]{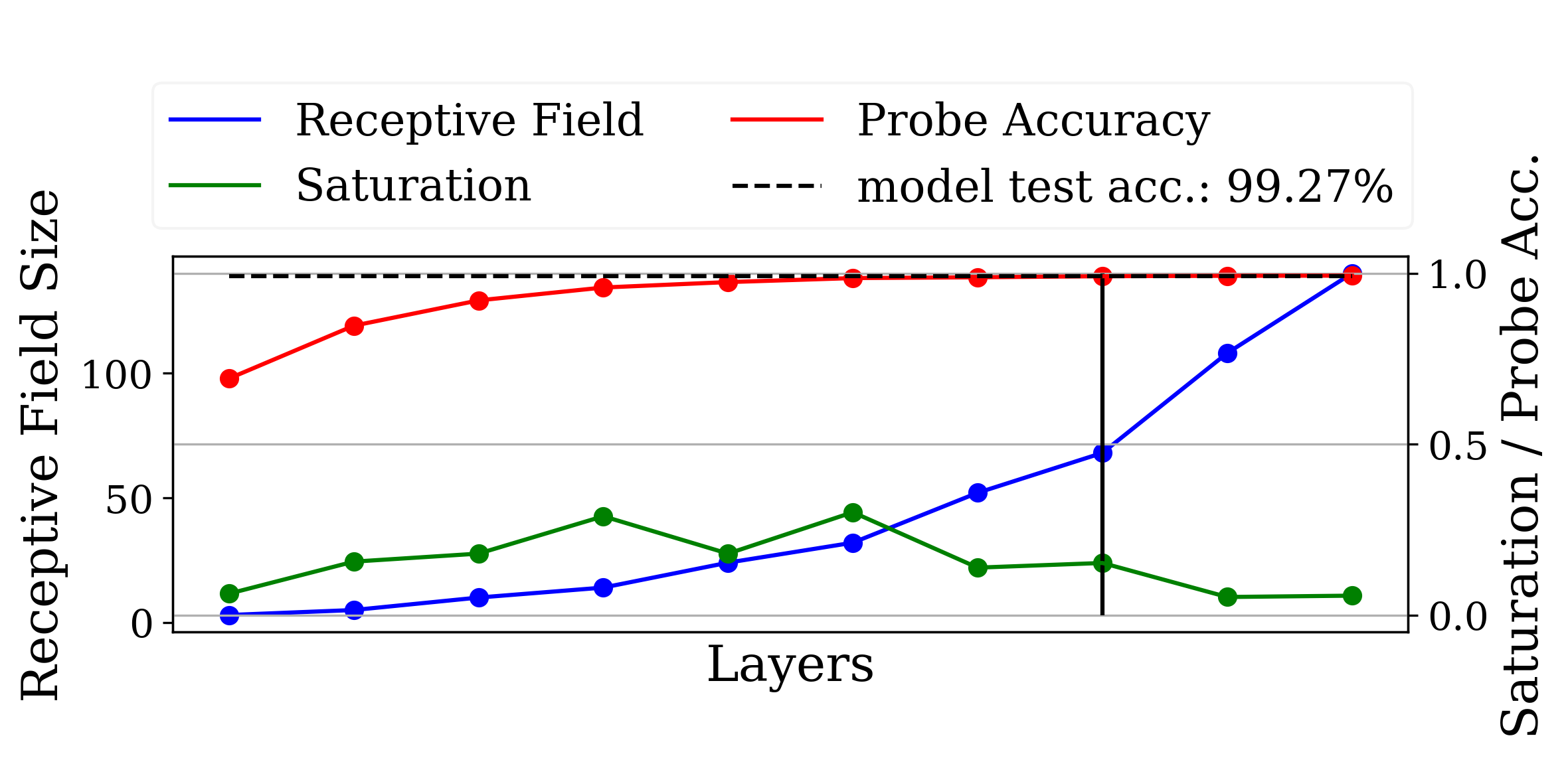}
	}\quad
	\subfloat[VGG16]{
	    \includegraphics[width=0.4\columnwidth]{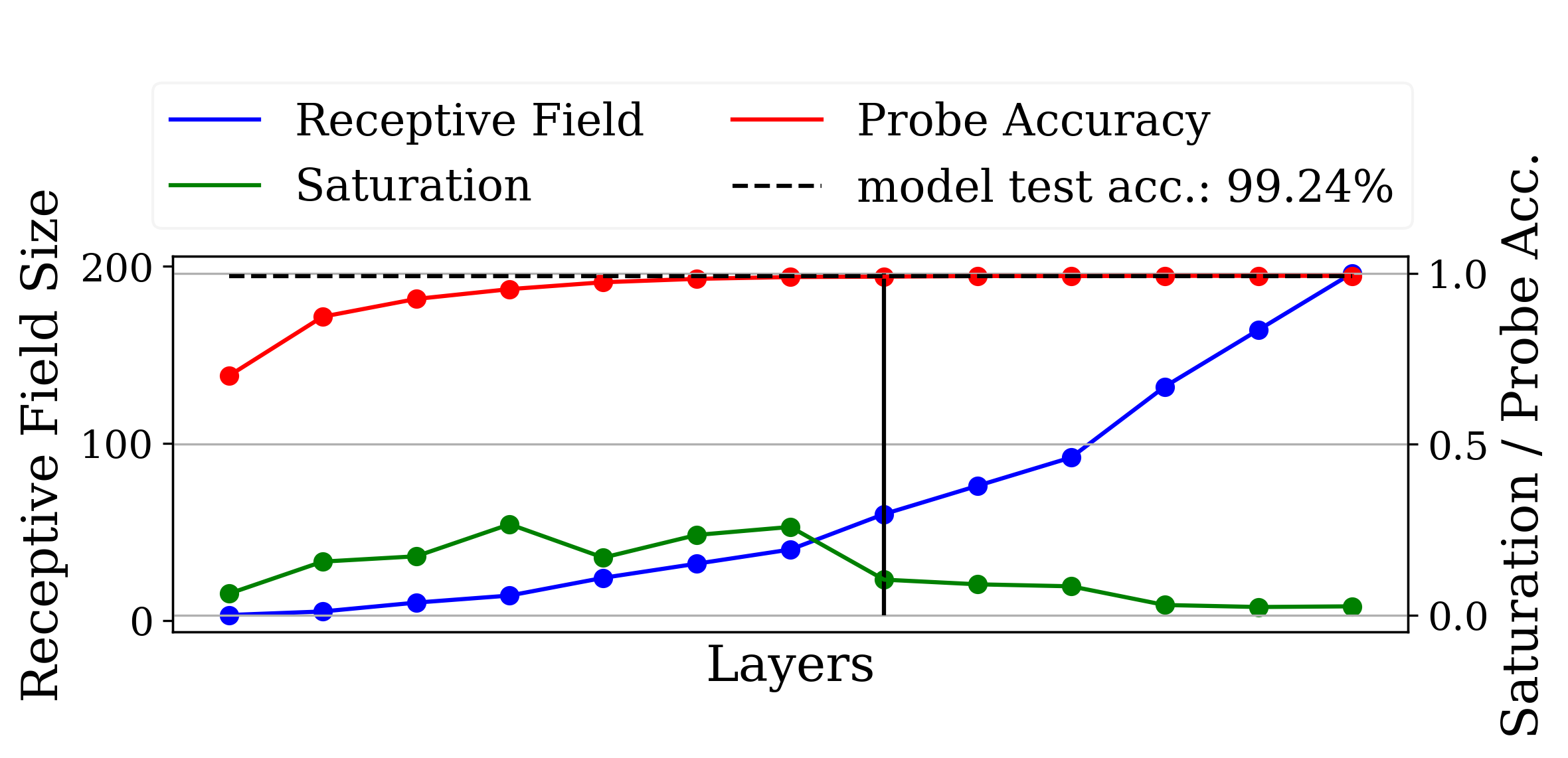}
	}\quad
	\subfloat[VGG19]{
	    \includegraphics[width=0.4\columnwidth]{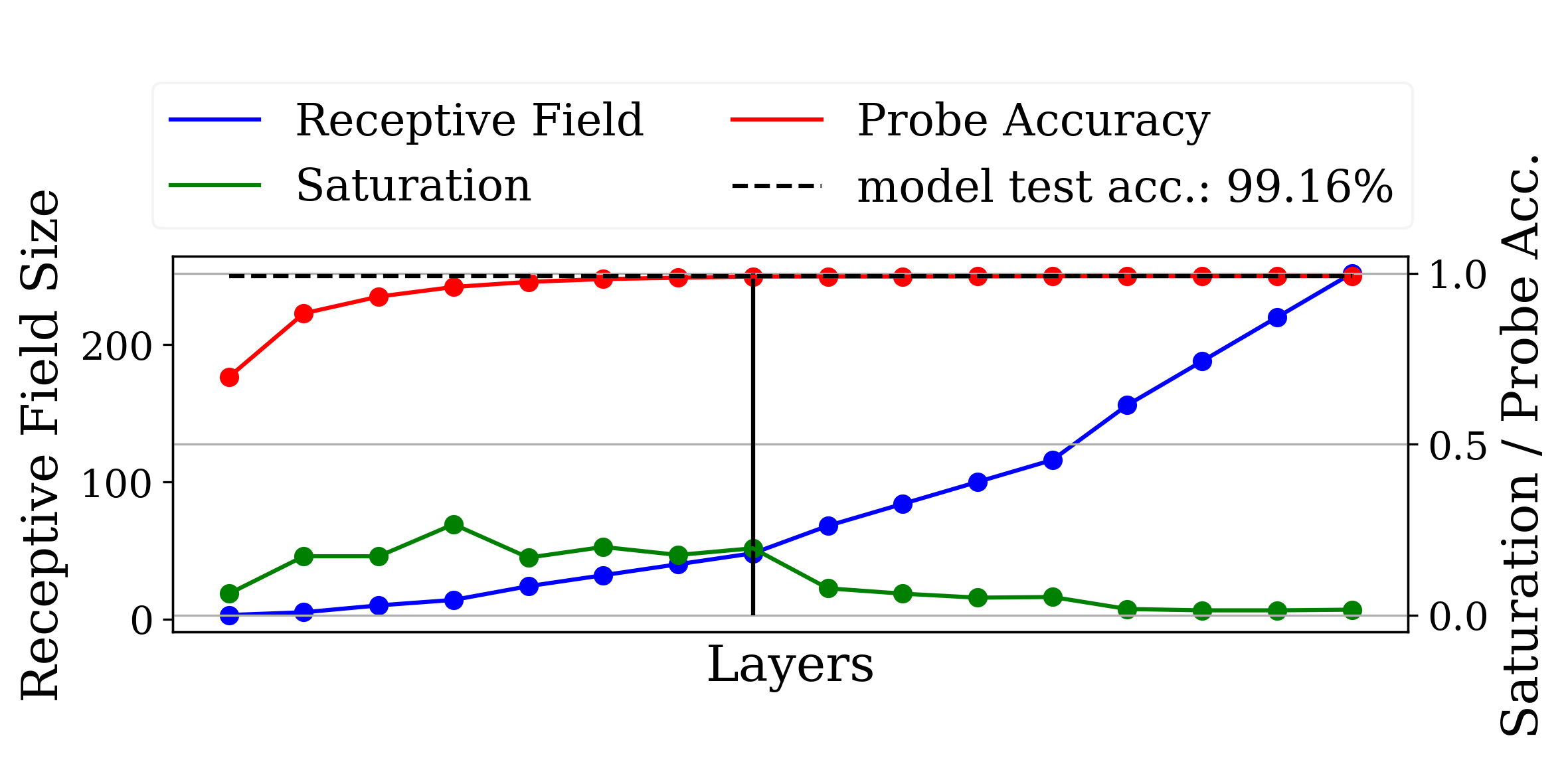}
	}
	\caption{The first convolutional layer processing input from a layer with a receptive field size greater than the input size marks a border between an "productive" and an "unproductive" part of the network. Productive meaning relatively high saturated and visible improving probe performance. The depicted models are from the VGG-family of networks by \citet{vgg} and trained on MNIST with an input size of $32 \times 32$ pixels.}
	\label{fig:vgg_receptive_field_MNIST}
	
\end{figure}

The following experiments show additional observations made regarding the role of the receptive field size.
When we training the models on the (arguably more difficult) TinyImageNet dataset using the same setup, the behavior stays consistent with our previous observations (see figure \ref{fig:vgg_receptive_field_TinyImageNet}).
However, if the problem is easy to solve it is possible for the model to mostly saturate probe performance before the saturation "tail" starts.
We can see an example of this in figure \ref{fig:vgg_receptive_field_MNIST}. When looking closely we can still see very slight improvements made until the border layers is reached. However, the improvements are hard to observe, which is also the reason why we only show MNIST results in the supplementary material.

We also observe that skip-connections have a statistically significant performance impact and that different combinations of residual connections distribute the inference process differently. To test this, we disabled different combinations of residual connections in ResNet18 while training Cifar10.

\subsection{TinyImageNet Results}

\FloatBarrier

\FloatBarrier
\clearpage

\subsection{MNIST results}

\FloatBarrier

\FloatBarrier
\clearpage

\subsection{ResNet18 Cifar10 - Effect of disabling residual connections}

Results indicate that removing the first three residual connections harms performance. Removing the last or second-to-last skip connection is statistically indistinguishable from not removing any residual connections. Furthermore, removing the first skip connection harms performance relative to many other permutations of residual connections. Thus, it seems like early downsampling decreases network performance and the first skip connection mitigates that effect. Later residual connections do not have the same effect.

{
\sethlcolor{green}

\begin{center}
\begin{table}[htb!]
\small
\begin{tabular}{ c c c c c c c c c c c c c c }
 & \textbf{2,3,4} & \textbf{1,3,4} & \textbf{1,2,4} & \textbf{1,2,3} & \textbf{3,4} & \textbf{2,4} & \textbf{2,3} & \textbf{1,4} & \textbf{1,3} & \textbf{1,2} & \textbf{4} & \textbf{1} & \textbf{ResNet18} \\
Mean & 82.70 & 83.17 & 83.40 & 83.39 & 82.67 & 82.35 & 82.59 & 83.42 & 81.88 & 82.95 & 81.65 & 82.92 & 83.16 \\
n & 10 & 10 & 10 & 10 & 10 & 10 & 10 & 10 & 10 & 9 & 9 & 8 & 8
\end{tabular}
\caption{ResNet18 with some residual connections removed. ResNet18 has four residual connections, \textbf{bold} numbers indicate which residual connections were enabled. For instance, \textbf{2, 3, 4} indicates that residual connections 2, 3 and 4 were present in the network. Mean performance of \textbf{n} Cifar10 trainings.}
\end{table}
\end{center}

\begin{center}
\begin{table}[htb!]
\small
\begin{tabular}{ c c c c c c c c c c c c c c }
 & \textbf{2,3,4} & \textbf{1,3,4} & \textbf{1,2,4} & \textbf{1,2,3} & \textbf{3,4} & \textbf{2,4} & \textbf{2,3} & \textbf{1,4} & \textbf{1,3} & \textbf{1,2} & \textbf{4} & \textbf{1} & \textbf{ResNet18} \\
\textbf{2,3,4} &  & 0.026 & \hl{0.004} & \hl{0.015} & 0.879 & 0.340 & 0.638 & \hl{0.001} & 0.563 & 0.256 & \hl{0.014} & 0.287 & 0.260 \\
\textbf{1,3,4} &-2.421 &   & 0.254 & 0.380 & \hl{0.018} & 0.028 & \hl{0.013} & 0.167 & 0.363 & 0.273 & \hl{0.001} & 0.177 & 0.983 \\
\textbf{1,2,4} &\hl{-3.297} & -1.178 &   & 0.963 & \hl{0.003} & \hl{0.008} & \hl{0.002} & 0.933 & 0.288 & 0.051 & \hl{0.000} & 0.030 & 0.562 \\
\textbf{1,2,3} &\hl{-2.701} & -0.901 & 0.047 &   & \hl{0.011} & \hl{0.014} & \hl{0.008} & 0.907 & 0.294 & 0.113 & \hl{0.001} & 0.087 & 0.609 \\
\textbf{3,4} &0.154 & \hl{2.615} & \hl{3.480} & \hl{2.847} &   & 0.385 & 0.740 & \hl{0.001} & 0.579 & 0.198 & \hl{0.016} & 0.219 & 0.228 \\
\textbf{2,4} &0.981 & 2.387 & \hl{2.960} & \hl{2.720} & 0.891 &   & 0.520 & \hl{0.006} & 0.746 & 0.121 & 0.169 & 0.149 & 0.130 \\
\textbf{2,3} &0.479 & \hl{2.761} & \hl{3.570} & \hl{2.989} & 0.337 & -0.656 &   & \hl{0.001} & 0.617 & 0.135 & 0.028 & 0.153 & 0.179 \\
\textbf{1,4} &\hl{-3.746} & -1.438 & -0.085 & -0.119 & \hl{-3.958} & \hl{-3.113} & \hl{-3.986} &   & 0.282 & 0.026 & \hl{0.000} & \hl{0.011} & 0.522 \\
\textbf{1,3} &0.589 & 0.933 & 1.095 & 1.080 & 0.565 & 0.329 & 0.508 & 1.109 &   & 0.475 & 0.879 & 0.513 & 0.432 \\
\textbf{1,2} &-1.176 & 1.132 & 2.098 & 1.673 & -1.341 & -1.632 & -1.569 & 2.447 & -0.731 &   & \hl{0.005} & 0.884 & 0.617 \\
\textbf{4} &\hl{2.750} & \hl{4.126} & \hl{4.613} & \hl{4.264} & \hl{2.671} & 1.437 & 2.402 & \hl{4.819} & 0.154 & \hl{3.291} &   & \hl{0.007} & \hl{0.014} \\
\textbf{1} &-1.102 & 1.412 & 2.384 & 1.821 & -1.279 & -1.515 & -1.502 & \hl{2.888} & -0.670 & 0.149 & \hl{-3.136} &   & 0.579 \\
\textbf{ResNet18} &-1.167 & 0.021 & 0.592 & 0.522 & -1.253 & -1.595 & -1.407 & 0.655 & -0.805 & -0.510 & \hl{-2.779} & -0.568 &  

\end{tabular}
\caption{Independent two-sample t-test, pairwise mean performance difference significance test, ResNet18 with some residual connections removed, trained on Cifar10. ResNet18 has four skip-connections, \textbf{bold} numbers indicate which residual connections were enabled. For instance, \textbf{2, 3, 4} indicates that residual connections 2, 3 and 4 were present in the network. P-values are in the upper triangle, t-values in the lower. Statistically significant values (p<0.25) are highlighted in \hl{green}. n=10 or as in the previous table.}
\end{table}
\end{center}
}
\FloatBarrier
\clearpage

\section{Receptive Field Analysis with Partial Solution Heatmaps}
\subsection{Receptive Field Analysis using Probes on VGG16}
\FloatBarrier

\begin{figure}[htb!]
	\centering
	\subfloat[VGG16 trained on Cifar10 using $32 \times 32$ pixel input size]{
	    \includegraphics[width=0.98\columnwidth]{figures/old_results/VGG16_Cifar10.png}
	} \quad
	\subfloat[Conv1]{
	    \includegraphics[width=0.08\columnwidth]{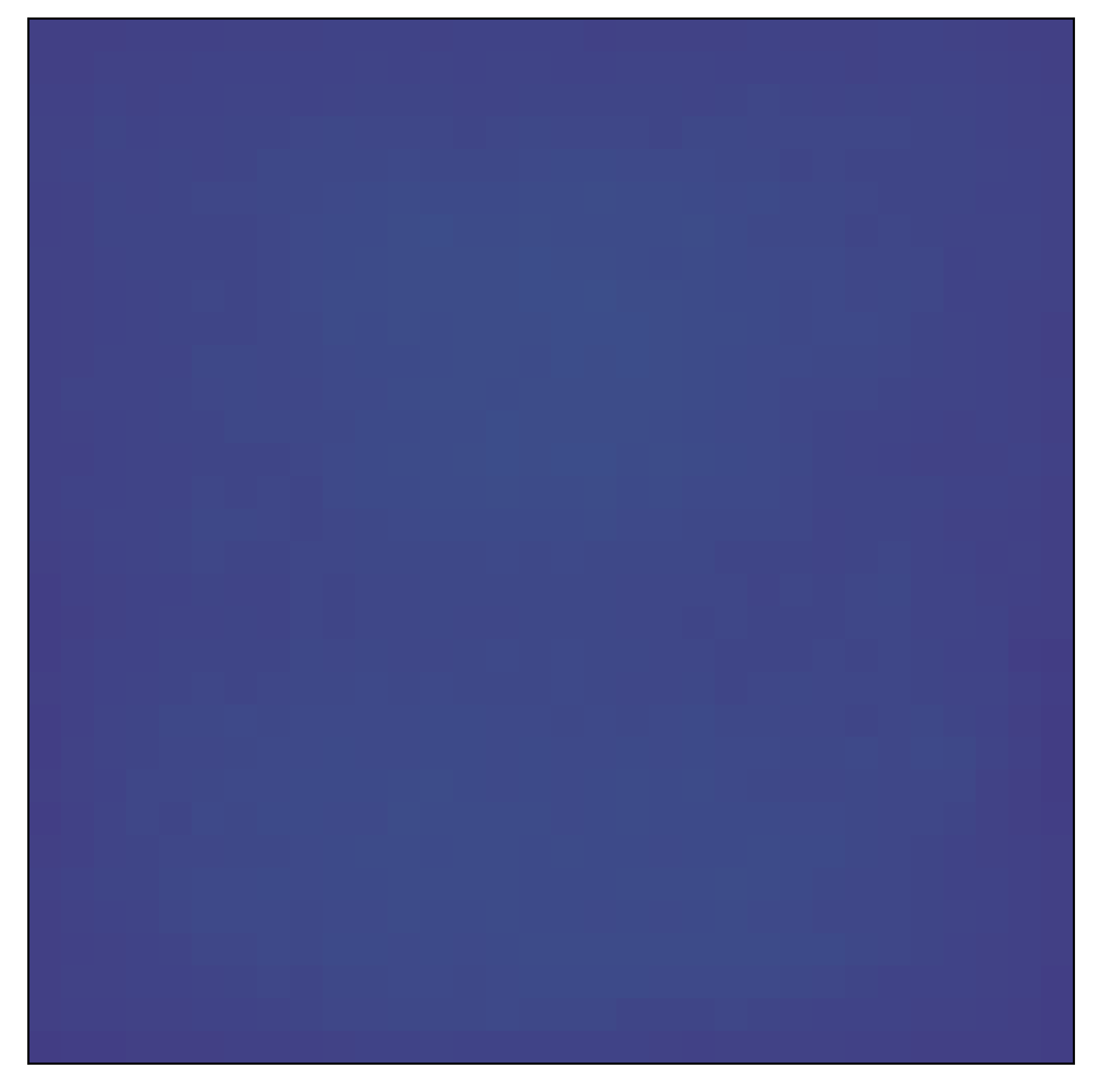}
	}
	\subfloat[Conv2]{
	    \includegraphics[width=0.08\columnwidth]{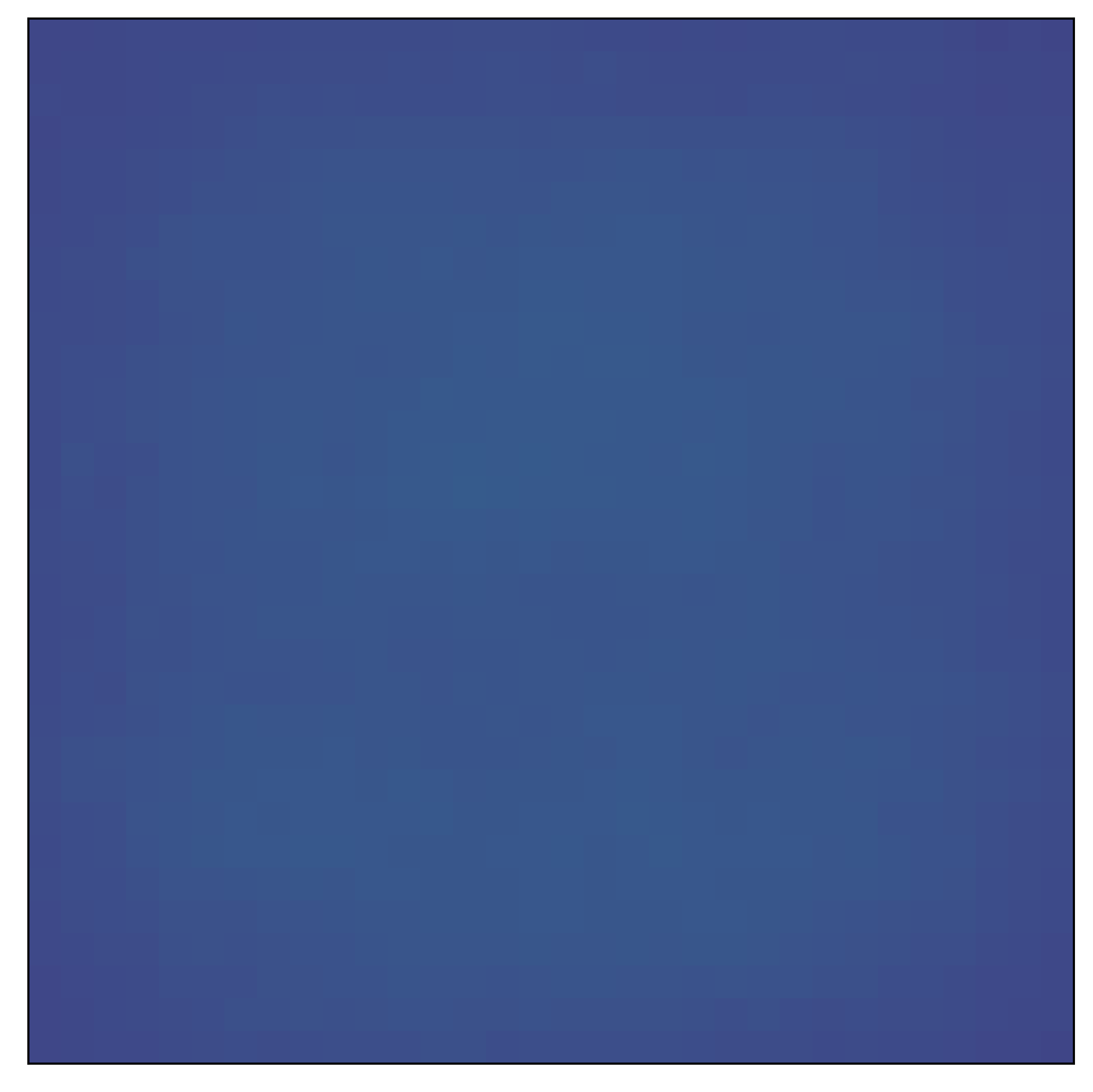}
	}
	\subfloat[Conv3]{
	    \includegraphics[width=0.08\columnwidth]{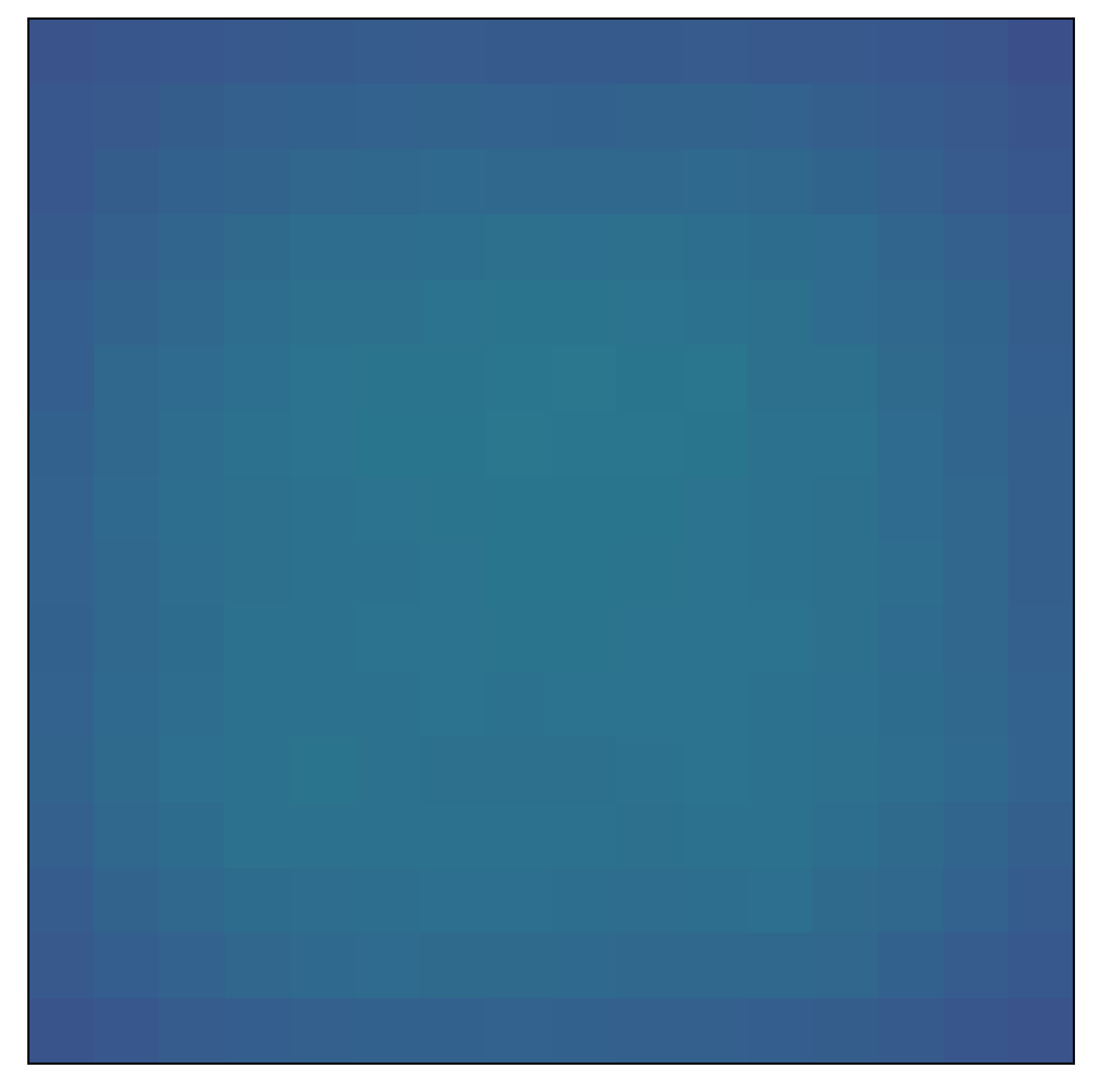}
	}
	\subfloat[Conv4]{
	    \includegraphics[width=0.08\columnwidth]{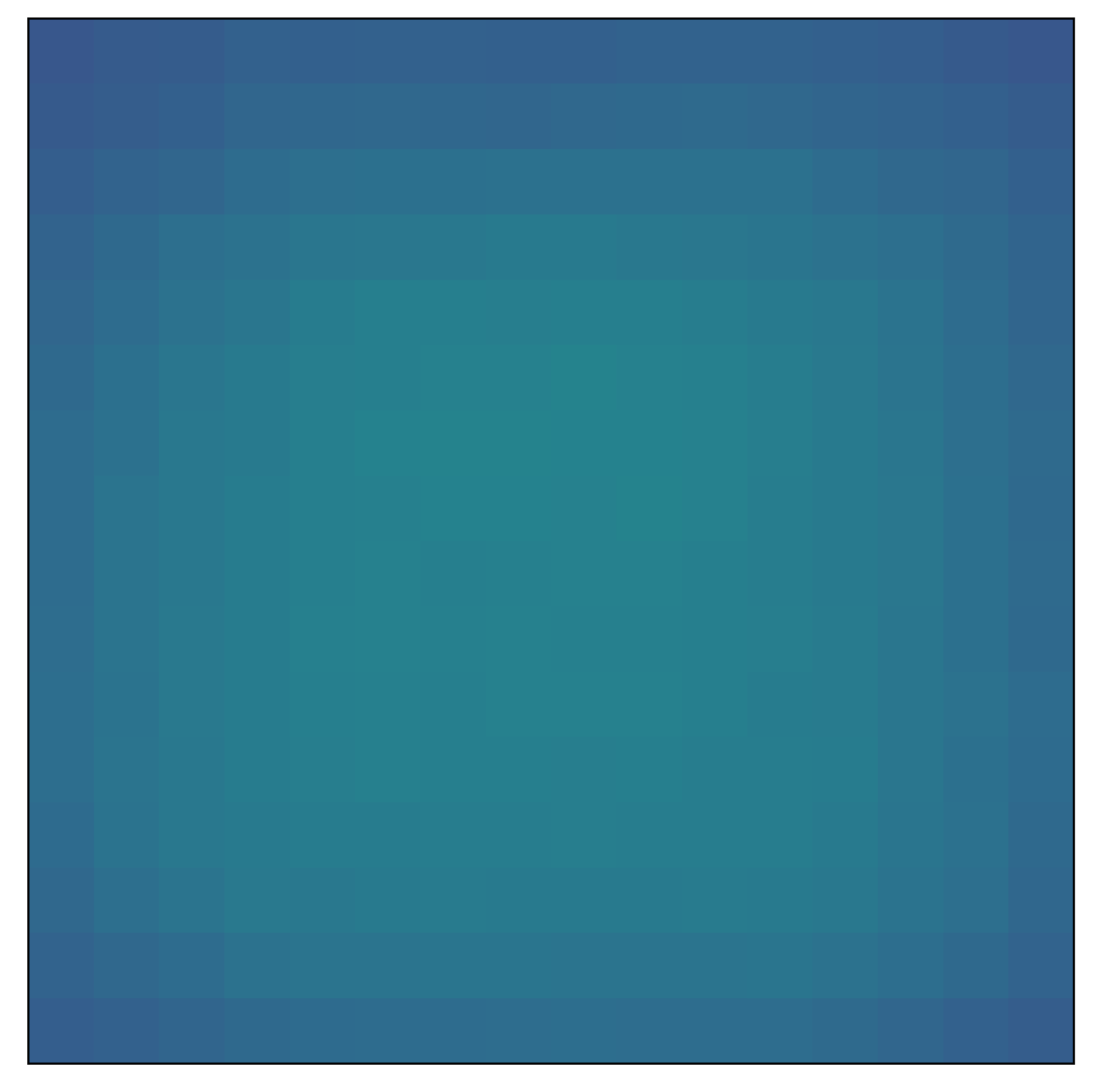}
	}
	\subfloat[Conv5]{
	    \includegraphics[width=0.08\columnwidth]{figures/receptive_field_feature_maps/features-14.png}
	}
	\subfloat[Conv6]{
	    \includegraphics[width=0.08\columnwidth]{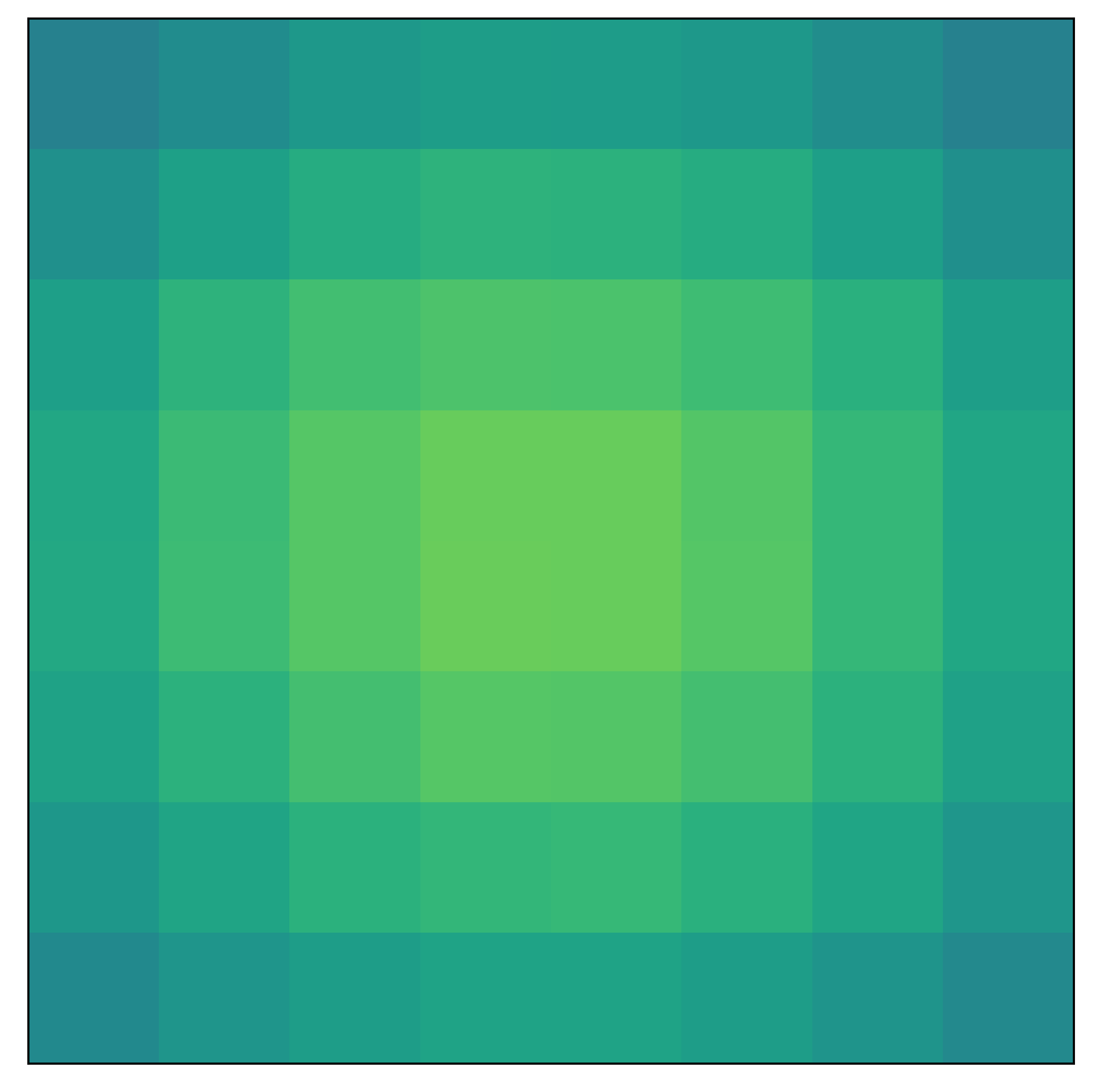}
	}
	\subfloat[Conv7]{
	    \includegraphics[width=0.08\columnwidth]{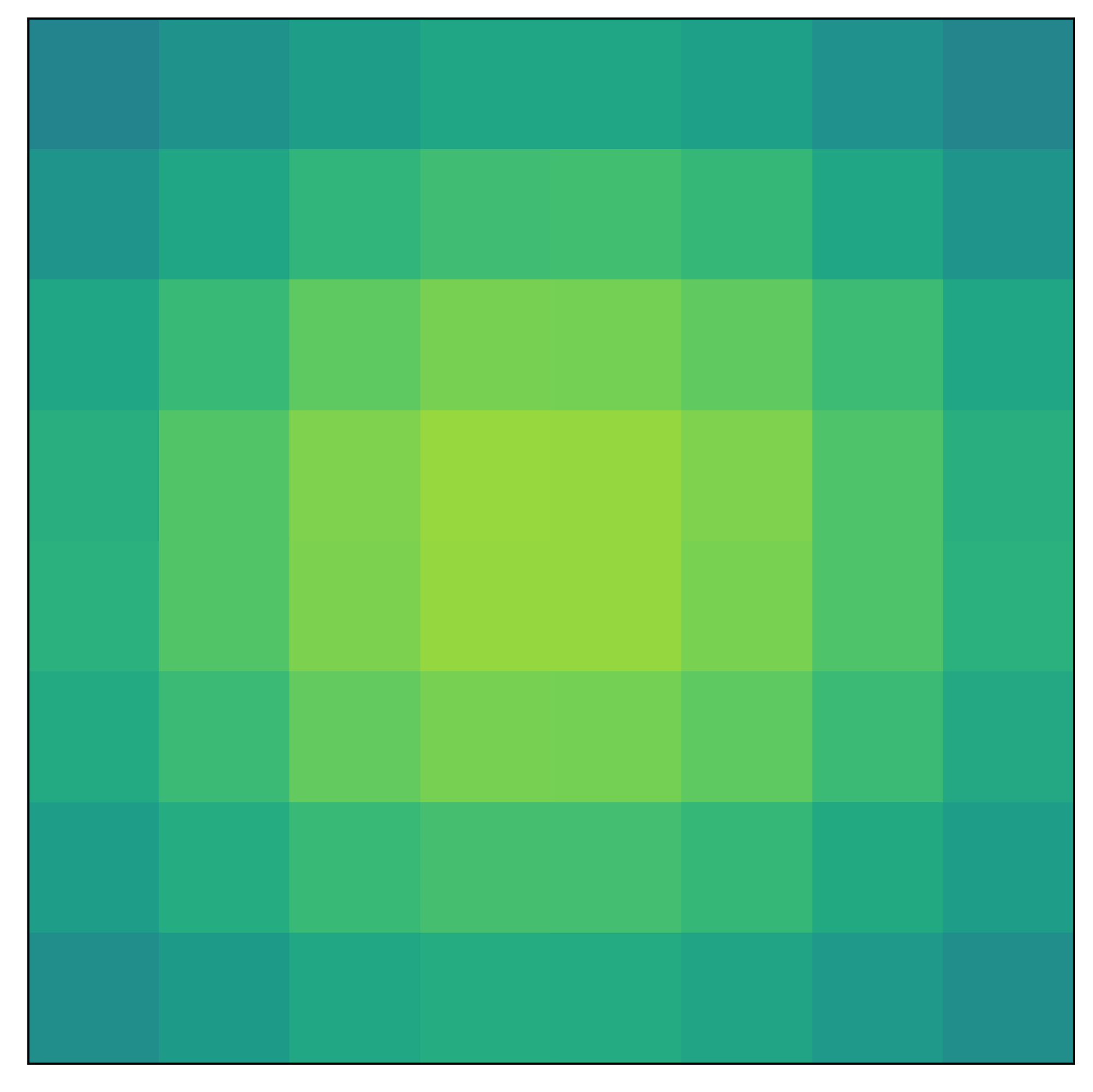}
	} \quad
	\subfloat[Conv8]{
	    \includegraphics[width=0.08\columnwidth]{figures/receptive_field_feature_maps/features-24.png}
	}
	\subfloat[Conv9]{
	    \includegraphics[width=0.08\columnwidth]{figures/receptive_field_feature_maps/features-27.png}
	}
	\subfloat[Conv10]{
	    \includegraphics[width=0.08\columnwidth]{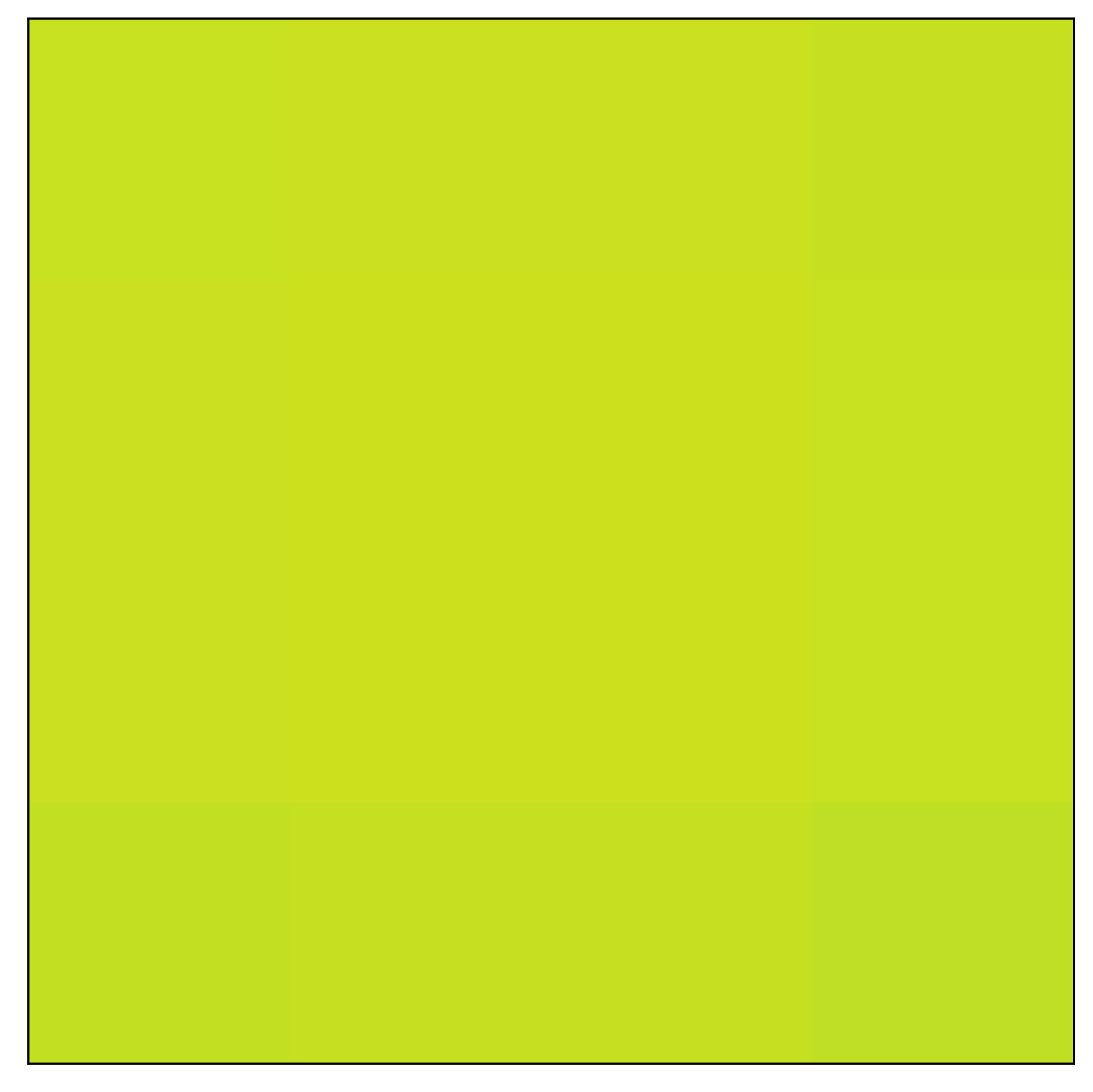}
	}
	\subfloat[Conv11]{
	    \includegraphics[width=0.08\columnwidth]{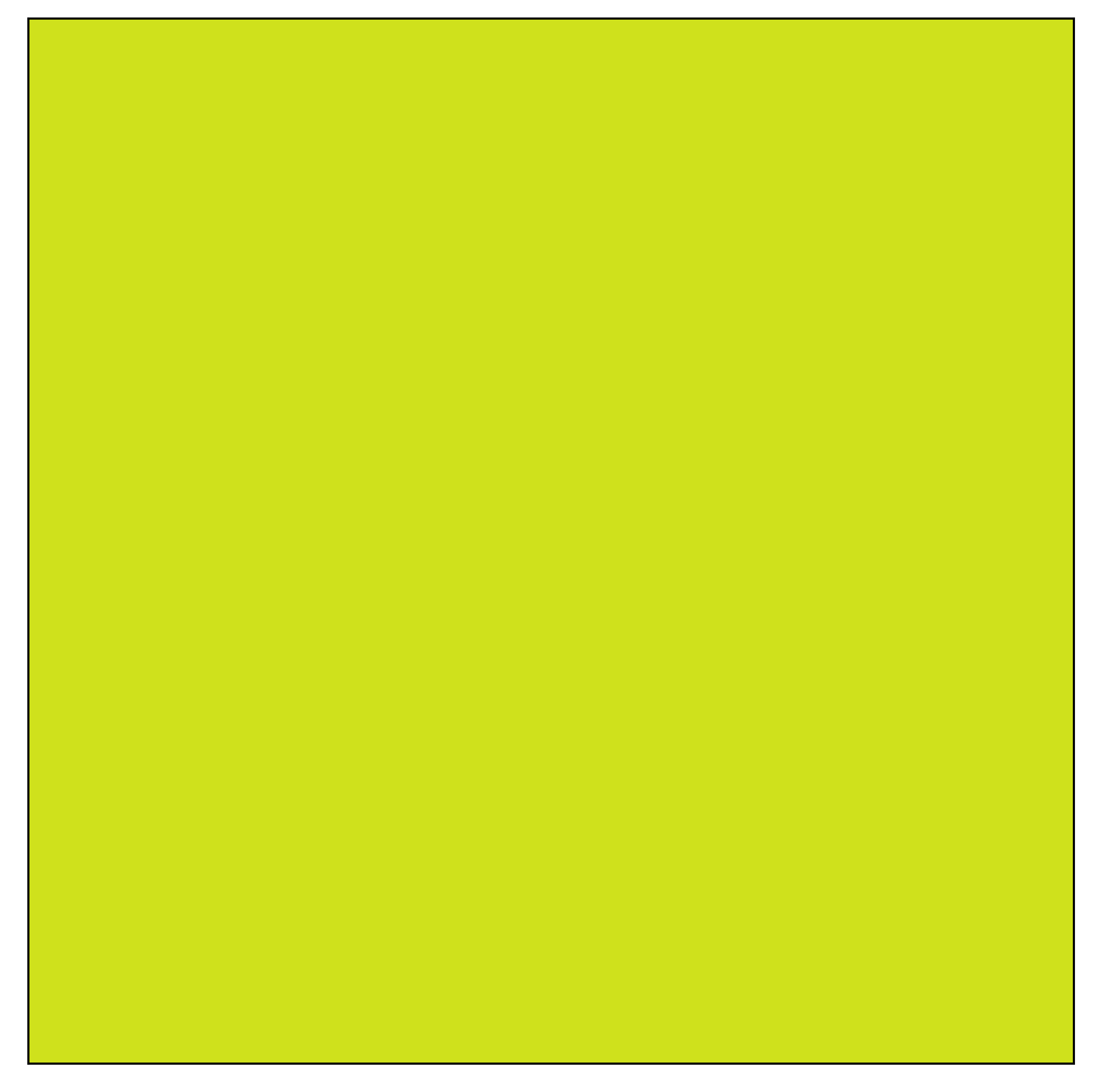}
	}
	\subfloat[Conv12]{
	    \includegraphics[width=0.08\columnwidth]{figures/receptive_field_feature_maps/features-37.png}
	}
	\subfloat[Conv13]{
	    \includegraphics[width=0.08\columnwidth]{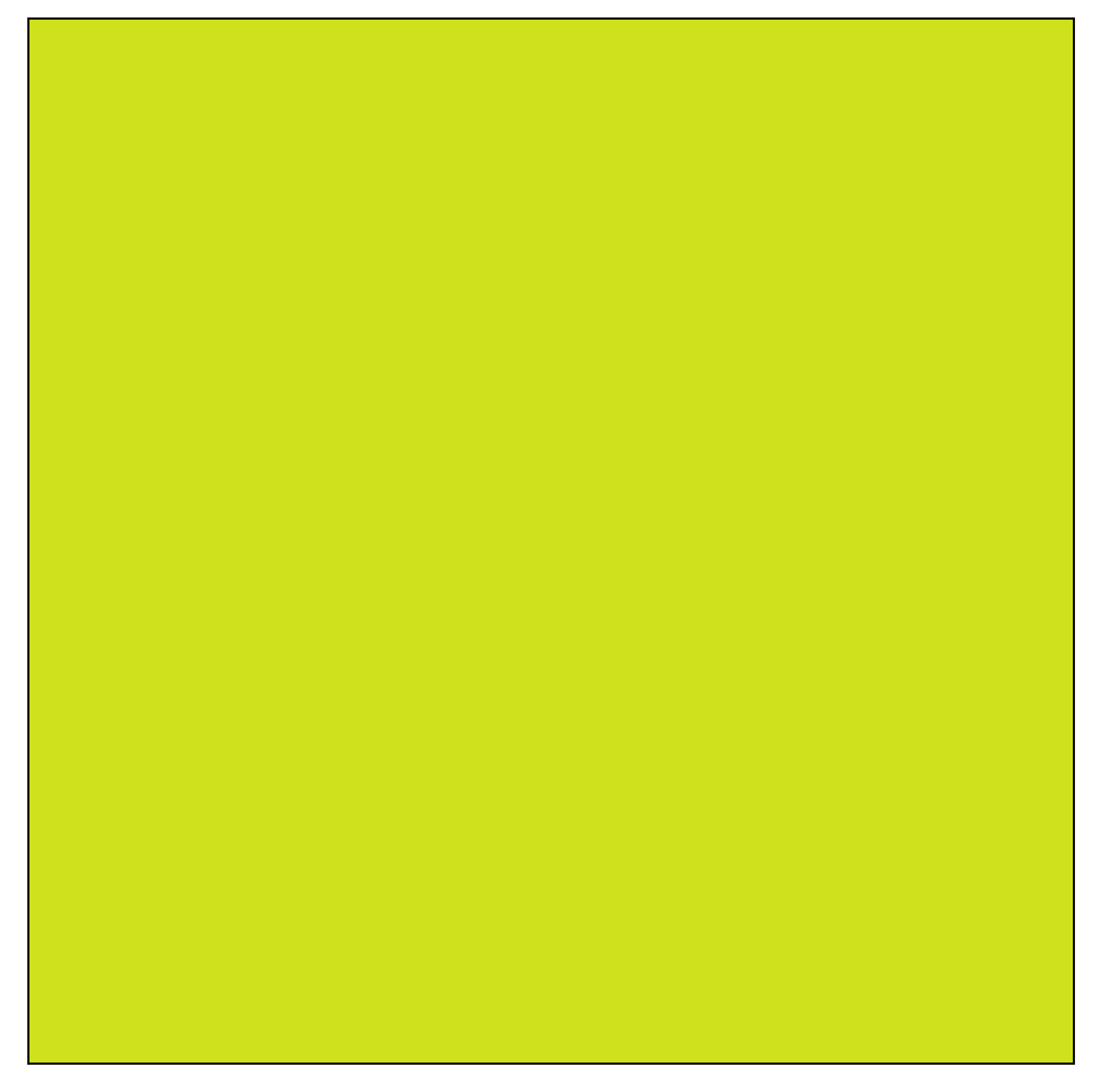}
	} \quad
	\subfloat{
	    \includegraphics[width=0.8\columnwidth]{figures/receptive_field_feature_maps/colorbar.png}
	}
	\caption{VGG16 trained on Cifar10 alongside the heatmaps generated from the relative accuracy of the partial solutions in each layer. Note that the border layer (Conv8) is the first layer to have partial solutions of equal quality to the models solution (measured in accuracy).}
	\label{fig:vgg_probe_heatmaps2a}

\end{figure}
\FloatBarrier
\clearpage

\subsection{Receptive Field Analysis using Probes on ResNet18}
\FloatBarrier

\begin{figure}[htb!]
	\centering
	\subfloat[ResNet18 trained on Cifar10 using $32 \times 32$ pixel input size]{
	    \includegraphics[width=0.98\columnwidth]{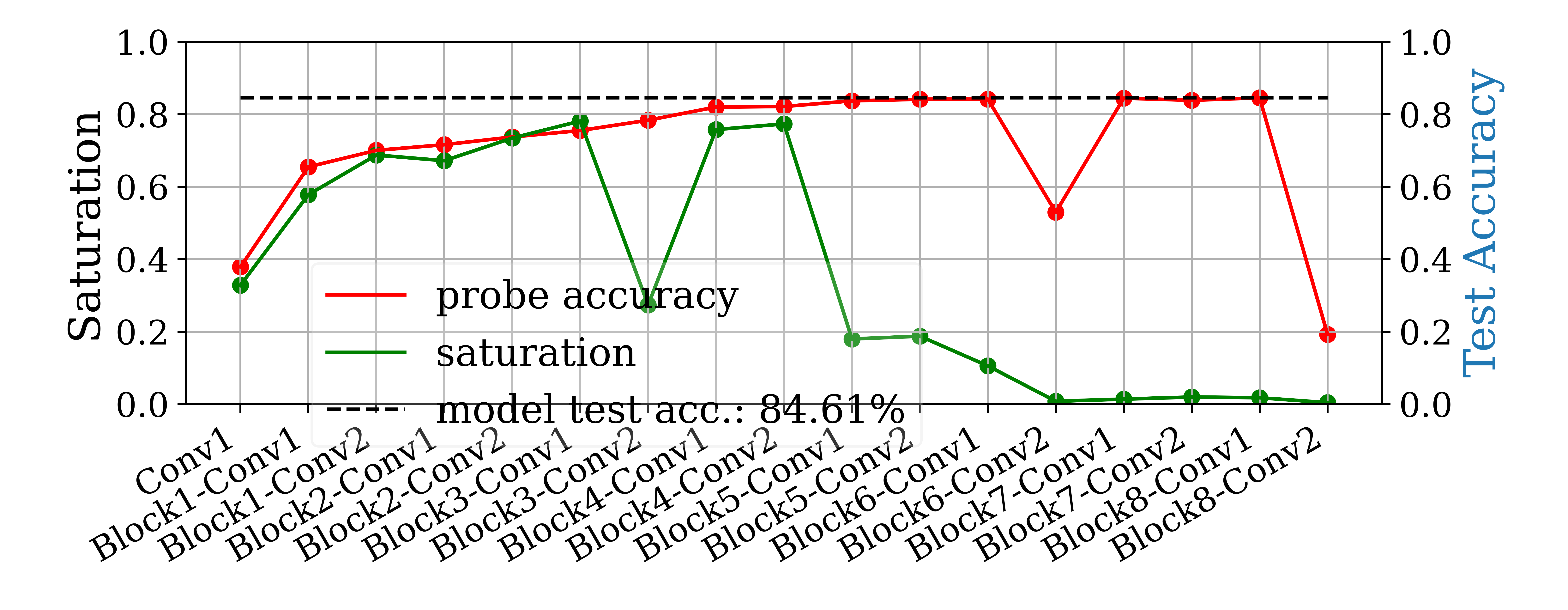}
	} \quad
	\subfloat[Conv1]{
	    \includegraphics[width=0.15\columnwidth]{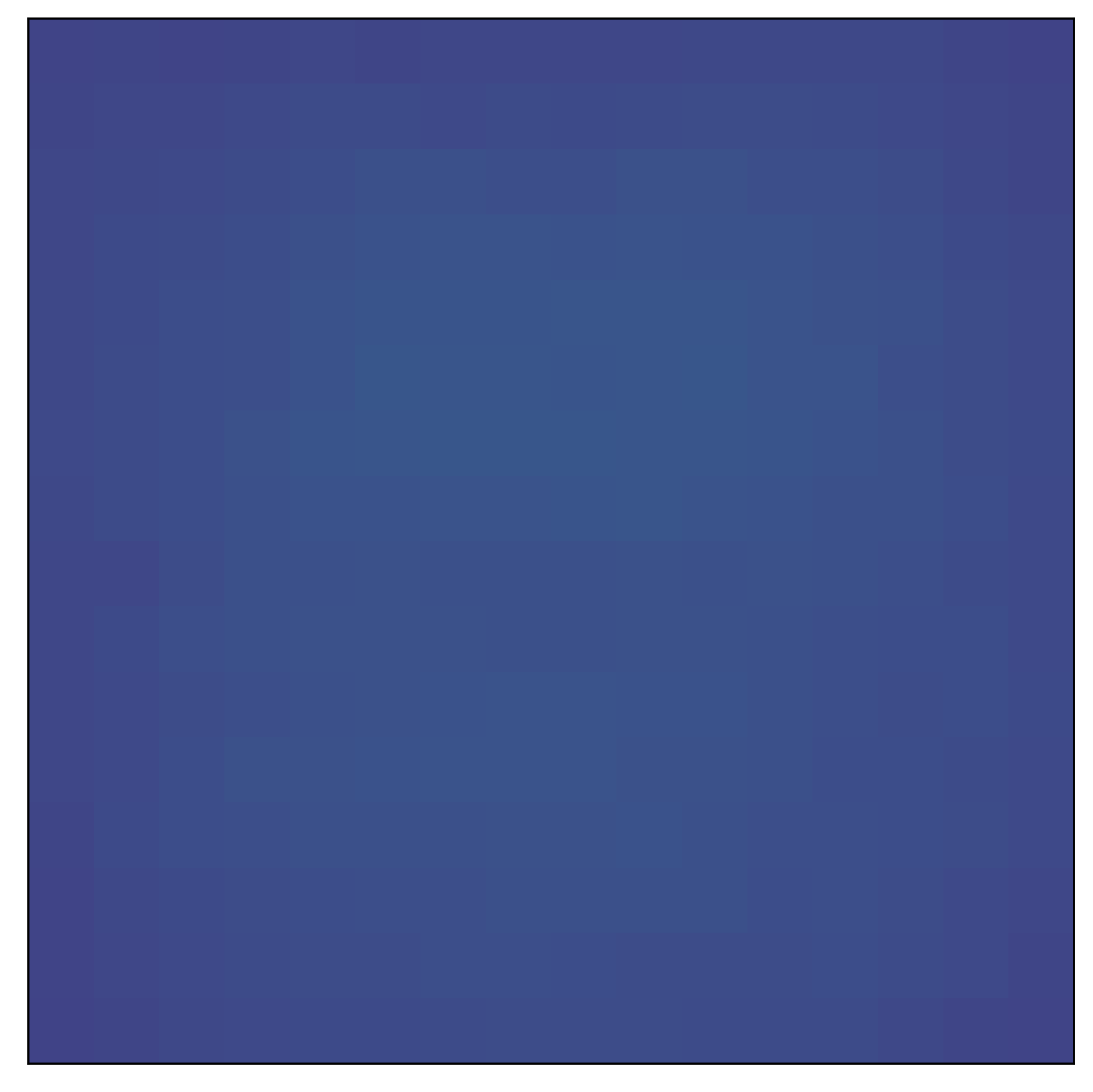}
	}
	\subfloat[Block1-Conv1]{
	    \includegraphics[width=0.15\columnwidth]{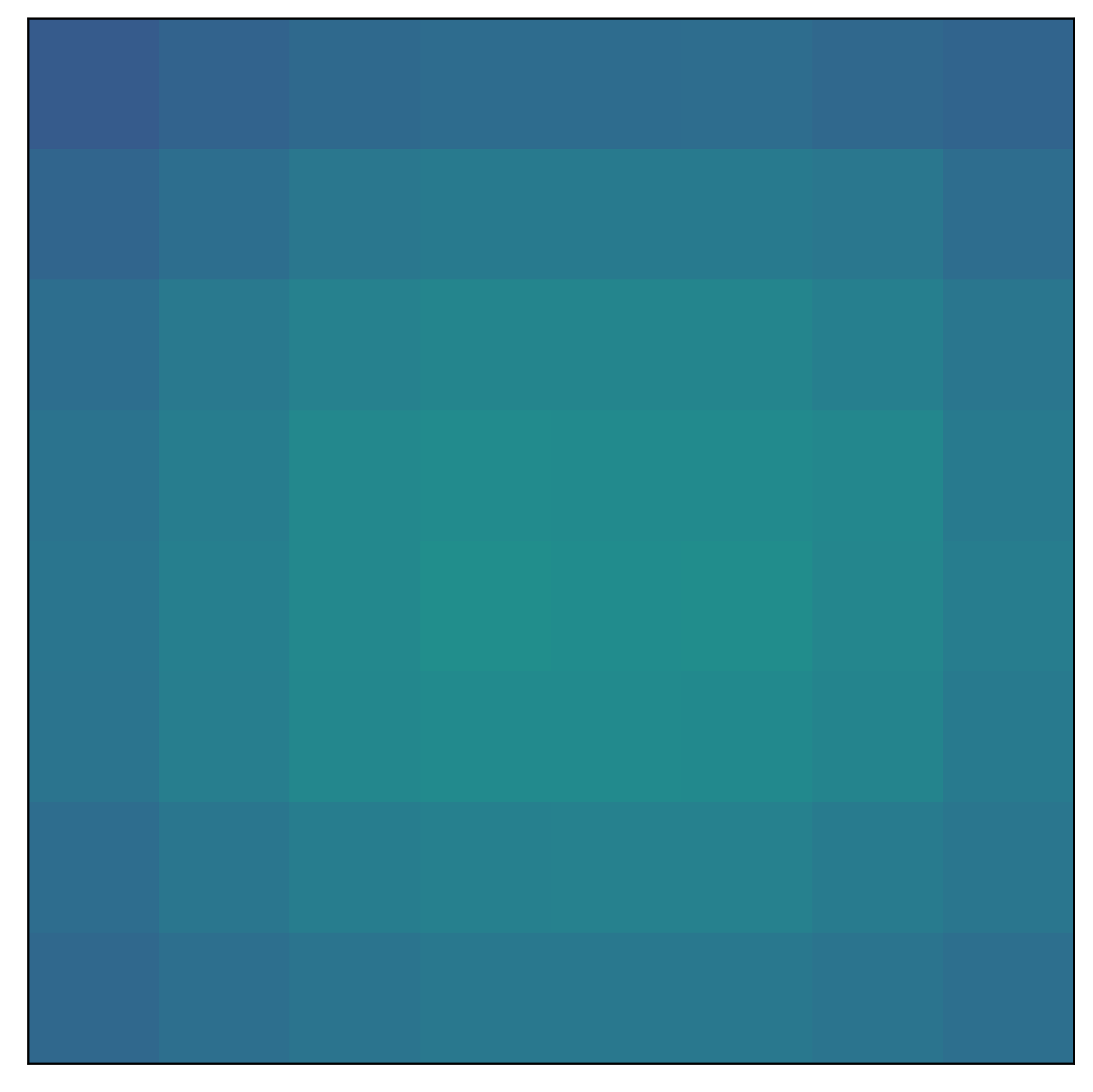}
	}
	\subfloat[Block1-Conv2]{
	    \includegraphics[width=0.15\columnwidth]{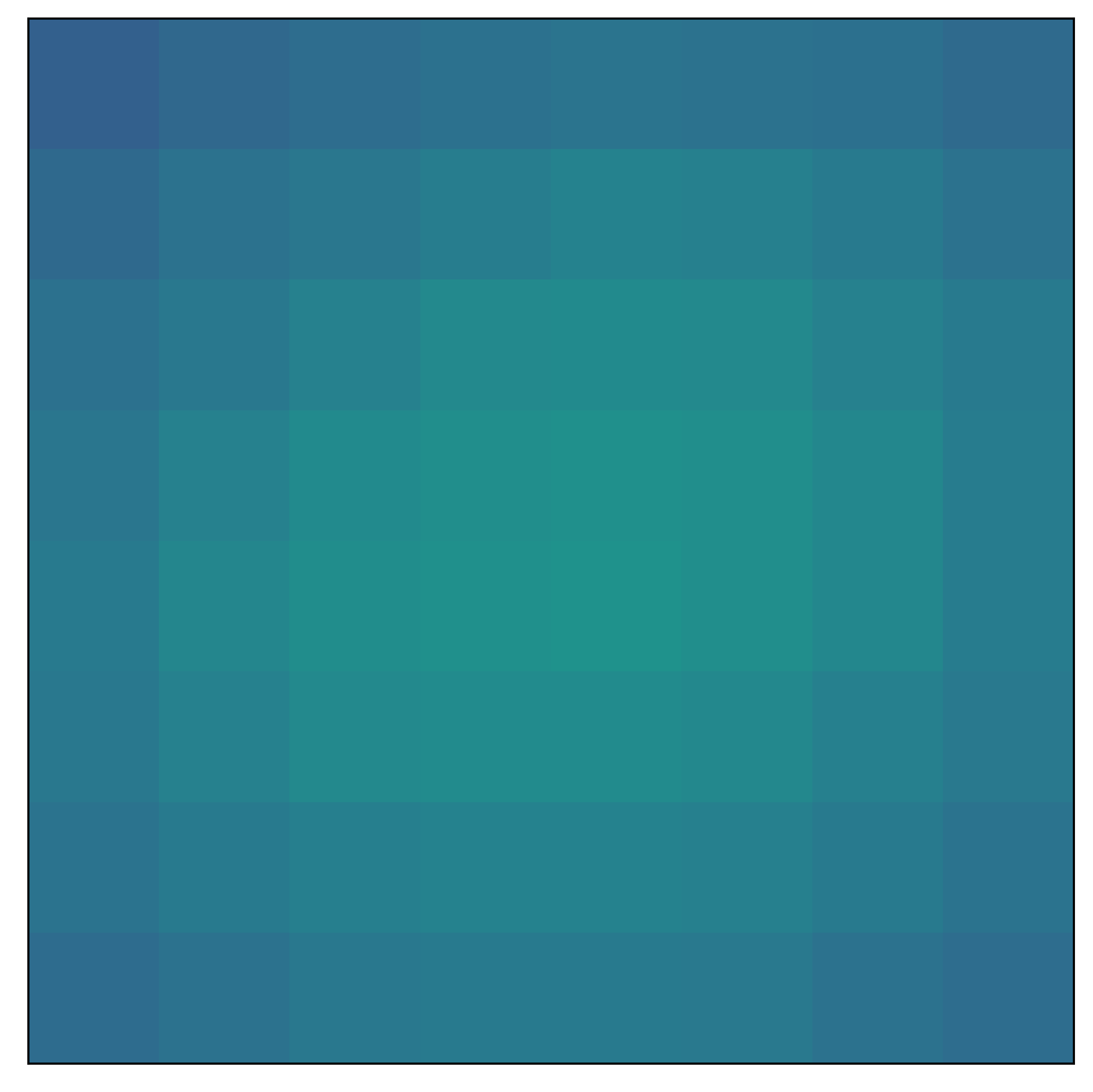}
	}
	\subfloat[Block2-Conv1]{
	    \includegraphics[width=0.15\columnwidth]{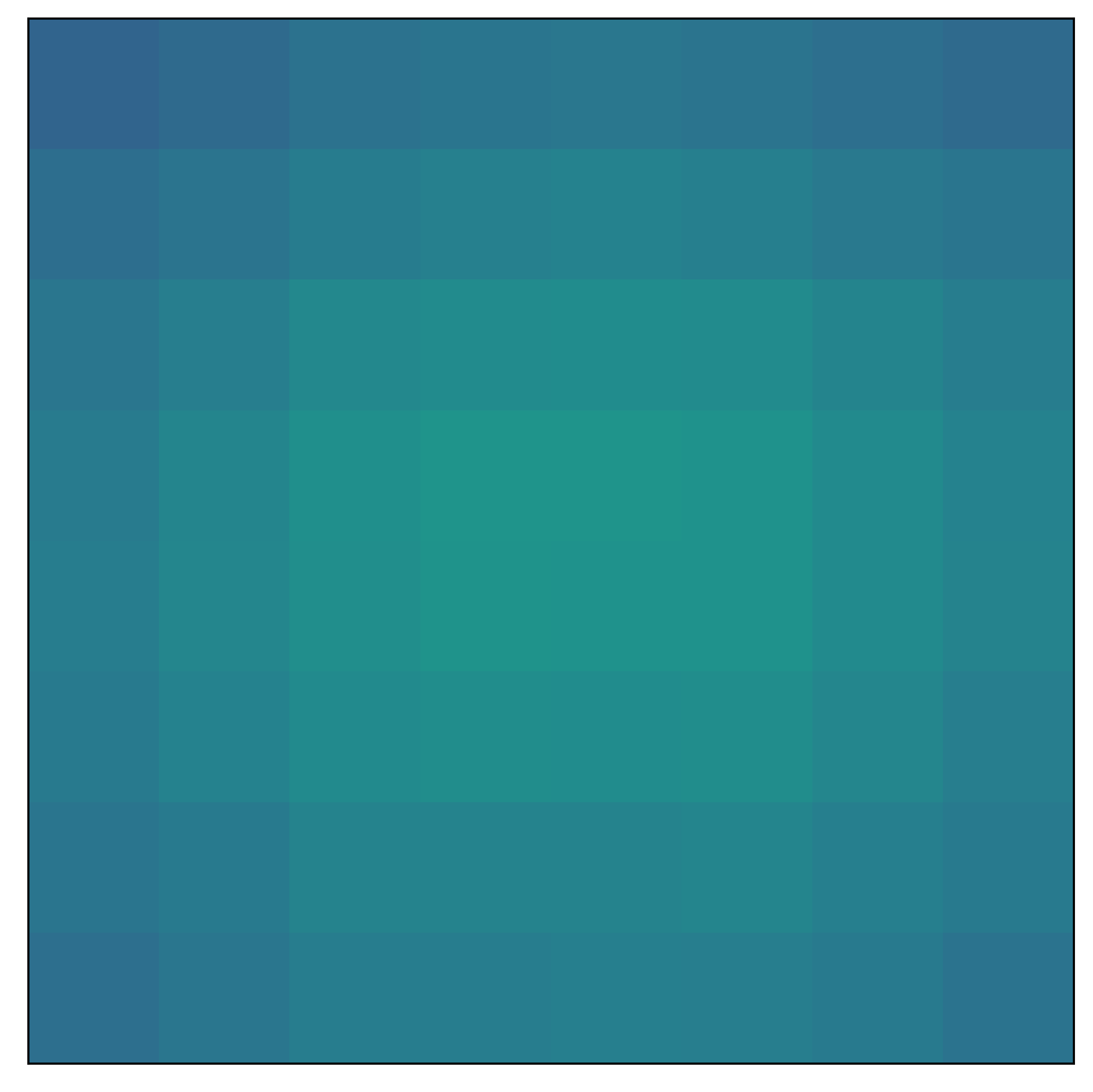}
	}
	\subfloat[Block2-Conv2]{
	    \includegraphics[width=0.15\columnwidth]{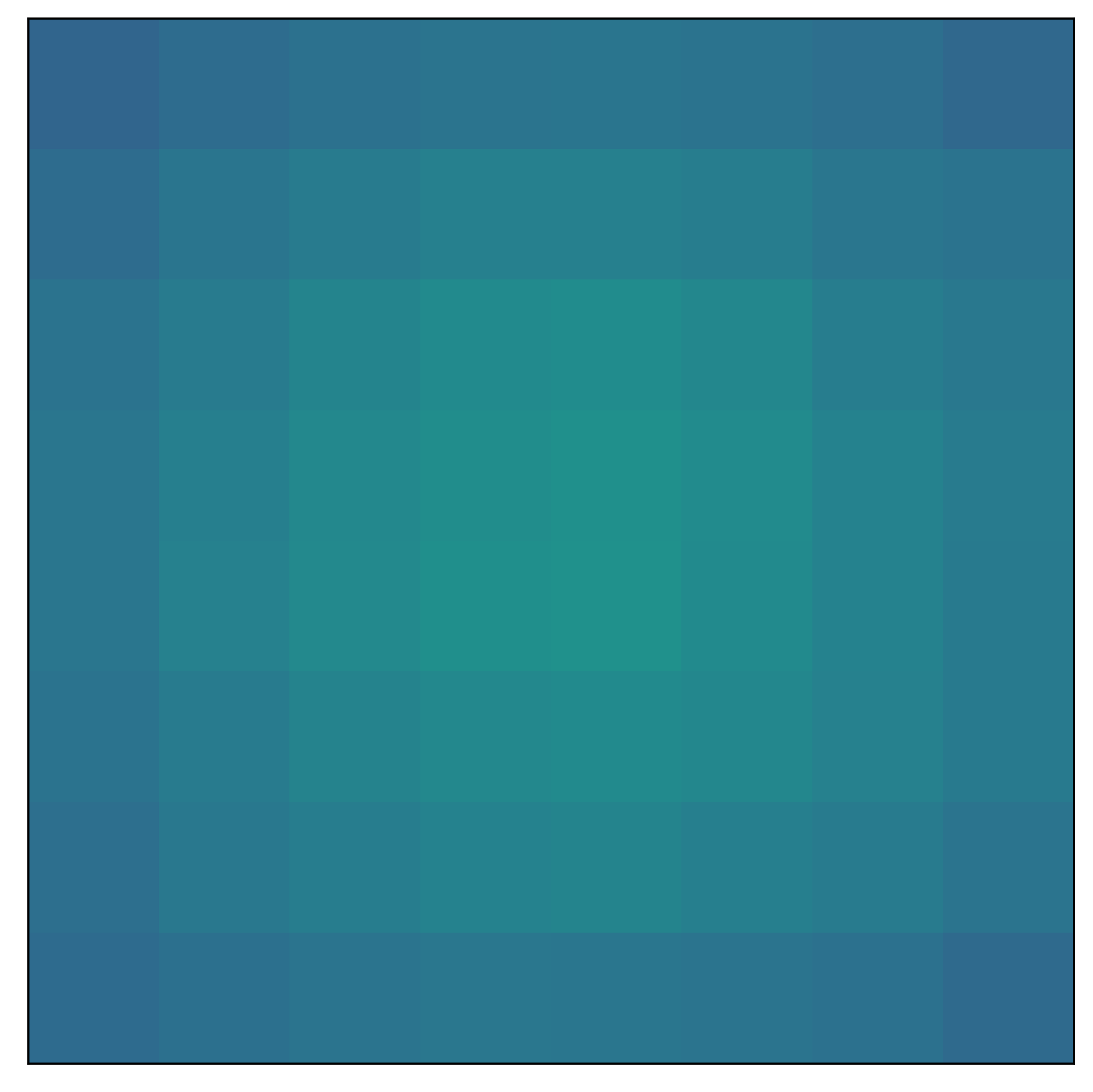}
	}
	\subfloat[Block3-Conv1]{
	    \includegraphics[width=0.15\columnwidth]{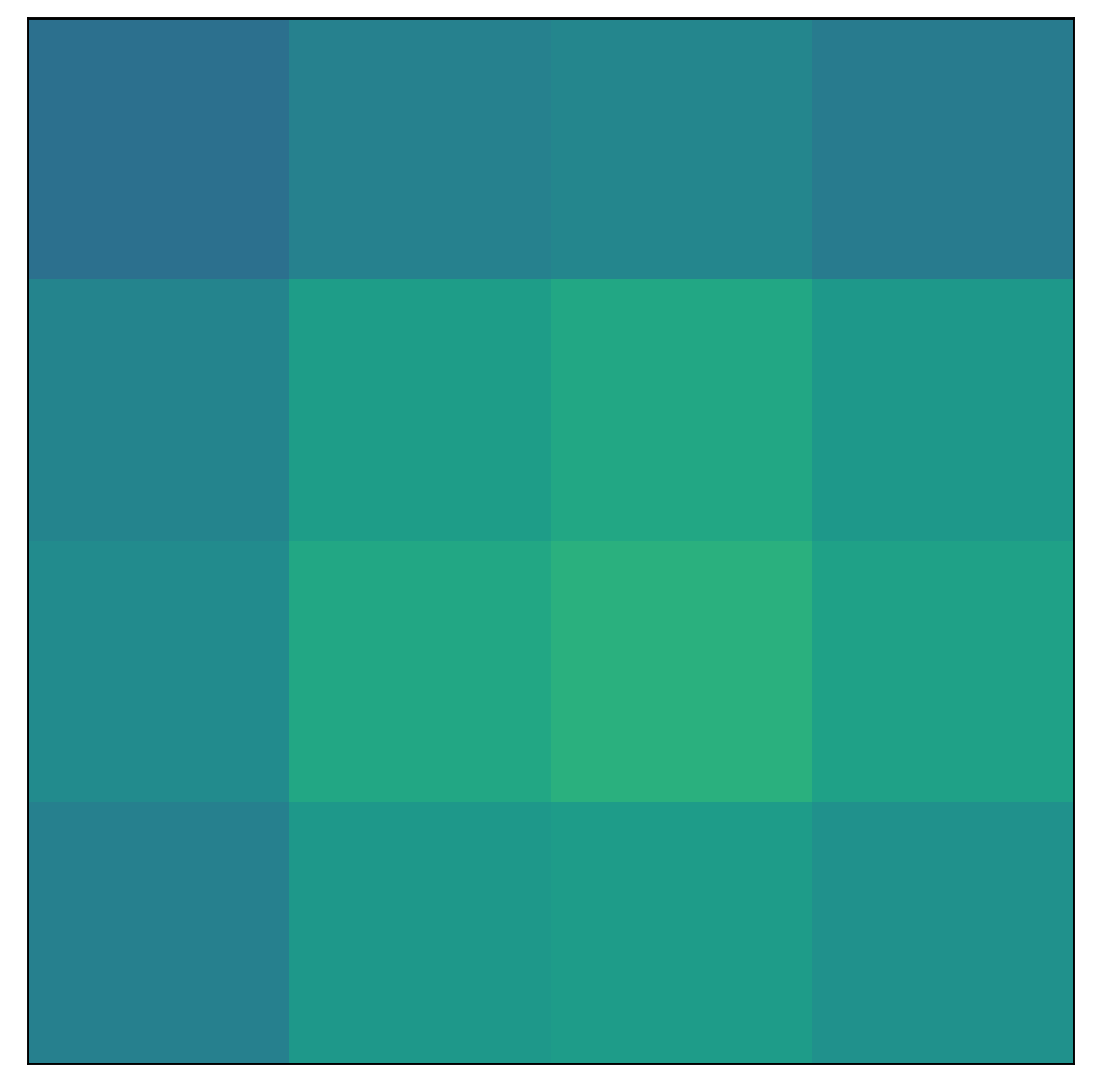}
	} \quad
	\subfloat[Block3-Conv2]{
	    \includegraphics[width=0.15\columnwidth]{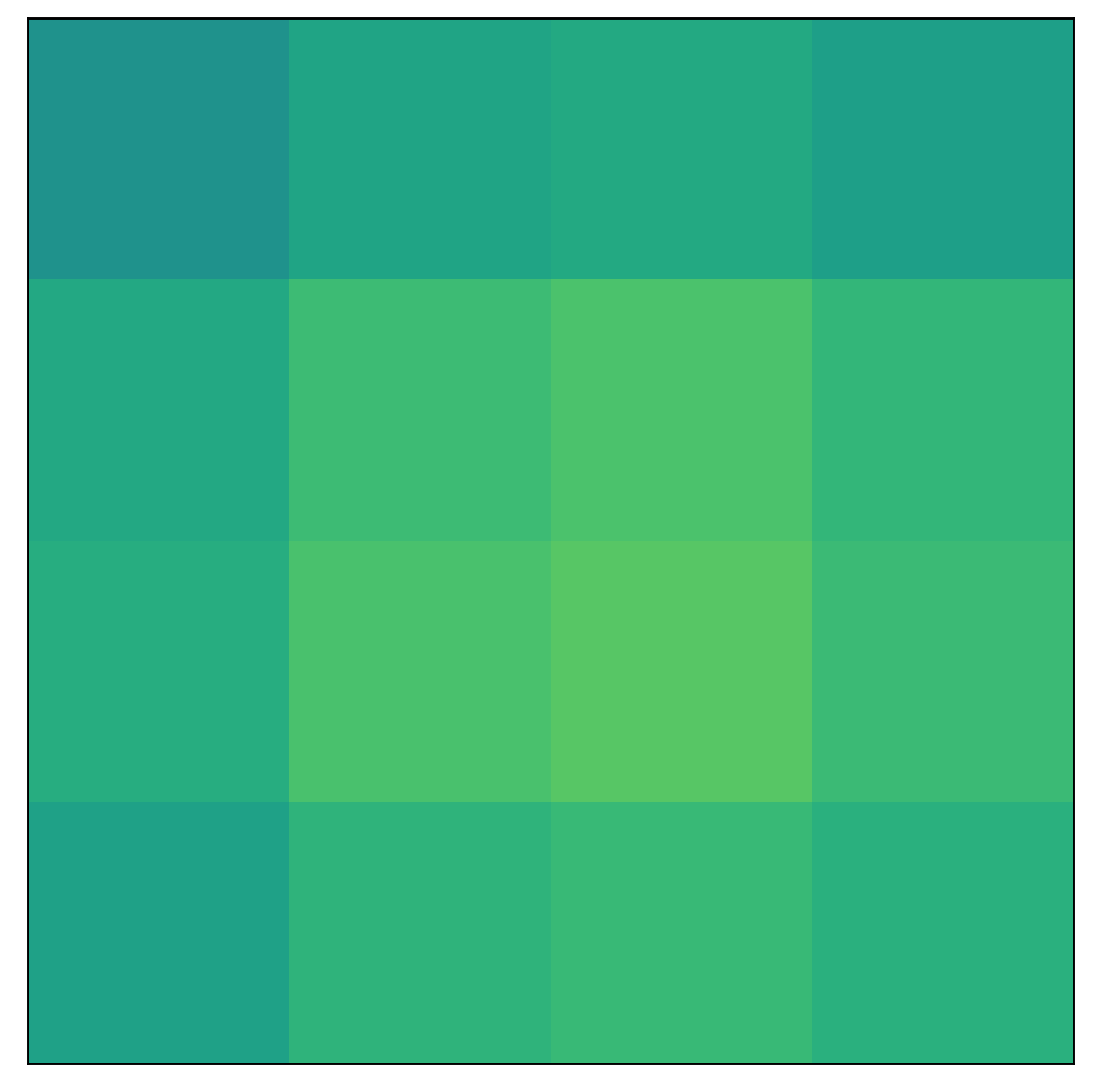}
	} 
	\subfloat[Block4-Conv1]{
	    \includegraphics[width=0.15\columnwidth]{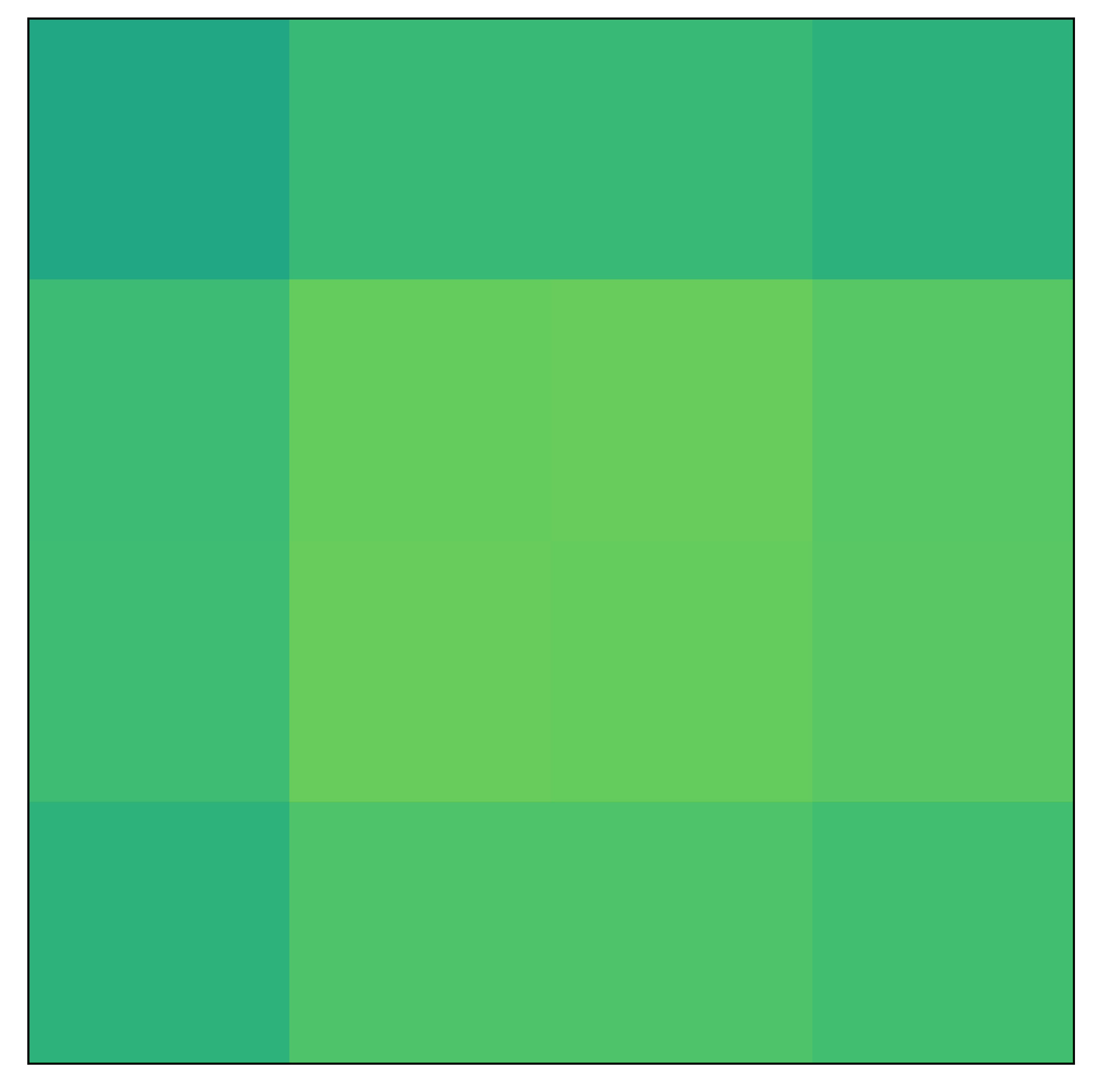}
	}
	\subfloat[Block4-Conv2]{
	    \includegraphics[width=0.15\columnwidth]{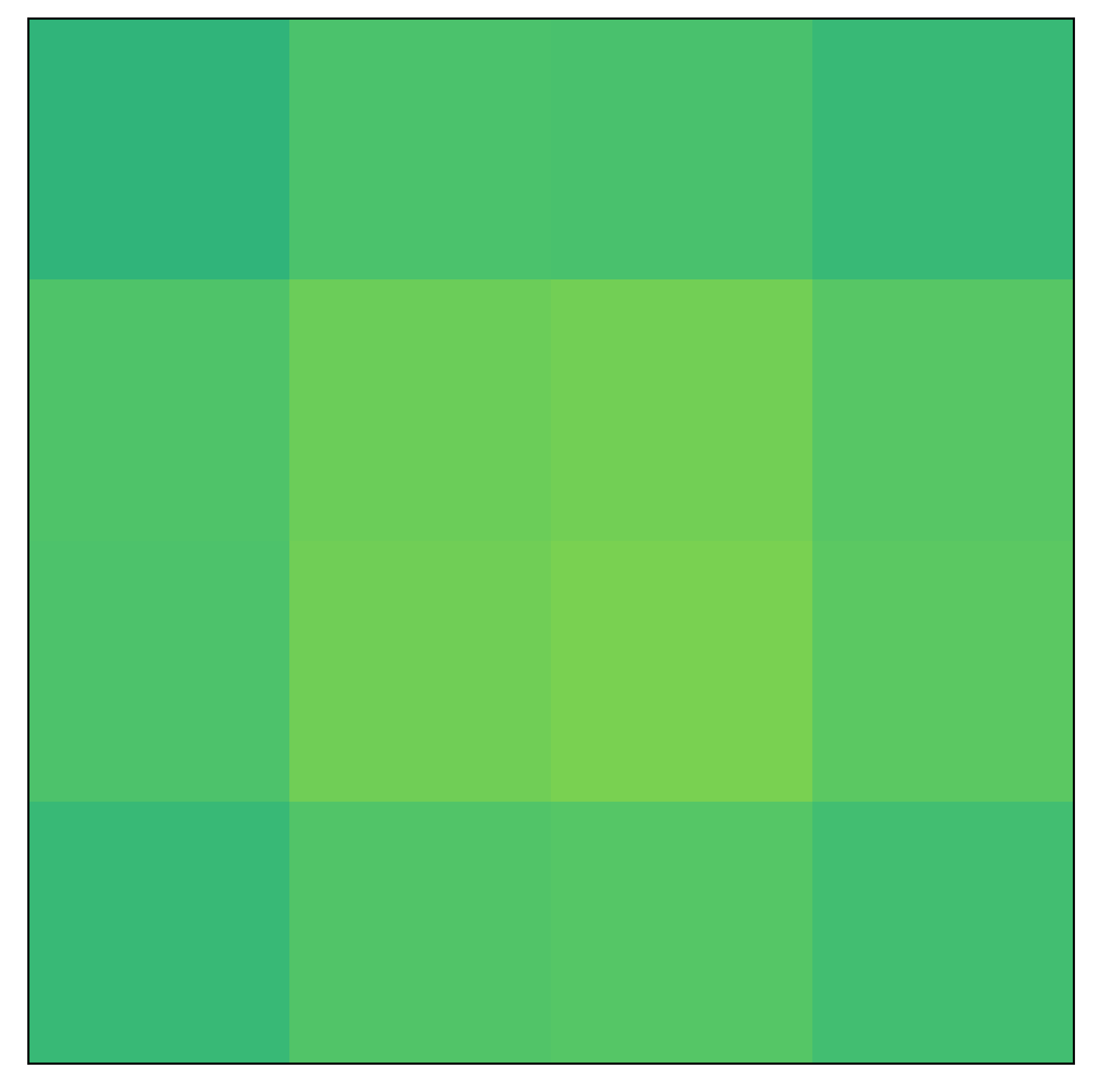}
	}
	\subfloat[Block5-Conv1]{
	    \includegraphics[width=0.15\columnwidth]{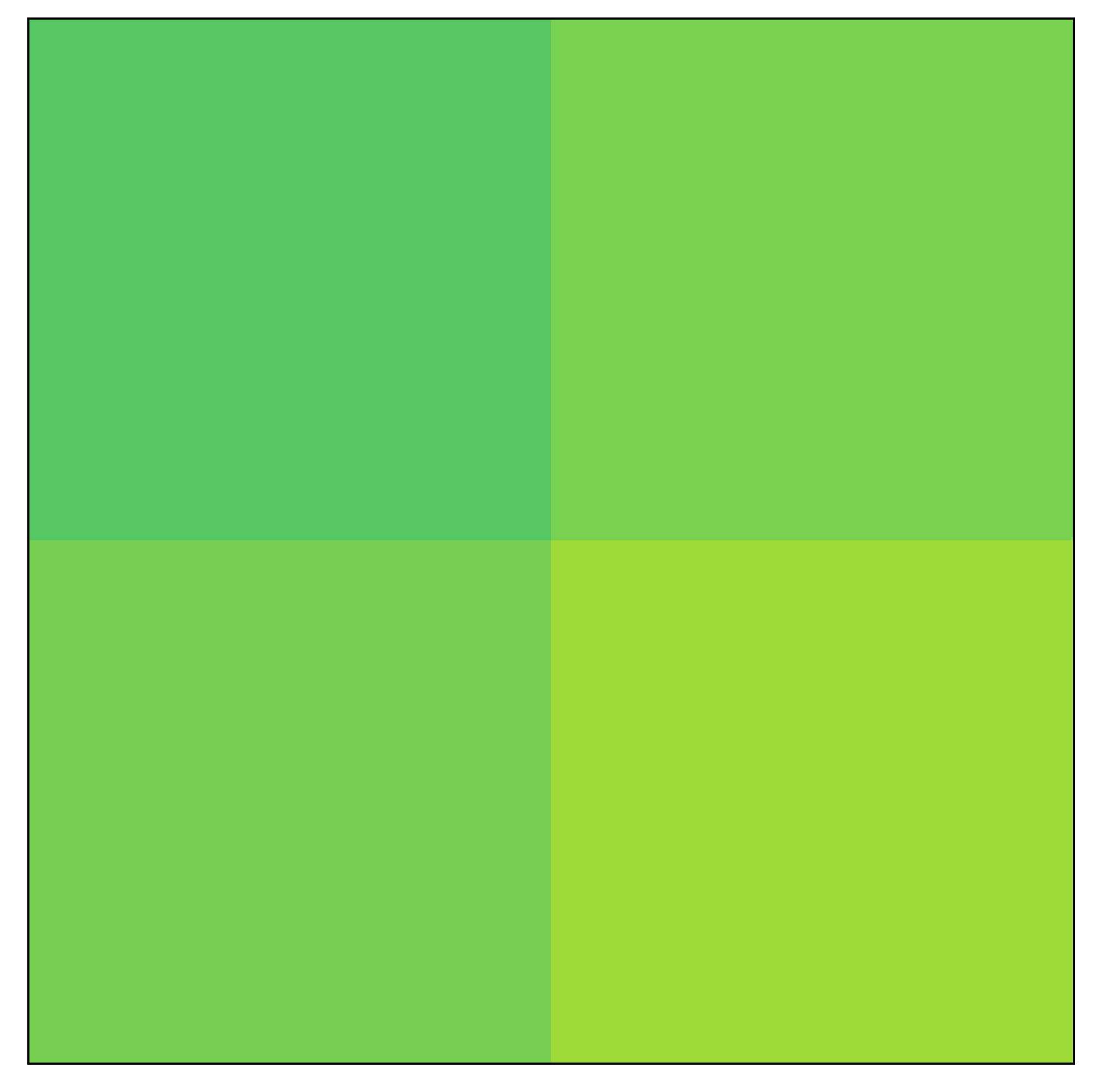}
	}
	\subfloat[Block5-Conv2]{
	    \includegraphics[width=0.15\columnwidth]{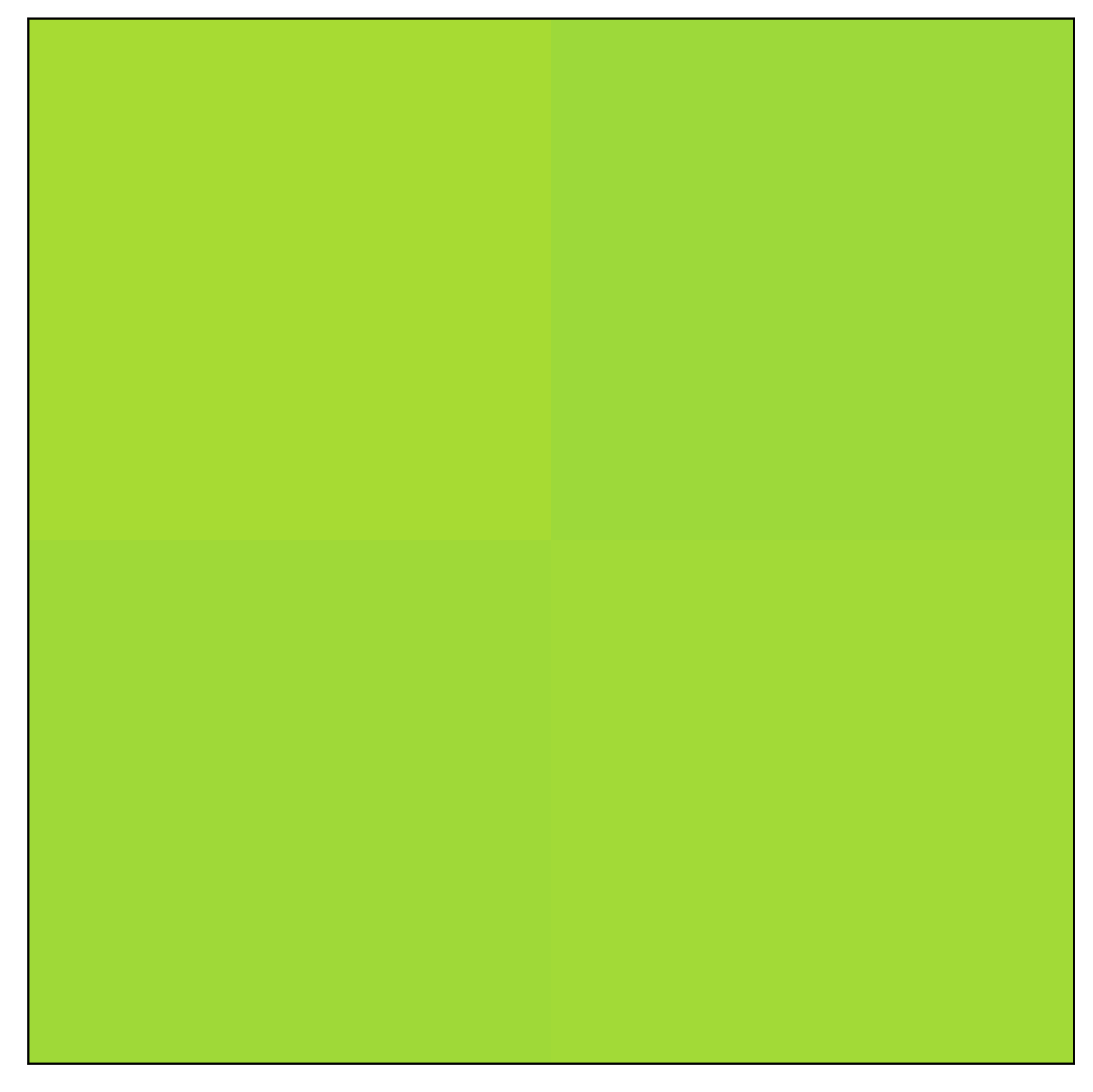}
	}
	\subfloat[Block6-Conv1]{
	    \includegraphics[width=0.15\columnwidth]{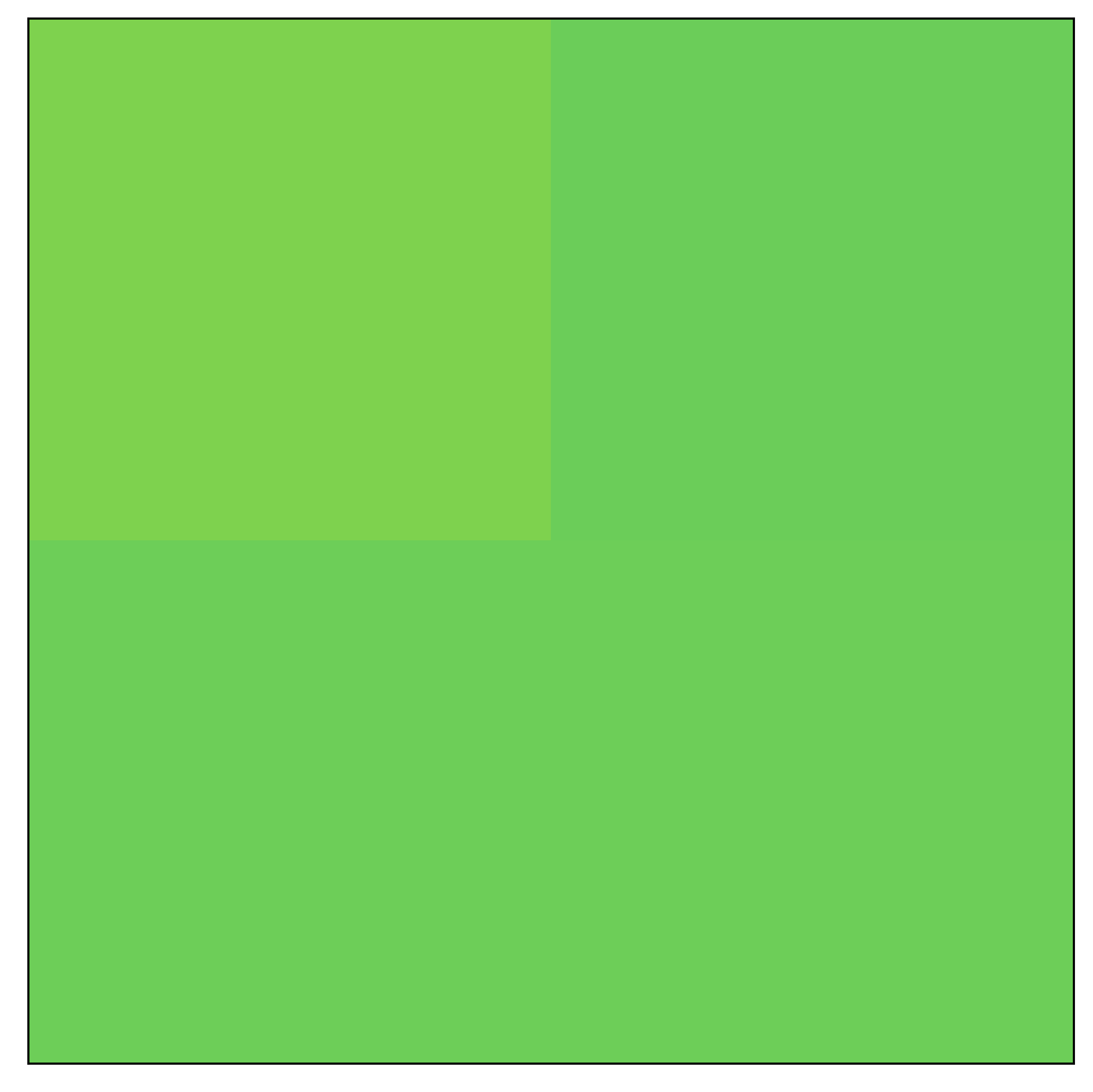}
	} \quad
	\subfloat[Block6-Conv2]{
	    \includegraphics[width=0.15\columnwidth]{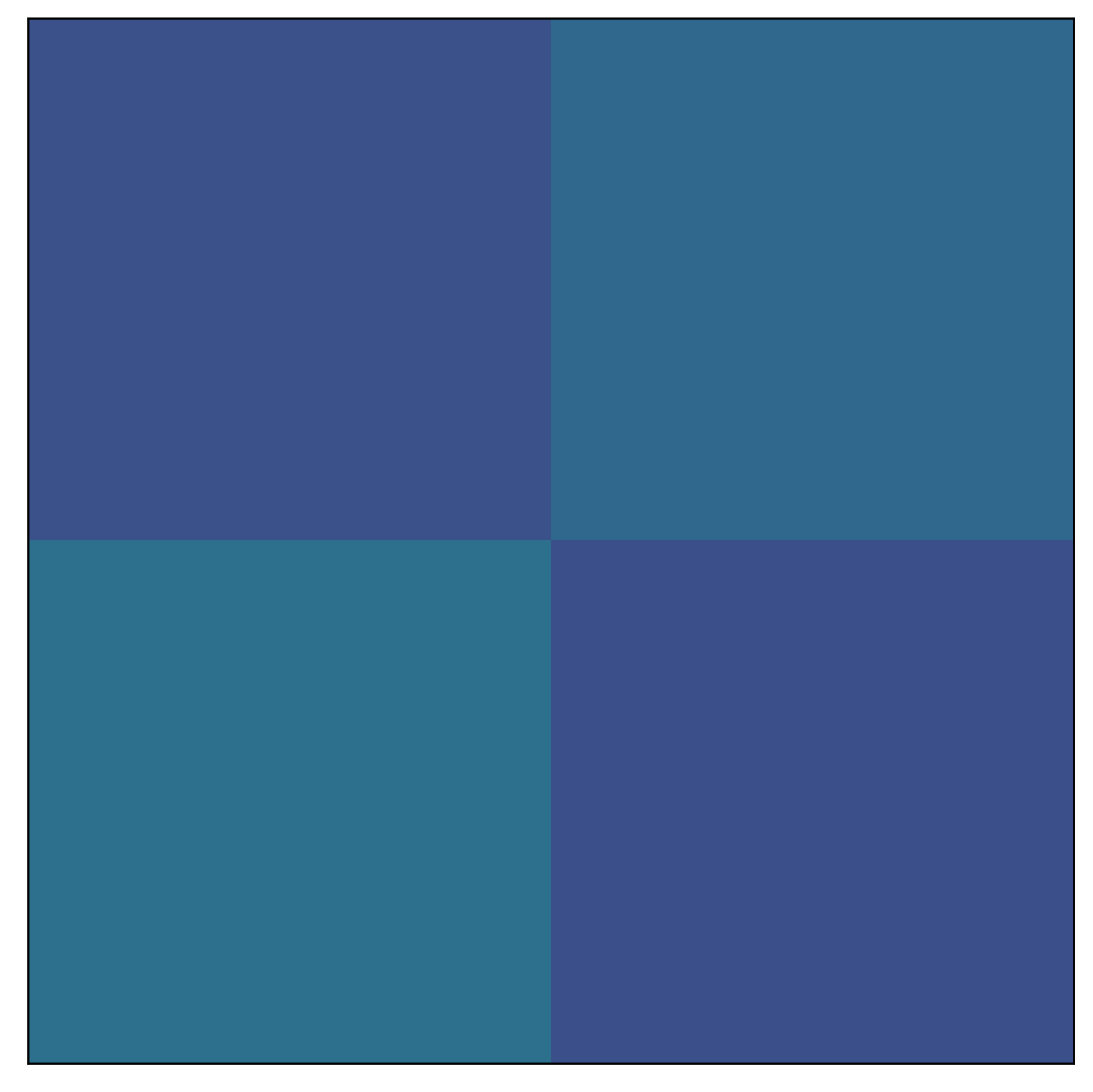}
	}
	\subfloat[Block7-Conv1]{
	    \includegraphics[width=0.15\columnwidth]{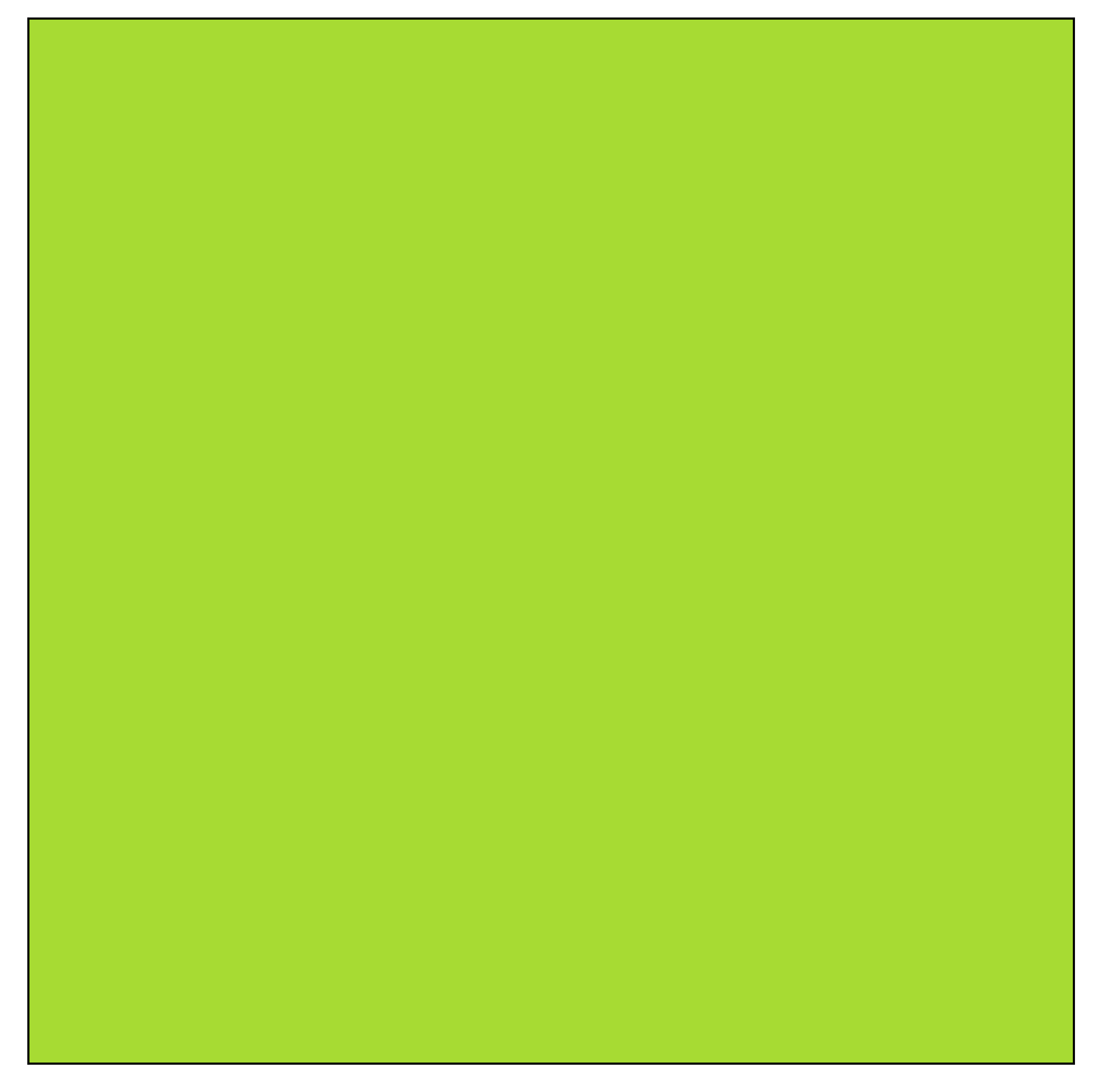}
	}
	\subfloat[Block7-Conv2]{
	    \includegraphics[width=0.15\columnwidth]{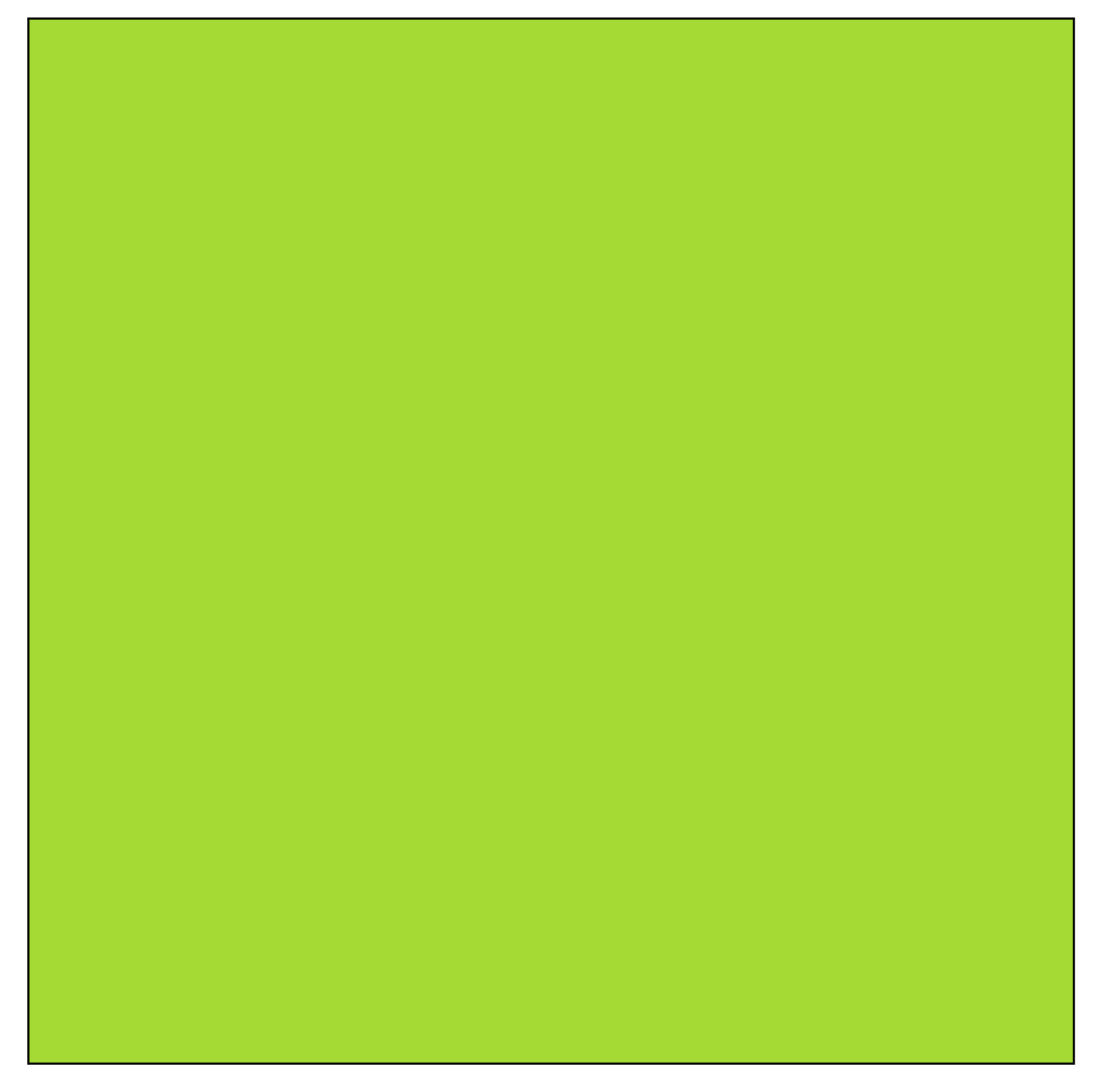}
	}
	\subfloat[Block8-Conv1]{
	    \includegraphics[width=0.15\columnwidth]{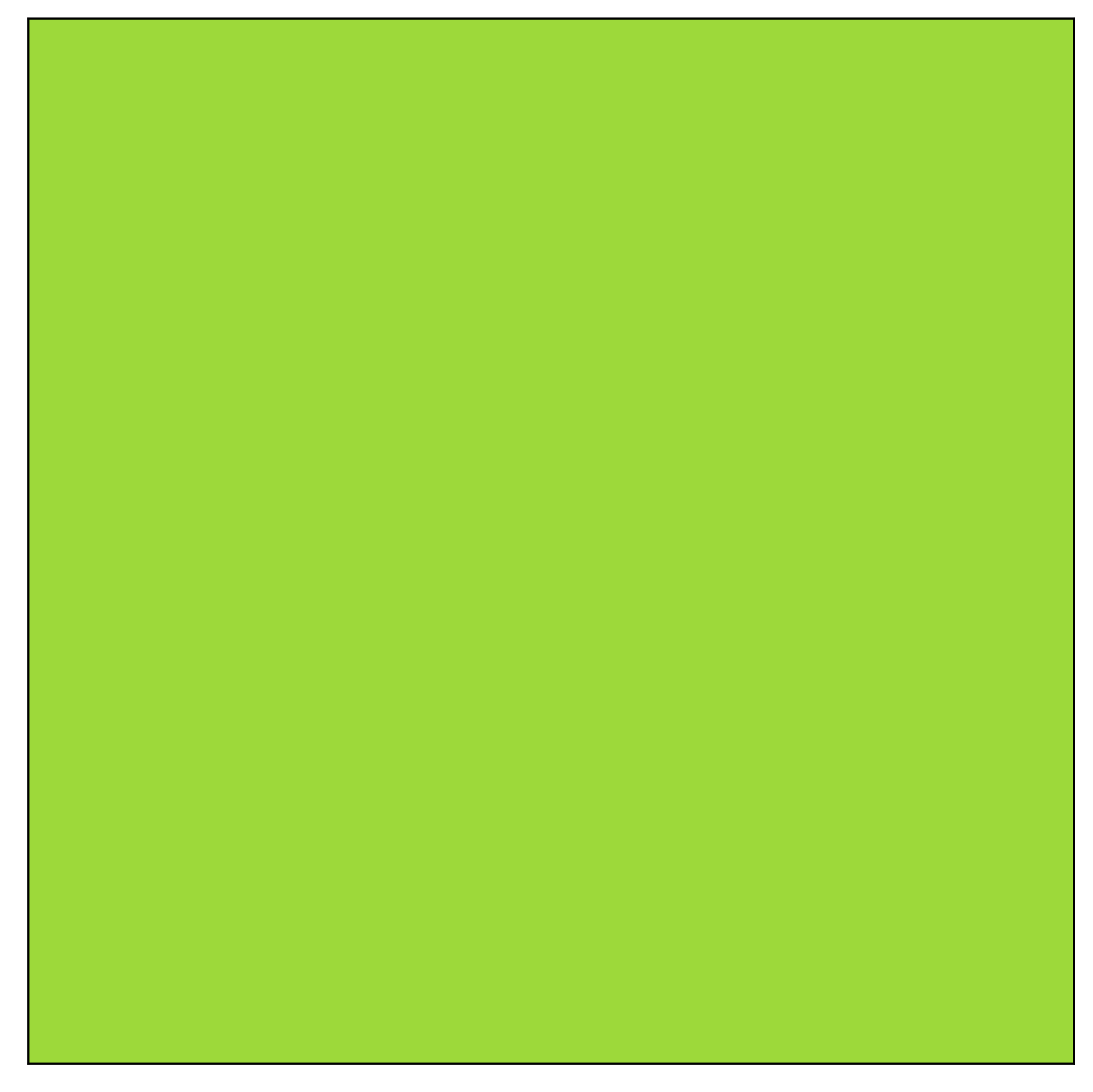}
	}
	\subfloat[Block8-Conv2]{
	    \includegraphics[width=0.15\columnwidth]{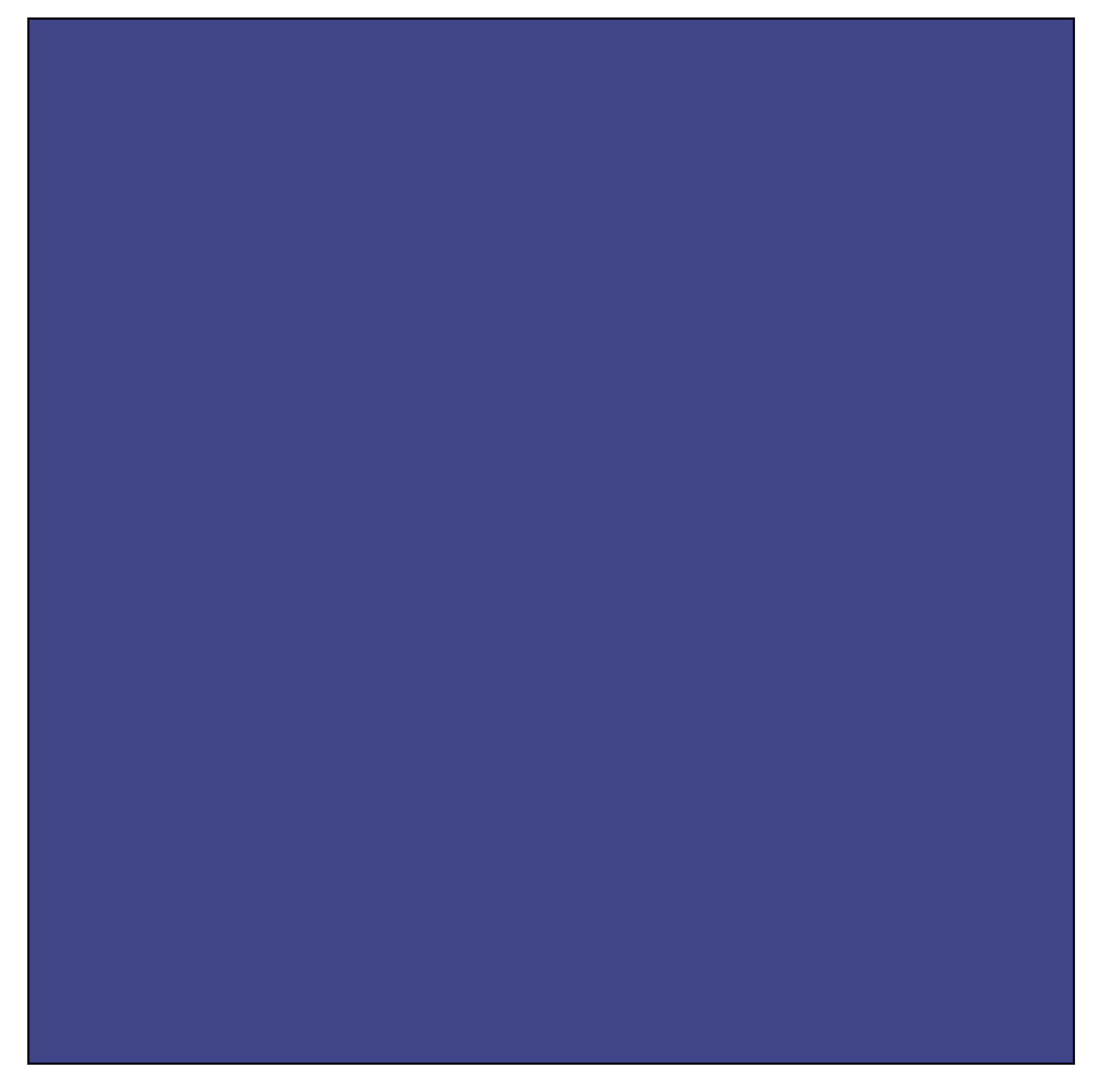}
	} \quad
	\subfloat{
	    \includegraphics[width=0.8\columnwidth]{figures/receptive_field_feature_maps/colorbar.png}
	}
	\caption{ResNet18 trained on Cifar10 alongside the heatmaps generated from the relative accuracy of the partial solutions in each layer. Note that "skipped" layer (Block6-Conv2 and Block8-Con2) have a worse performance and worst partial solutions.}
	\label{fig:vgg_probe_heatmaps2b}

\end{figure}
\FloatBarrier
\clearpage

\subsection{Receptive Field Analysis using Probes on ResNet18 optimized for Cifar10}
\FloatBarrier

\begin{figure}[htb!]
	\centering
	\subfloat[ResNet18 (Cifar10 Optimized) trained on Cifar10 using $32 \times 32$ pixel input size]{
	    \includegraphics[width=0.8\columnwidth]{figures/receptive_field_probe_sat/ResNet18_PC_Cifar10_32.png}
	} \quad
	\subfloat[Conv1]{
	    \includegraphics[width=0.15\columnwidth]{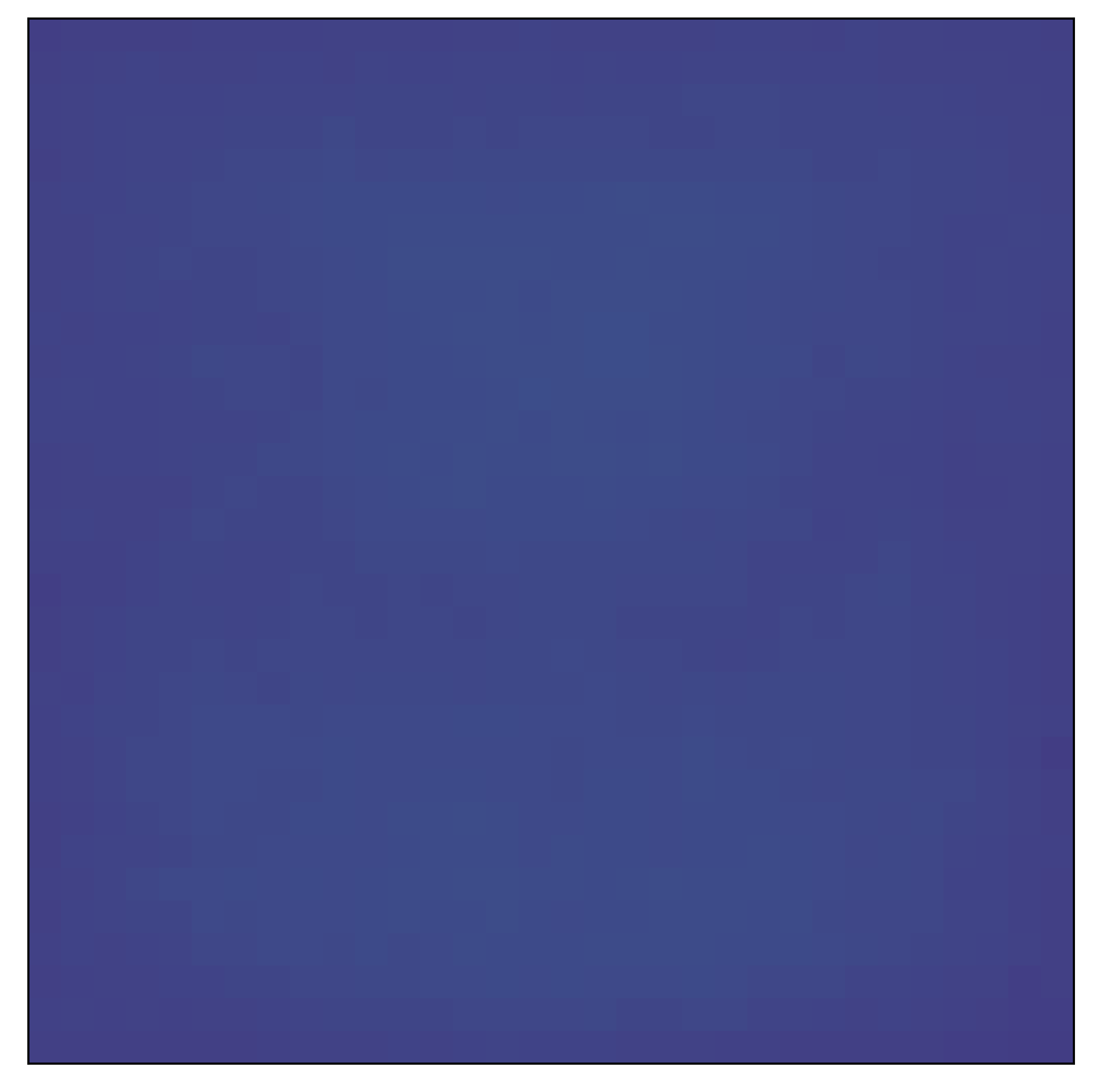}
	}
	\subfloat[Block1-Conv1]{
	    \includegraphics[width=0.15\columnwidth]{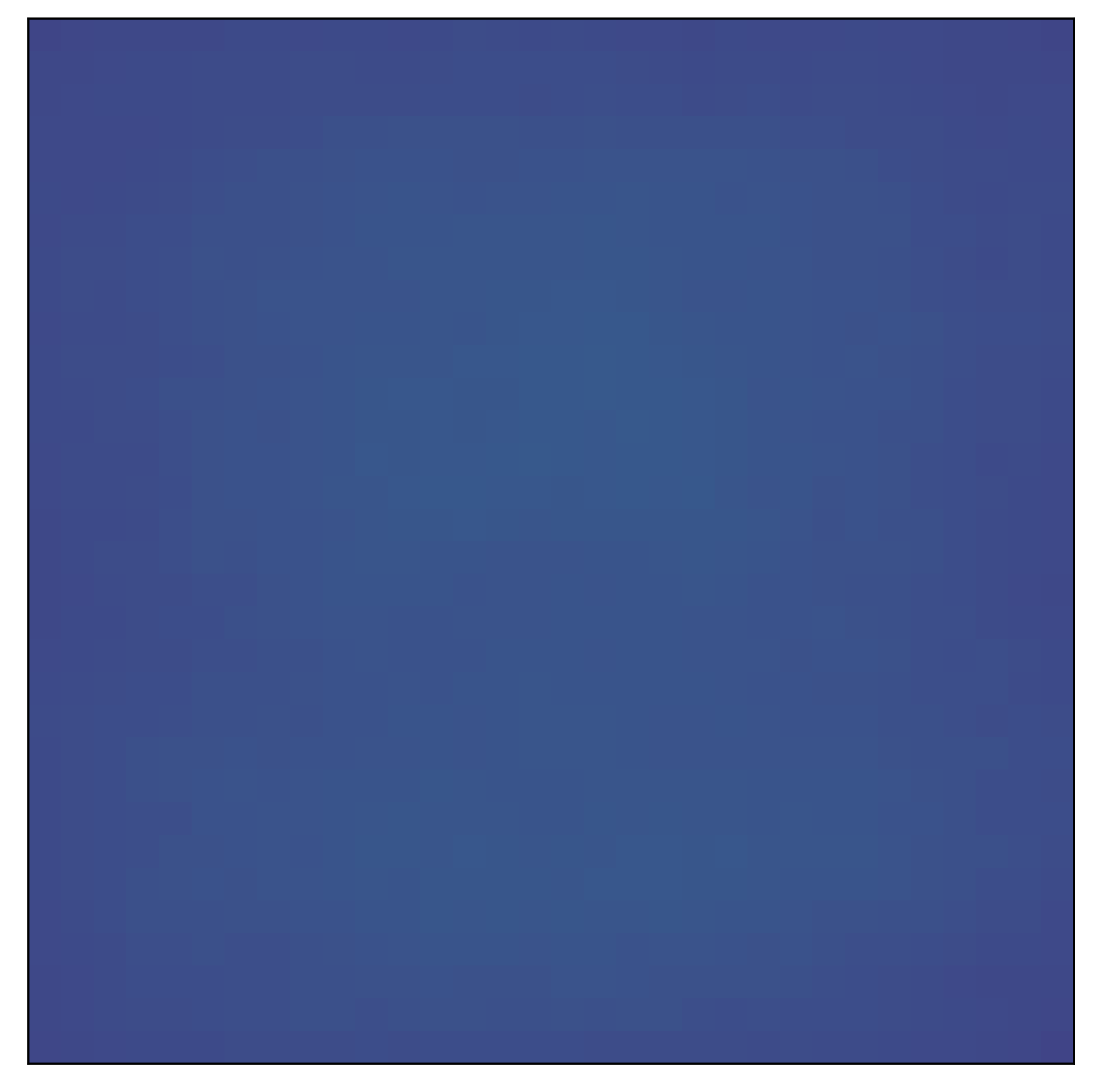}
	}
	\subfloat[Block1-Conv2]{
	    \includegraphics[width=0.15\columnwidth]{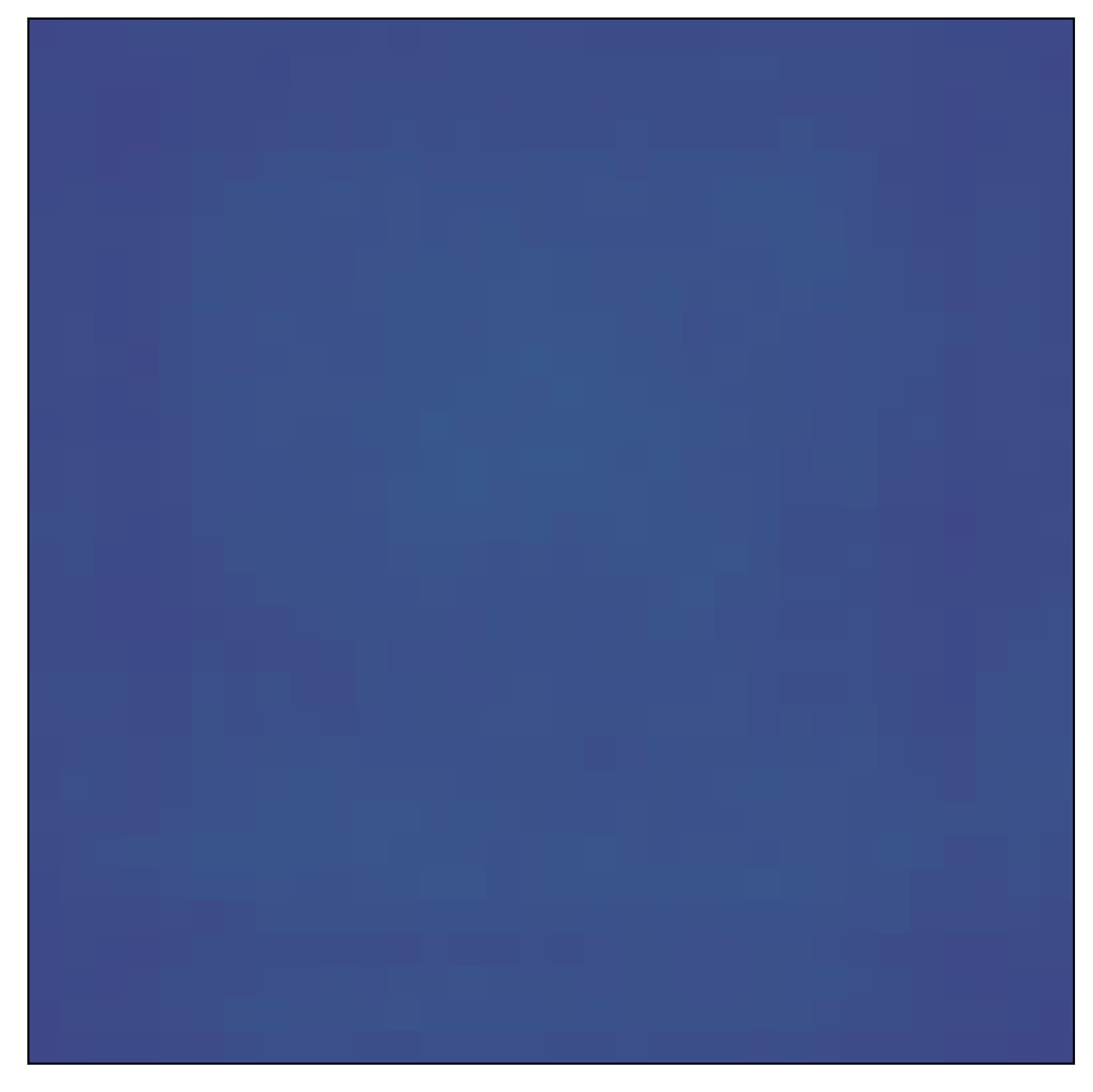}
	}
	\subfloat[Block2-Conv1]{
	    \includegraphics[width=0.15\columnwidth]{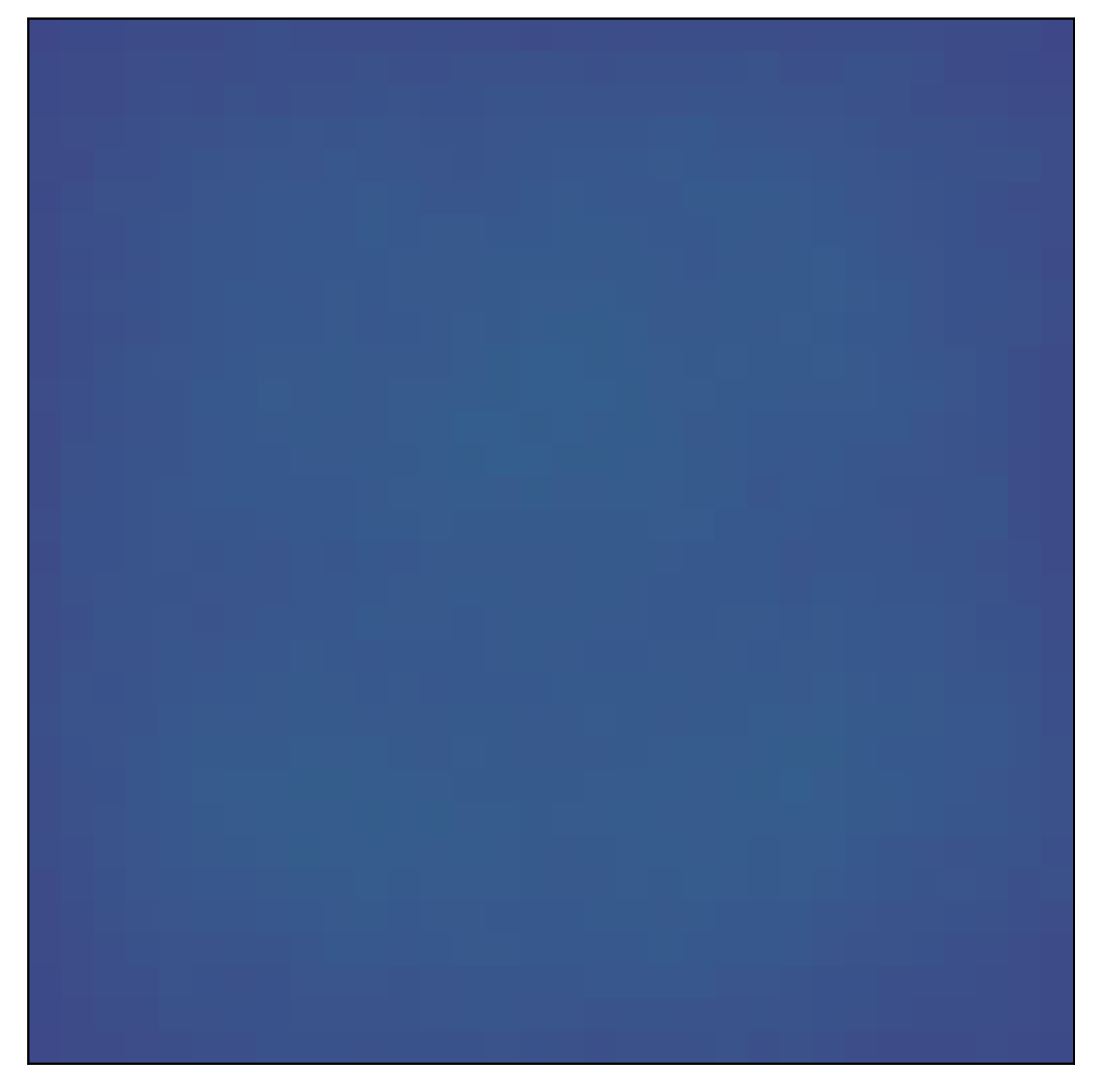}
	}
	\subfloat[Block2-Conv2]{
	    \includegraphics[width=0.15\columnwidth]{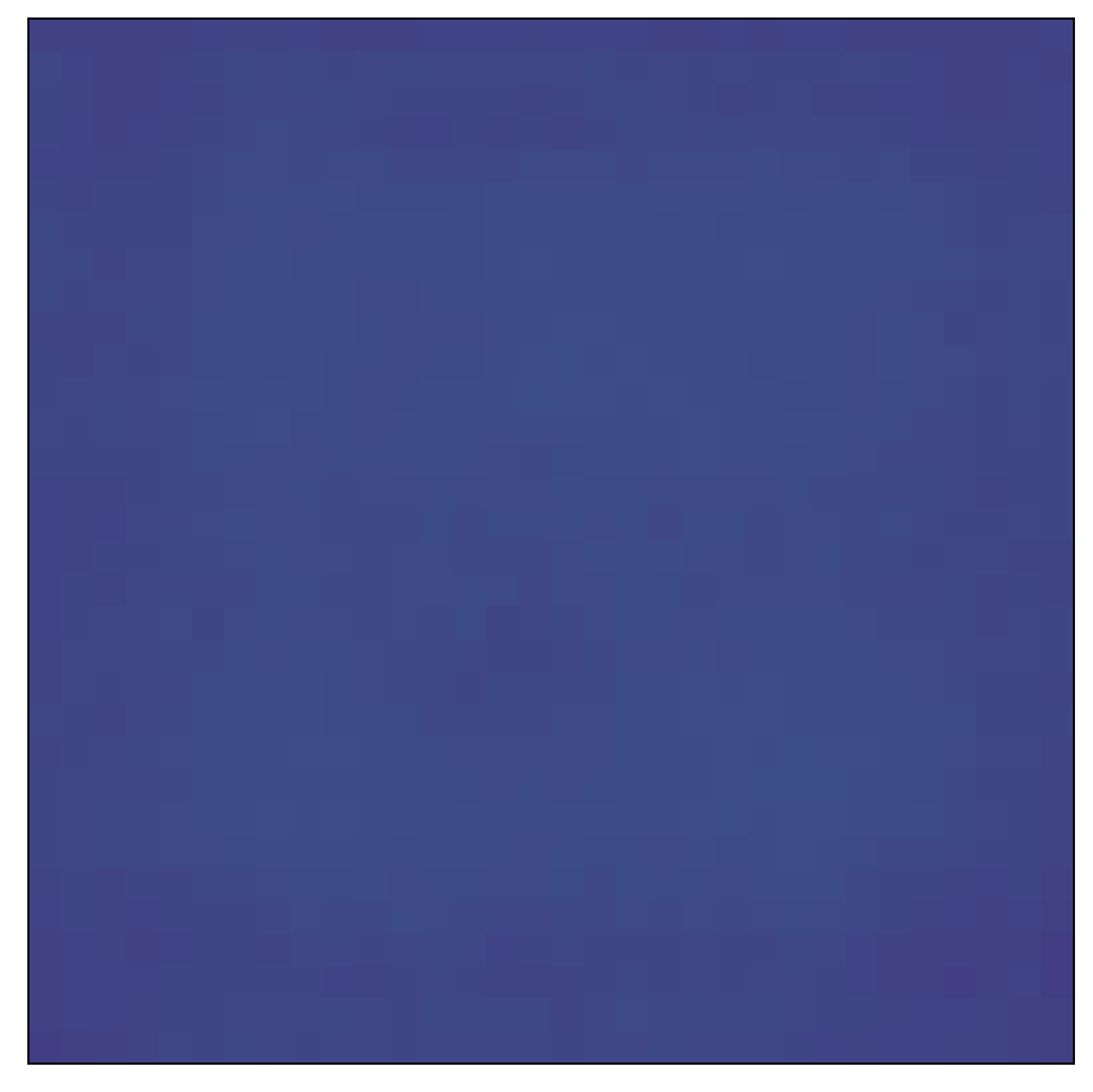}
	}
	\subfloat[Block3-Conv1]{
	    \includegraphics[width=0.15\columnwidth]{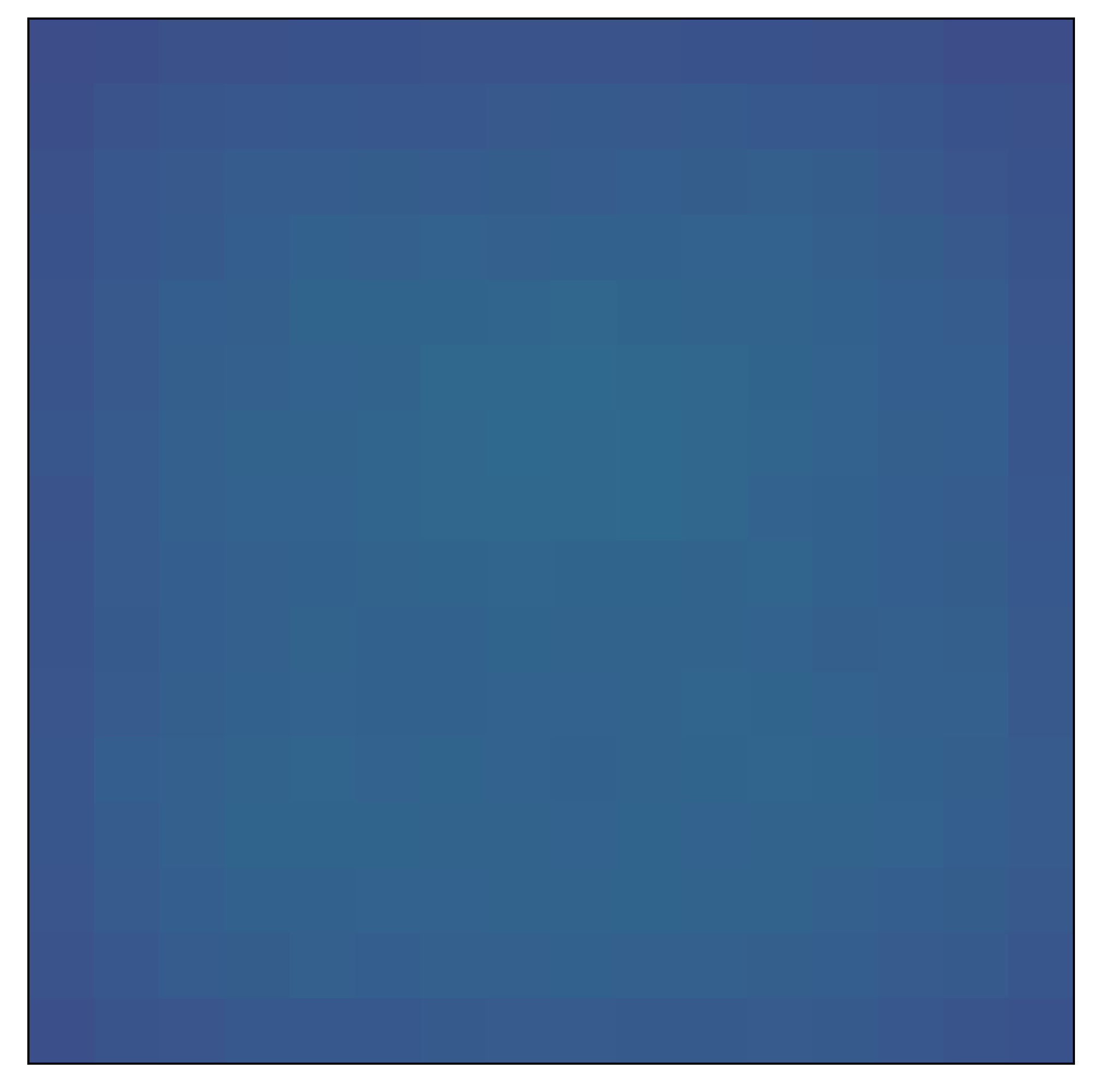}
	} \quad
	\subfloat[Block3-Conv2]{
	    \includegraphics[width=0.15\columnwidth]{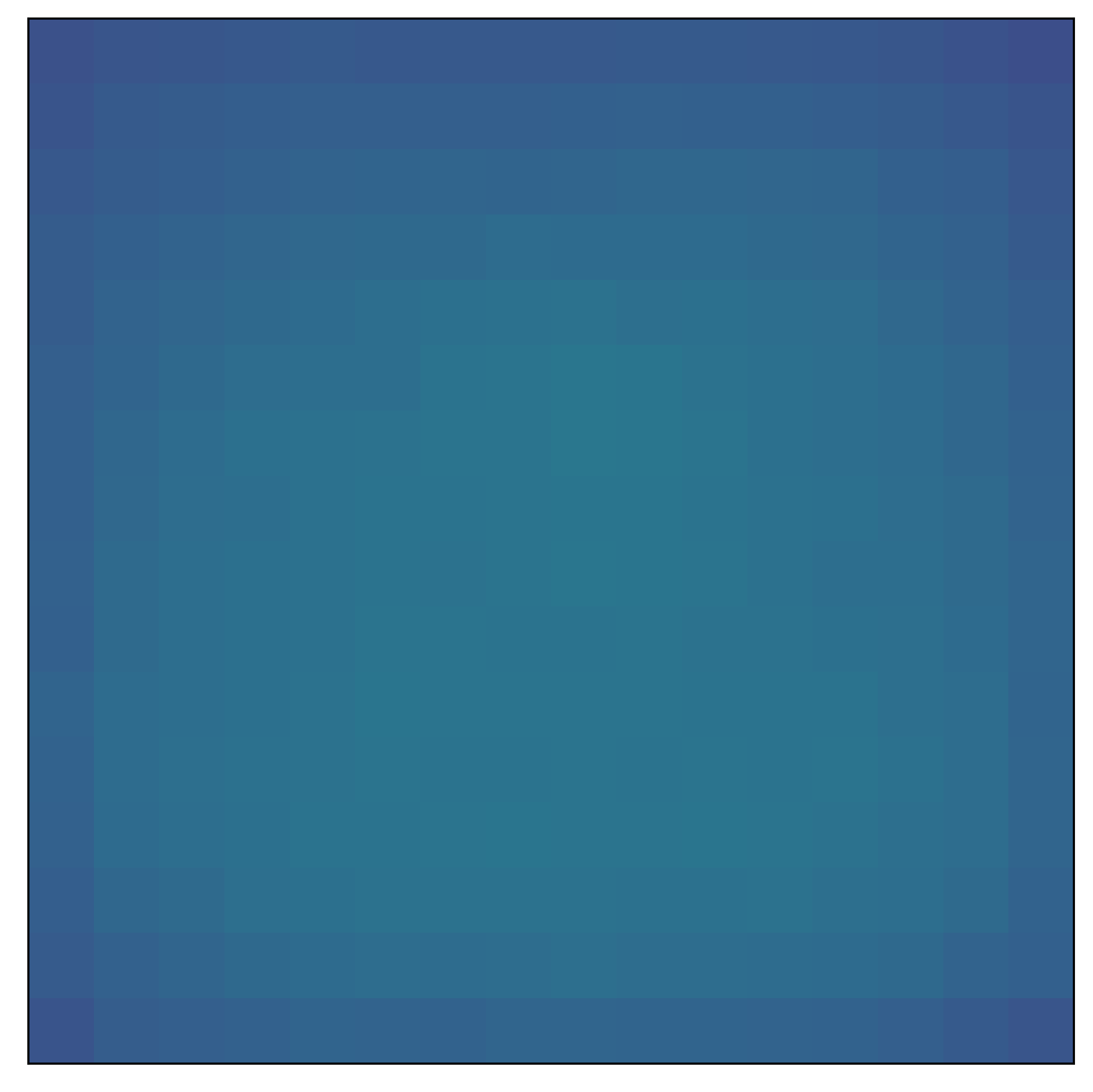}
	} 
	\subfloat[Block4-Conv1]{
	    \includegraphics[width=0.15\columnwidth]{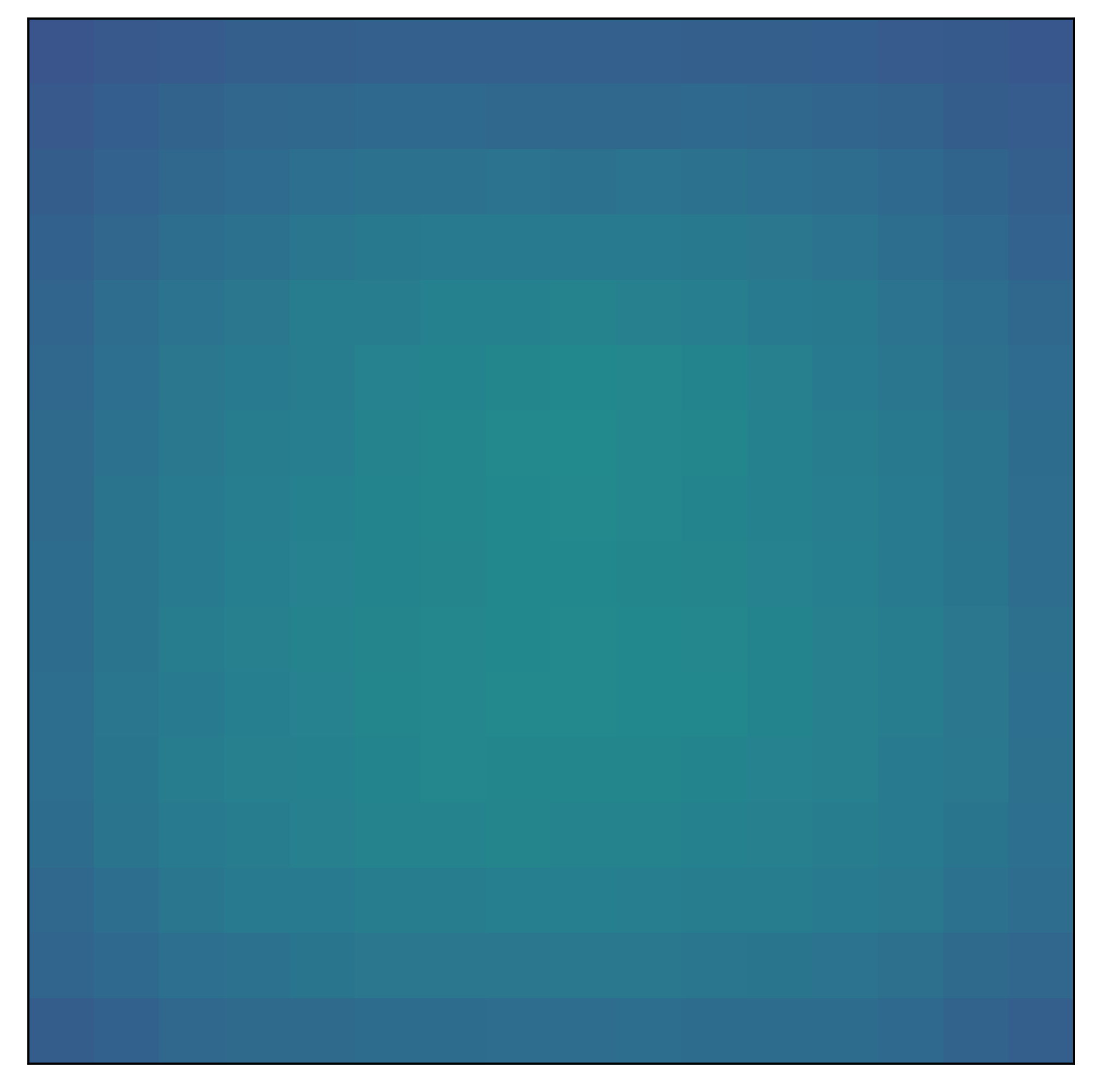}
	}
	\subfloat[Block4-Conv2]{
	    \includegraphics[width=0.15\columnwidth]{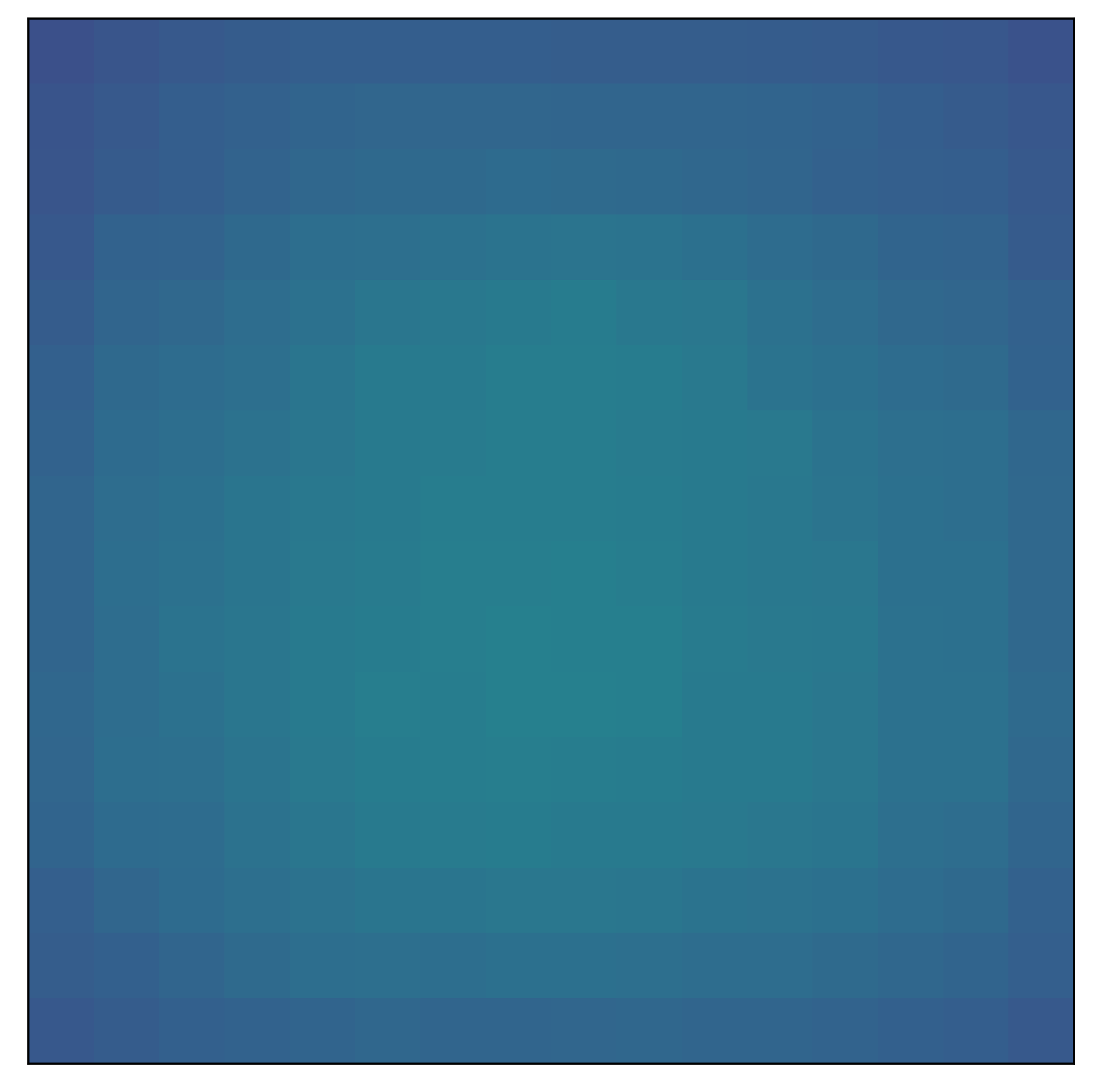}
	}
	\subfloat[Block5-Conv1]{
	    \includegraphics[width=0.15\columnwidth]{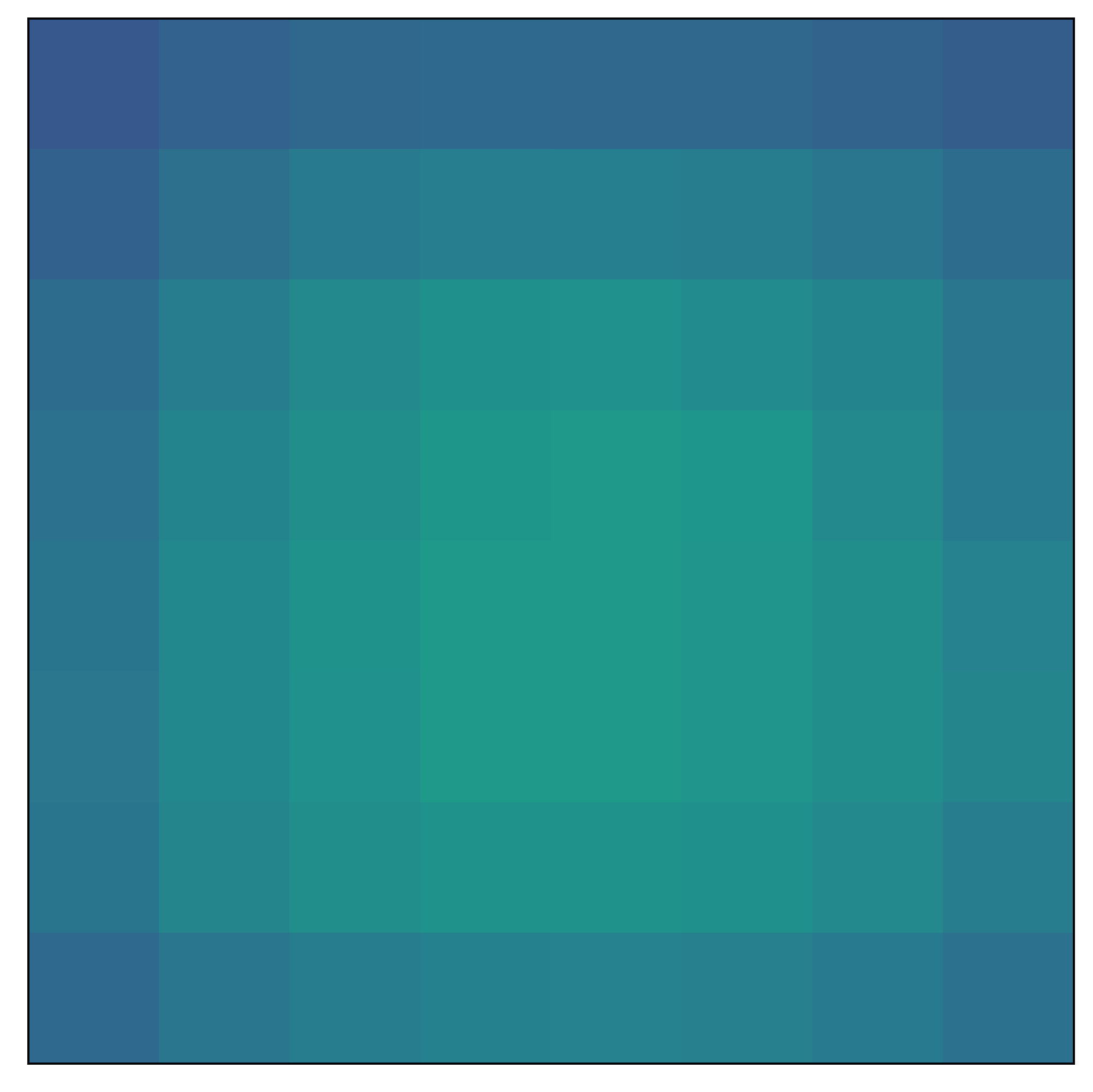}
	}
	\subfloat[Block5-Conv2]{
	    \includegraphics[width=0.15\columnwidth]{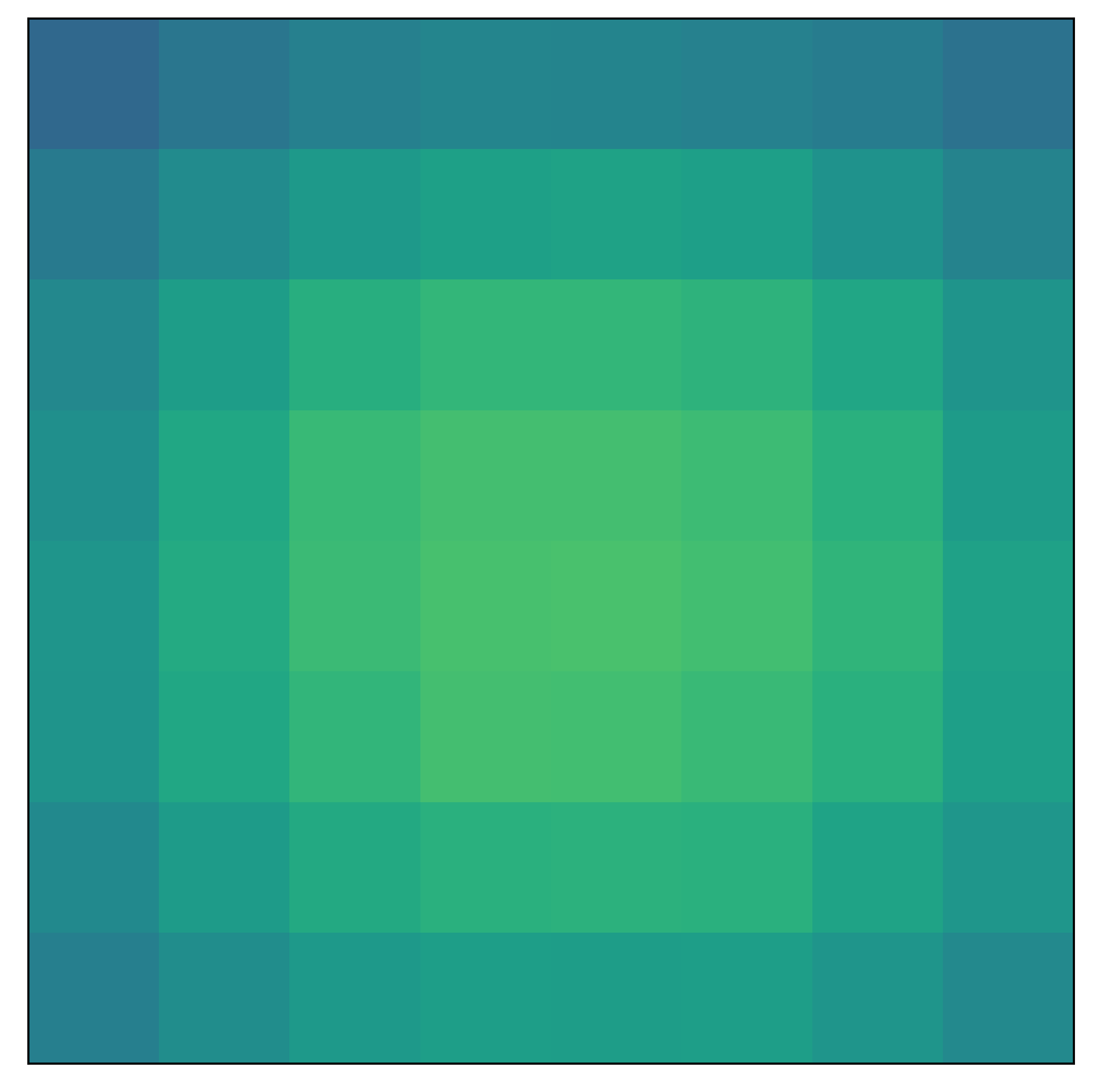}
	}
	\subfloat[Block6-Conv1]{
	    \includegraphics[width=0.15\columnwidth]{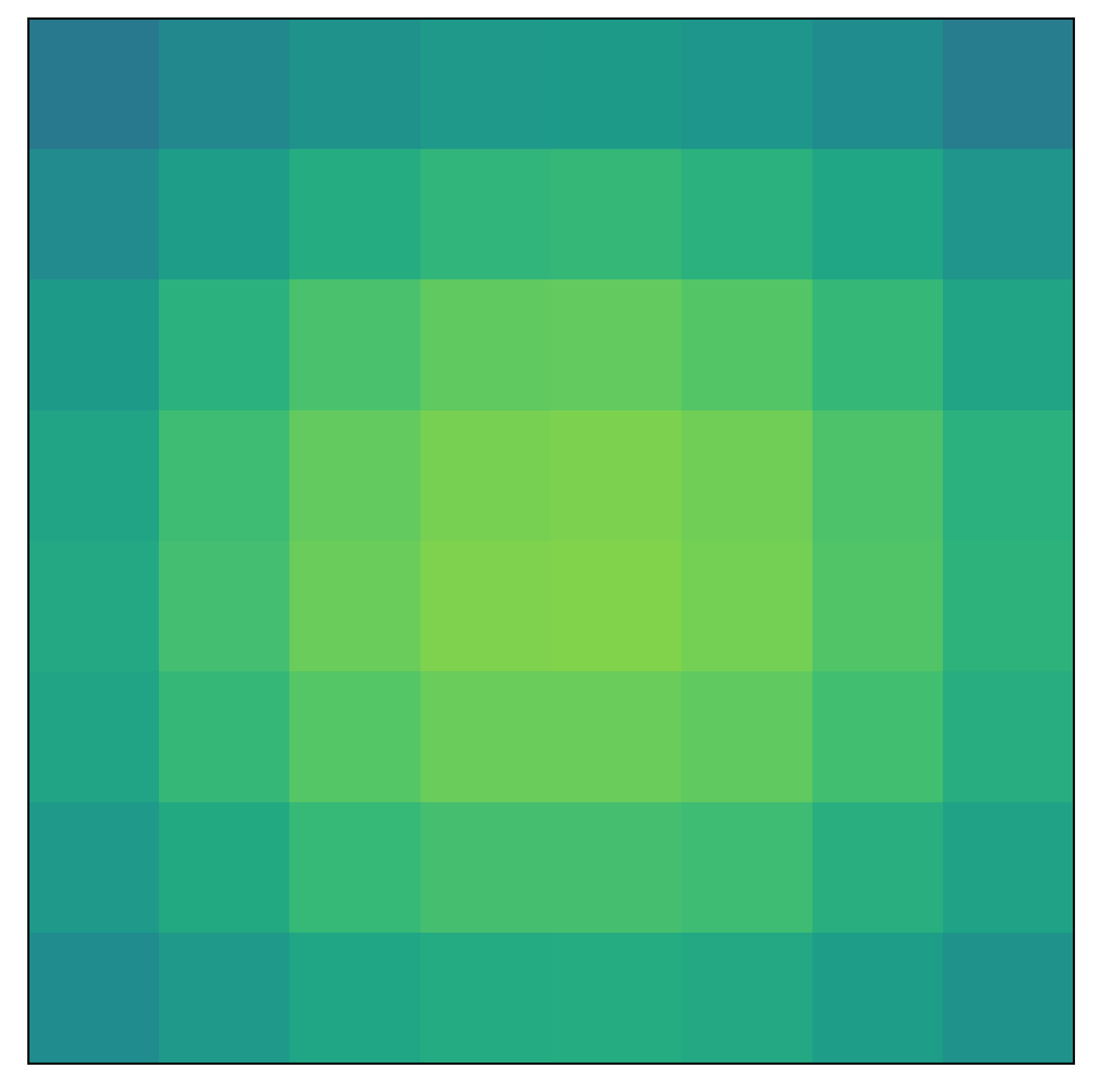}
	} \quad
	\subfloat[Block6-Conv2]{
	    \includegraphics[width=0.15\columnwidth]{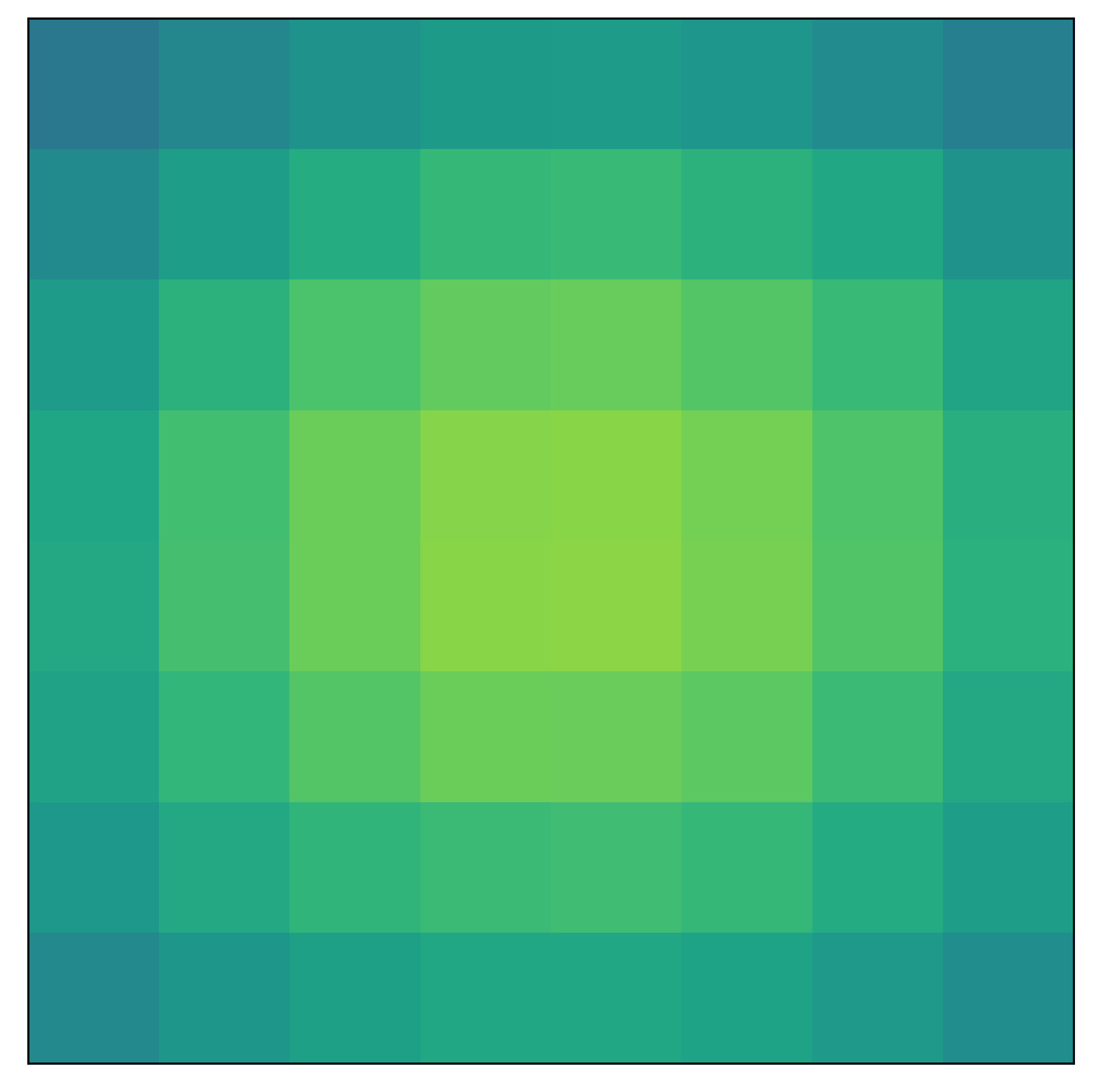}
	}
	\subfloat[Block7-Conv1]{
	    \includegraphics[width=0.15\columnwidth]{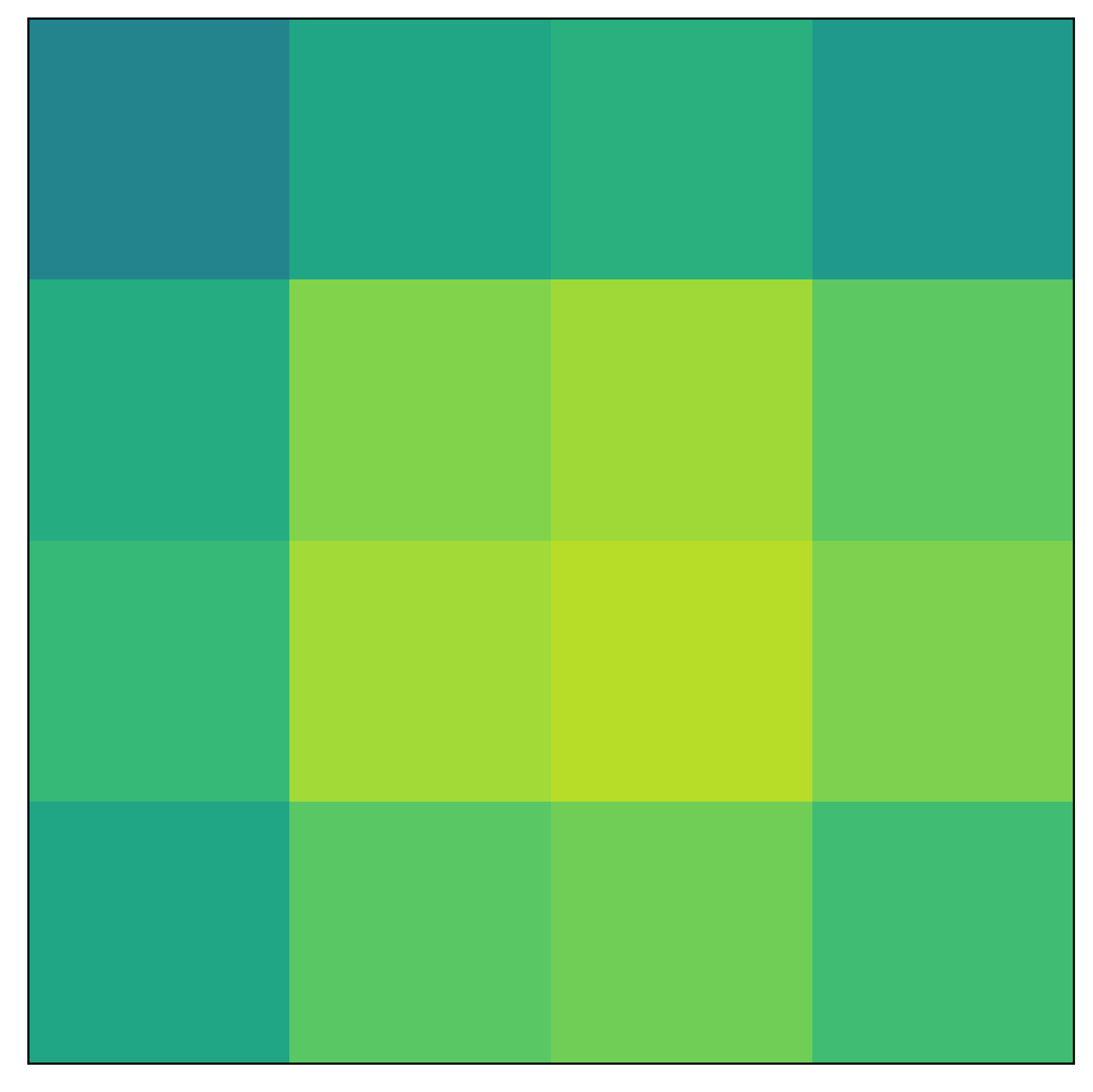}
	}
	\subfloat[Block7-Conv2]{
	    \includegraphics[width=0.15\columnwidth]{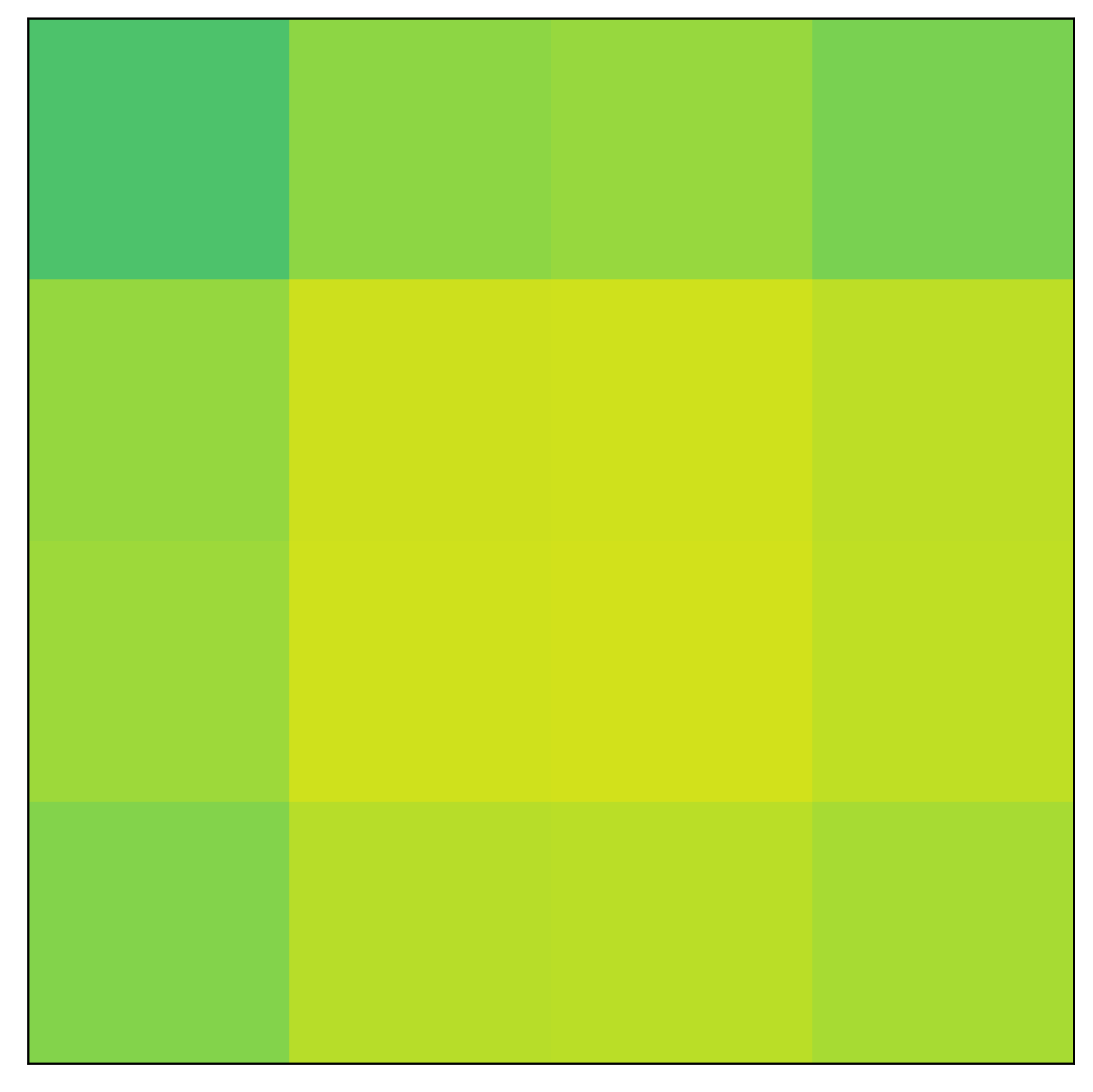}
	}
	\subfloat[Block8-Conv1]{
	    \includegraphics[width=0.15\columnwidth]{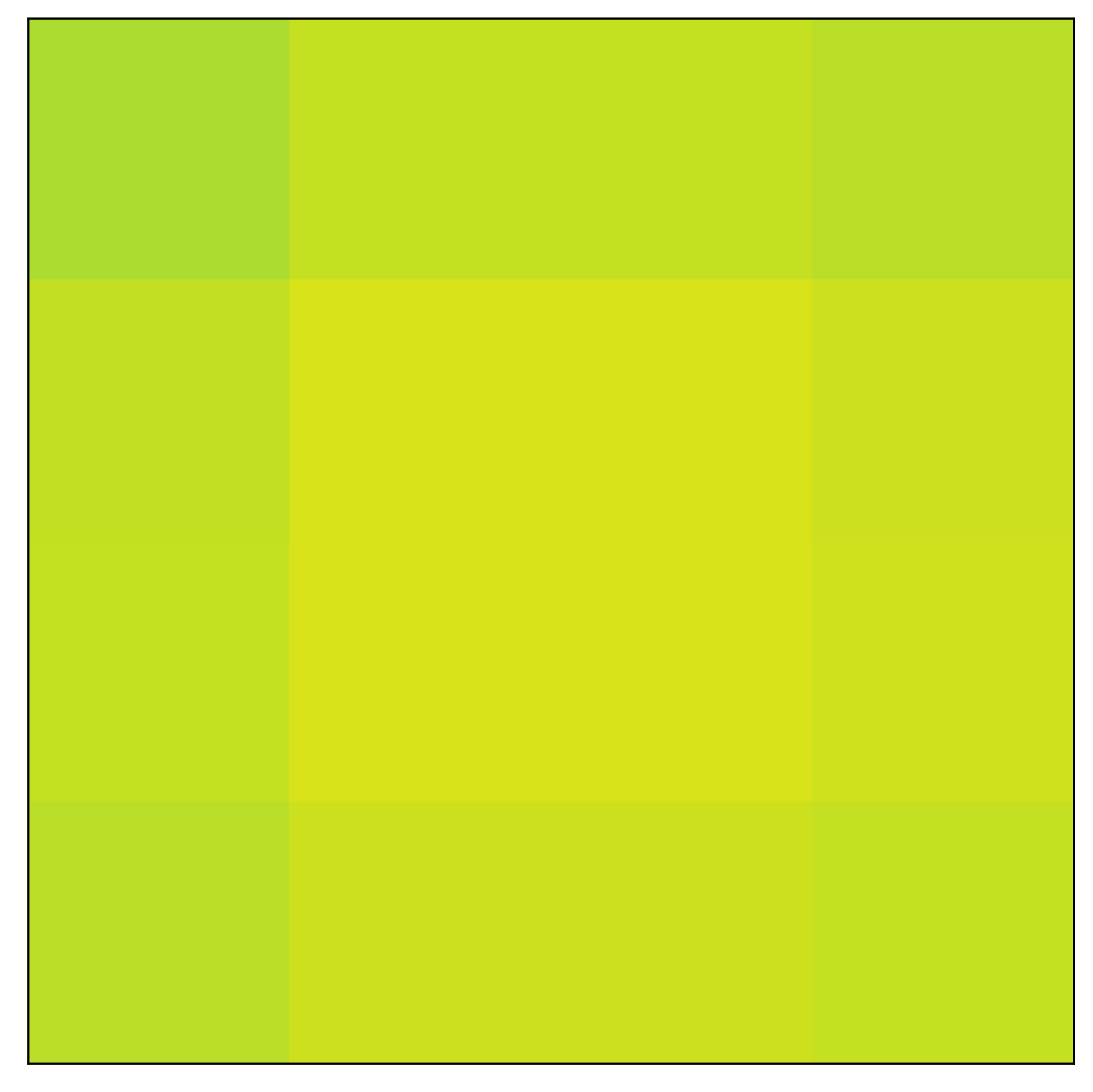}
	}
	\subfloat[Block8-Conv2]{
	    \includegraphics[width=0.15\columnwidth]{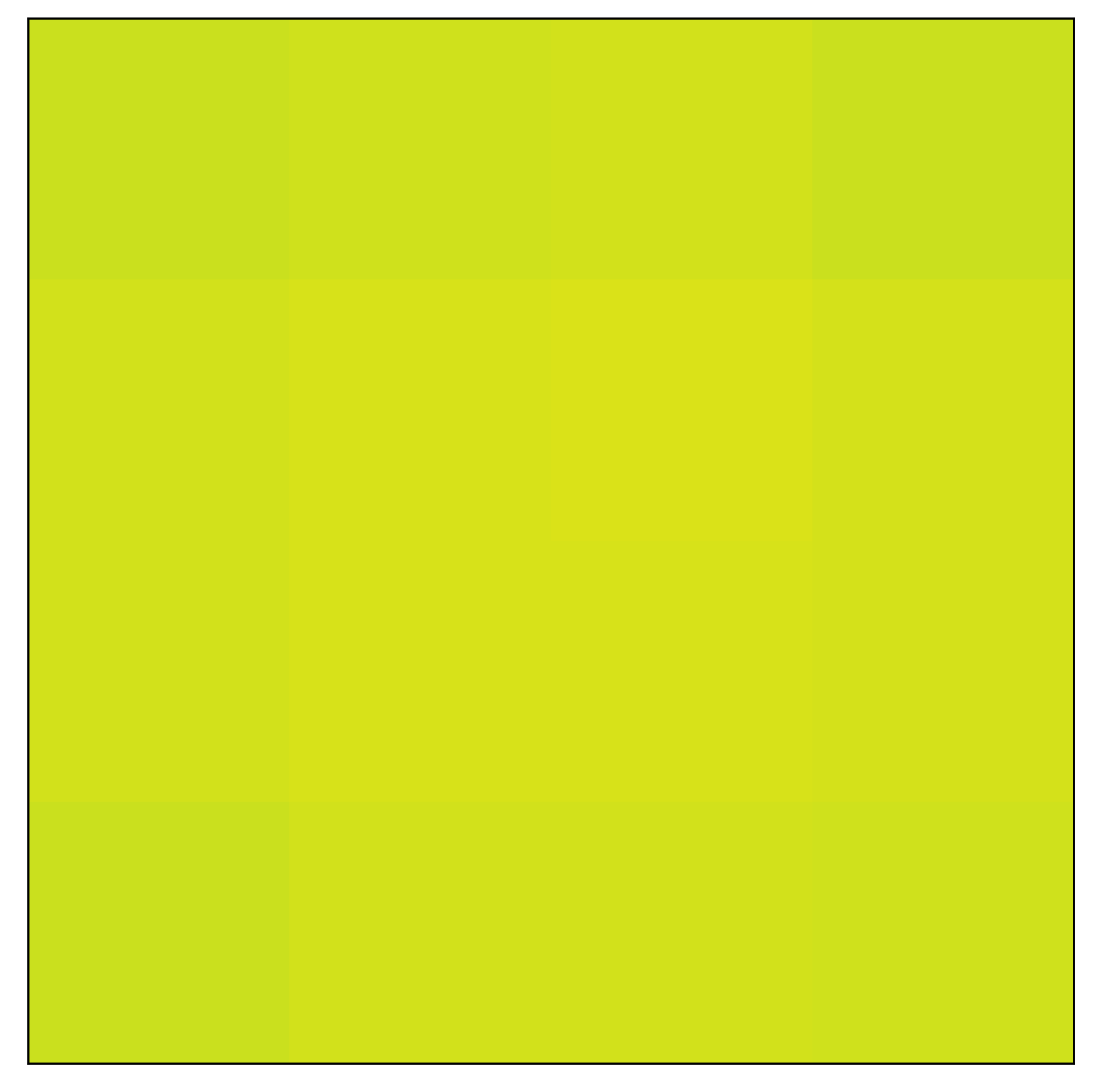}
	} \quad
	\subfloat{
	    \includegraphics[width=0.8\columnwidth]{figures/receptive_field_feature_maps/colorbar.png}
	}
	\caption{Cifar10 optimized ResNet18 trained on Cifar10 alongside the heatmaps generated from the relative accuracy of the partial solutions in each layer. Note that no layers are skipped and the solution quality of the partial solution develops up until the final residual block.}
	\label{fig:vgg_probe_heatmaps2c}

\end{figure}
\FloatBarrier
\clearpage

%%%%%%%%%%%%%%%%%%%%%%%%%%%%%%%%%%%%%%%%%%%%%%%%%%%%%%%%%%%%%%%%%%%%%%%%%%%%%%%
%%%%%%%%%%%%%%%%%%%%%%%%%%%%%%%%%%%%%%%%%%%%%%%%%%%%%%%%%%%%%%%%%%%%%%%%%%%%%%%

\end{document}